\documentclass{article} 
\usepackage{iclr2027_conference,times}

\usepackage{amsmath,amsfonts,bm}

\def\eqref#1{equation~\ref{#1}}

\def\1{\bm{1}}

\DeclareMathAlphabet{\mathsfit}{\encodingdefault}{\sfdefault}{m}{sl}
\SetMathAlphabet{\mathsfit}{bold}{\encodingdefault}{\sfdefault}{bx}{n}

\usepackage{hyperref}
\usepackage{url}

\usepackage{amsmath}
\usepackage{graphicx}
\usepackage{subcaption}
\usepackage{multirow}
\usepackage{tabularx}
\usepackage{amssymb}
\usepackage{makecell}
\usepackage{booktabs}

\title{SCRIBE: Scalable Contextual Representations of Interpretable Brain-Grounded Semantics}

\author{Camila Kolling \qquad Omer Moussa \qquad Emin Çelik \qquad Mariya Toneva\\
Max Planck Institute for Software Systems (MPI-SWS) \\
\texttt{\{ckolling, omoussa, emcelik, mtoneva\}@mpi-sws.org} \\
}

\def\methodlong{Scalable Contextual Representations of Interpretable Brain-grounded sEmantics}
\def\method{SCRIBE}

\renewcommand{\headrulewidth}{0pt} 

\iclrfinalcopy 
\begin{document}

\maketitle
\lhead{} 

\begin{abstract}
Understanding how the human brain constructs meaning from real-world language remains a fundamental challenge, as semantics depend on context that varies in natural settings. To enable the study of real-world linguistic meaning in the brain, we introduce SCRIBE, a framework grounded in a neuroscience-informed semantic space that leverages an ensemble of open-source large language models to generate interpretable, word-level semantic representations of arbitrary text. 
In a behavioral study with $921$ participants, we show that human semantic feature judgments vary systematically with context, and that \method{} captures these shifts, reaching $75$--$81\%$ of the estimated human noise ceiling.
Using brain activity recorded while participants listen to naturalistic stories, we further demonstrate that \method{} predicts neural responses in language-selective brain regions, significantly outperforming non-contextual semantic representations.
Variance partitioning reveals that SCRIBE explains unique brain-predictive information beyond syntactic features, suggesting that its representations capture aspects of linguistic meaning not accounted for by the measured syntactic features.
Together, these findings establish SCRIBE as a scalable and interpretable foundation for studying natural language semantics in the brain.
\end{abstract}

\section{Introduction}
\label{sec:intro}

Understanding how the human brain constructs meaning from language is a central goal in cognitive neuroscience and artificial intelligence. 
Semantic representations, which capture the meanings of words and concepts, are inherently dynamic and context-dependent. For example, the word ``knife'' may evoke harm and negative affect in a sentence such as ``he cut himself with a knife'', but a more neutral, functional meaning in ``he chopped vegetables with a knife''.
This context dependence  makes semantics difficult to study in the brain, as the effectively unbounded range of contexts complicates experimental design and the collection of human annotations that capture meaning as it occurs in real-world language.

Prior work has primarily proposed semantic representations along two distinct directions.
One line of research studies isolated concepts using feature-based representations grounded in cognitive and neuroscience theories~\citep{binder2016toward, mcrae2005semantic, devereux2014centre, lynott2020lancaster}.
These representations are interpretable and have been successfully used to model semantic structure and predict brain activity~\citep{mitchell2008predicting, anderson2017predicting}.
However, even when extended to sentence-level modeling~\citep{anderson2017predicting}, these approaches assume that word meanings remain fixed across contexts and combine them using simplified compositional schemes. In addition, these representations often rely on manual feature annotation and fixed vocabularies, making them difficult to scale to arbitrary natural language inputs.

A second line of work aims to study semantics using distributed representations derived from large language models (LLMs)~\citep{devlin2019bert, touvron2023llama, jiang2023mistral7b, grattafiori2024llama}.
These representations are contextual, scalable, and have been shown to predict neural responses during natural language comprehension~\citep{toneva2019interpreting, pereira2018toward}.
However, because internal LLM representations entangle semantic, syntactic, and other lexical properties of language, their alignment with the brain is difficult to attribute to semantic processing, limiting their interpretability for studying meaning in the brain.

In this work, we bridge these two approaches by introducing \methodlong~(\method), an approach for generating scalable, interpretable, context-dependent semantic representations.
Grounded in a neuroscience-informed semantic space~\citep{binder2016toward}, \method{} leverages an ensemble of open-source LLMs to extend traditional feature-based representations to arbitrary natural language contexts.

We validate \method{} along four complementary dimensions.
First, we validate \method{} scores against existing human semantic feature norms for isolated out-of-context concepts~\citep{binder2016toward}, establishing its semantic validity.
Second, we extend this evaluation to contextual semantics by collecting a new dataset of human ratings of concept--property pairs, both in and out of context, from $921$ participants.
To our knowledge, this constitutes the first human-annotated resource of context-dependent semantic feature norms, which we will publicly release.
We then show that these judgments are reliable and systematically vary with linguistic context, and that \method{} captures these shifts, reaching $75$--$81\%$ of the human noise ceiling across conditions.
Third, we demonstrate that contextual \method{} representations significantly outperform their out-of-context counterparts at predicting neural responses during naturalistic story comprehension.
We show that \method{} captures unique brain-predictive information beyond the closest interpretable representation, QAEmb~\citep{singh2025evaluating}.
Finally, variance partitioning with independently defined syntactic features reveals that \method{} retains more unique brain-predictive variance than QAEmb while exhibiting less overlap with syntactic features in language-related regions, supporting greater semantic specificity of its neural predictivity.
Together, results establish \method{} as a scalable, interpretable, context-aware foundation for studying how the brain represents meaning in natural language.

\section{Related work}
\label{section:related_work}
\textbf{Feature-based interpretable semantic representations.}
A central question in cognitive science is how to represent meaning, with a long line of work modeling concepts through explicit semantic features derived from human judgments and experiential attributes~\citep{mcrae2005semantic, devereux2014centre, lynott2020lancaster, hoffman2013semantic}. 
\citet{binder2016toward} consolidated this line of work into a comprehensive, neurobiologically grounded semantic space spanning sensory, motor, affective, social, and cognitive dimensions. 
Such interpretable representations have been used to model brain activity at both the word and sentence level~\citep{mitchell2008predicting, fernandino2015predicting, anderson2017predicting}. 
However, these approaches rely on out-of-context representations and manually annotated datasets with limited vocabularies, limiting their scalability to settings where meaning varies across contexts. 
Our work leverages LLMs to generate context-aware semantic representations, enabling a scalable and semantically grounded feature space.

\textbf{LLM-based contextual semantic representations.}
A second line of work models brain activity using contextualized LLM embeddings~\citep{toneva2019interpreting,schrimpf2021neural,goldstein2022shared}, which capture a wide range of linguistic and semantic phenomena~\citep{jawahar2019does, brown2020language}. Despite their impressive brain alignment, LLMs' distributed and high-dimensional representations are difficult to interpret and to relate explicitly to semantic processing in the brain. In fact, prior work has shown that the brain alignment of pretrained LLMs depends in part on syntax \citep{oota2023joint,fresen2024language}. In contrast, we leverage LLMs as semantic annotators to map concepts to an interpretable and cognitively grounded feature space, which can then be used to predict brain activity.

\textbf{Bridging interpretability and contextual semantic representations.}
Previous work has shown that semantic properties can be recovered from LLM representations using probes~\citep{chersoni2021decoding, turton2021deriving, turton2020extrapolating}.
However, these approaches do not assess whether modern LLMs can directly annotate such properties through prompting, analogous to the human semantic annotation procedure used to construct semantic feature spaces.
Moreover, although LLM representations are inherently context sensitive, these approaches average them across contexts rather than modeling how semantic properties vary with context~\citep{turton2021deriving}.
More recently, structured prompting has enabled LLMs to directly produce explicit, interpretable features, for example question-answering-based embedding methods such as QAEmb~\citep{singh2025evaluating}.
QAEmb represents stimuli using binary features obtained from LLM responses to yes/no questions derived from plausible neurolinguistic theories.
This makes QAEmb the closest methodological comparison to \method{}: both approaches use LLMs as annotators to transform natural-language stimuli into explicit, cognitively motivated, interpretable word-level features for brain encoding.
They differ, however, in what these features are designed to represent.
QAEmb provides a general framework spanning neuroscience-motivated linguistic properties, whereas \method{} is specifically designed to represent conceptual semantics within the independently defined neurocognitive semantic space of \citet{binder2016toward}, using LLMs to estimate how semantic properties vary with context.

Our approach bridges these directions by using LLMs as context-aware semantic annotators within a cognitively grounded feature space, yielding representations that are simultaneously interpretable, context-aware, and explicitly semantic.

\begin{figure}[t]
    \centering

    \begin{subfigure}{\linewidth}
        \centering
        \includegraphics[width=0.8\linewidth,]{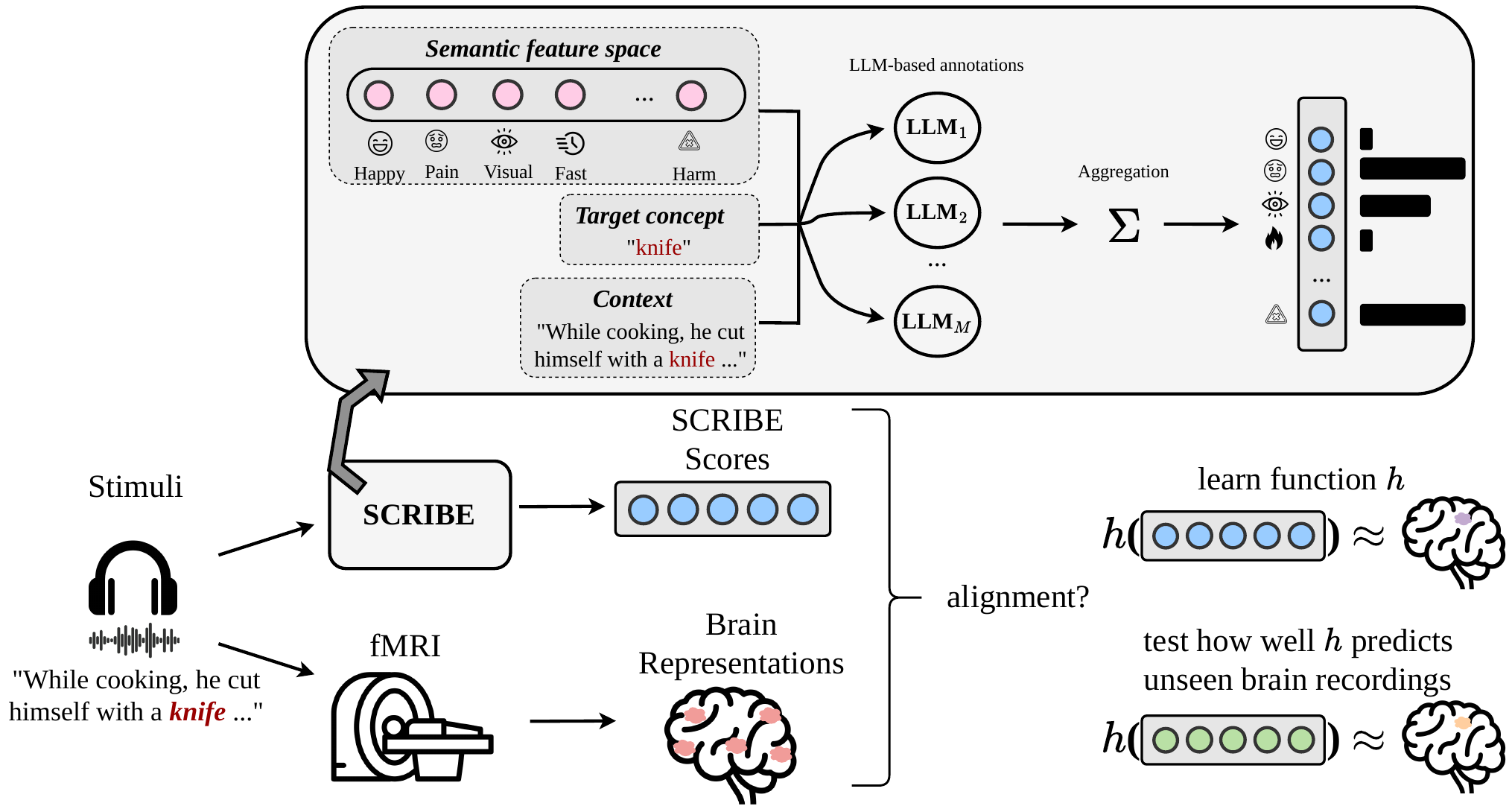}
    \end{subfigure}

    \caption{
    \method{} constructs interpretable, context-aware semantic features that are used to predict brain activity, enabling the modeling of the dynamic nature of meaning in the brain. 
    It combines three information sources: a cognitively grounded semantic feature space~\citep{binder2016toward}, a target concept, and the linguistic context in which the concept appears. 
    For each semantic property in the feature space, LLMs are prompted independently to assign a score reflecting how strongly the property applies to the concept in context. 
    These property-wise scores are then aggregated across models to produce the final representations, which are used as input to a brain encoding model that predicts fMRI responses.
    }
    \label{fig:scribe_method_schematic}
\end{figure}

\section{Methods}
\label{sec:method}

We first describe \method{} (Section~\ref{subsec:method_overview}), then how we evaluate its alignment with human semantic judgments (Section~\ref{subsec:method_human_eval}), and with brain activity (Section~\ref{subsec:modeling_brain}).

\subsection{\method: interpretable and context-aware semantic representations}
\label{subsec:method_overview}

\method{} maps a target concept, optionally embedded in a linguistic context, to a vector of interpretable semantic feature scores (Figure~\ref{fig:scribe_method_schematic}). 

\textbf{Semantic feature space.}
We build on the feature space of \citet{binder2016toward}, which defines $65$ semantic properties spanning sensory (e.g., \emph{color}, \emph{brightness}), motor (e.g., \emph{body parts}, \emph{movement}), affective (e.g., \emph{pleasantness}), and higher-level cognitive dimensions, derived from a synthesis of findings in cognitive neuroscience and psychology (Appendix~\ref{appendix:binder_features}).
Because each dimension is interpretable and neurocognitively grounded, we treat this space as a fixed set of semantic dimensions along which meaning can be expressed and modulated by context.
\citet{binder2016toward} also provide human ratings ($0$--$6$), averaged across participants, of how strongly each property applies to $535$ English words, which we use as a reference for out-of-context meaning.

\textbf{LLM-based annotation.}
Instead of manual annotation, \method{} prompts an LLM independently for each property to rate how strongly it applies to the target concept on the same $0$--$6$ scale used by \citet{binder2016toward}, mirroring their protocol (Appendix~\ref{appendix:nc_annotation_prompt}); the model outputs a single number, obtained with greedy decoding for deterministic annotations.
Prompt ablations show that results are robust to variations in prompt wording and format (Appendix~\ref{appendix:results_prompt_ablation}).
We use four open-source instruction-tuned LLMs, \texttt{llama3.1-8b}, \texttt{mistral-7b}, \texttt{qwen2.5-7b} and \texttt{llama3.1-70b}~\citep{grattafiori2024llama,jiang2023mistral7b,yang2024qwen25}, and average their scores to obtain the final representation, mitigating model-specific biases and yielding higher correlations with human judgments than any individual model (Sections~\ref{subsec:results_noncontextual_validation} and~\ref{subsec:results_contextual_validation}).

\textbf{Context-aware annotation.}
Because meaning is shaped by context, \method{} conditions each estimate on the surrounding text: given a concept and the sentence in which it appears, the LLM rates how strongly each property applies to the concept \emph{in that context} (Appendix~\ref{appendix:contextual_annotation_prompt}).
The same concept can thus receive different semantic profiles depending on its usage, while remaining in the same interpretable space; because annotation is automated, it scales to naturalistic stimuli.

\subsection{Evaluating alignment with human semantic judgments}
\label{subsec:method_human_eval}

We compare \method{} against two sets of human ratings: the out-of-context norms of \citet{binder2016toward}, and our newly collected ratings of concept--property pairs in and out of context (Section~\ref{subsec:results_contextual_validation}; Appendix~\ref{appendix:results_human_survey}).
For our newly collected dataset, we estimate the reliability of averaged human ratings using the one-way random-effects intraclass correlation $\mathrm{ICC}(1,n)$~\citep{shrout1979intraclass}, computed on $n=8$ ratings per item (a concept--property pair in a given context condition) from distinct participants and averaged over $20$ balanced subsamples; we also report $\mathrm{ICC}(1,1)$ for single rater agreement (Appendix~\ref{appendix:human_reliability}).
The estimated \emph{human noise ceiling}, $\sqrt{\mathrm{ICC}(1,n)}$, represents the expected correlation between a noise-free model and the $n$-rater mean.
We report the Pearson correlation $r$ between \method{} scores and the mean human ratings, normalized by this ceiling (with $1$ indicating human-level agreement), with bootstrap confidence intervals over items.
We also report the mean squared error (MSE), with significance assessed by permutation tests that shuffle \method{} scores across items.
For \citet{binder2016toward}, which provide only participant-averaged ratings, we report raw correlations without noise-ceiling normalization.

\subsection{Modeling semantic representations in the brain}
\label{subsec:modeling_brain}


\textbf{Brain encoding.}
We use a subset of the \emph{Moth Radio Hour} dataset~\citep{lebel2023natural}: fMRI recordings from $7$ participants listening to autobiographical stories ($10$--$15$ minutes each, TR$=2$s). This dataset is widely used in prior work~\citep{moussa2024improving, antonello2023scaling, oota2024speech} and enables direct comparison with QAEmb~\citep{singh2025evaluating}.
The encoding models are trained on $9$ stories and evaluated on $1$ held-out story (Appendix~\ref{appendix:fmri_data}).
Following standard voxel-wise encoding~\citep{oota2024speech, antonello2023scaling}, word-level \method{} features are aligned to the fMRI signal using Lanczos interpolation, augmented with time-lagged copies spanning the preceding $4$s, and used to fit ridge regression models with cross-validation.

\textbf{Evaluation.}
Prediction accuracy is the Pearson correlation between predicted and held-out responses, normalized by voxel-wise noise ceilings~\citep{lebel2023natural, moussa2024improving}.
We restrict analyses to voxels with a noise ceiling above $0.4$ and report results across all retained voxels and within language-related regions previously defined using the Glasser atlas~\citep{glasser2016multi, moussa2024improving}; differences are tested with Wilcoxon signed-rank tests (Appendix~\ref{appendix:brain_encoding}).

\textbf{Baseline.}
We compare \method{} with QAEmb~\citep{singh2025evaluating}, the closest existing interpretable stimulus representation (see Section~\ref{section:related_work} for a discussion), which represents each word using LLM responses to $35$ neuroscience-motivated yes/no questions automatically selected from an initial pool of $606$.
We use the publicly released annotations.

\subsubsection{Annotation design: word types and context size}
\label{subsubsec:annotation_design}

Since function words (e.g., articles, prepositions, auxiliaries) carry limited lexical meaning and may add noise, we annotate only content words (nouns, proper nouns, verbs, adjectives, adverbs, numerals, and pronouns), assigning zero-valued features to all other tokens.
For context, we compare windows of $w$ preceding tokens with the full preceding context, and use $w=10$.
Both choices were selected through ablation studies across individual models and the ensemble (Appendix~\ref{appendix:results_brain_wordtypes_contextlen}): broader word sets outperform more restrictive ones (e.g., the nouns, verbs, and adjectives of \citet{binder2016toward}), likely due to reduced sparsity, while gains from additional context saturate, consistent with prior work~\citep{jain2018incorporating, toneva2019interpreting}.

\section{Results}
\label{sec:results}

We evaluate \method{} along four directions: \emph{semantic validity}, whether its scores align with human ratings of isolated concepts (Section~\ref{subsec:results_noncontextual_validation}); \emph{context awareness}, whether human judgments shift with context and \method{} captures these shifts (Section~\ref{subsec:results_contextual_validation}); \emph{neural predictivity}, how well it predicts brain activity during natural language comprehension (Section~\ref{subsec:results_brain_encoding}); and \emph{semantic specificity}, whether its brain-predictive information is distinct from syntactic features (Section~\ref{subsec:results_vp}).

\textbf{Implementation details.} 
All experiments were performed on internal compute clusters, using two NVIDIA H100 PCIe GPUs with $\sim80$GB GPU memory per device.
The computation of the experiments took a total of $\sim10$ days.

\subsection{\method{} generates semantic features that align with human judgments}
\label{subsec:results_noncontextual_validation}

\begin{table}[t]
    \centering
    \caption{Alignment between SCRIBE scores and human ratings for isolated concepts (i.e., without additional context)~\citep{binder2016toward}.
    We report Pearson correlation ($r$) and mean squared error (MSE), with corresponding permutation-based $p$-values. Overall, SCRIBE aligns significantly with human semantic ratings, with the ensemble achieving the best performance.}
    \small{
    \begin{tabular}{lcccc}
    \textbf{Model} & \textbf{Pearson $\uparrow$} & \textbf{$p_r$} & \textbf{MSE $\downarrow$} & \textbf{$p_{\text{MSE}}$} \\
    \midrule
    Mistral 7B Instruct     & 0.61 & $<0.001$ & 2.16 & $<0.001$ \\
    Qwen 2.5 7B Instruct    & 0.69 & $<0.001$ & 1.32 & $<0.001$ \\
    LLaMA 3.1 8B Instruct   & 0.67 & $<0.001$ & 1.70 & $<0.001$ \\
    LLaMA 3.1 70B Instruct  & 0.74 & $<0.001$ & 1.53 & $<0.001$ \\
    Ensemble                & \textbf{0.77} & $<0.001$ & \textbf{1.07} & $<0.001$ \\
    \end{tabular}
    }
    \label{tab:nc_llm_human_alignment}
\end{table}

We first evaluate whether \method~can produce semantically meaningful representations in the absence of context, treating SCRIBE as a scalable proxy for human semantic ratings over isolated concepts.
To this end, we compare SCRIBE scores for isolated concepts (i.e., without context) to human semantic feature ratings~\citep{binder2016toward}, using the original dataset of $535$ words introduced in that work and described in Section~\ref{subsec:method_overview}, as a reference benchmark.
For each concept, we quantify the similarity between the SCRIBE scores and human ratings using Pearson correlation and mean squared error (MSE). 
We assess statistical significance using permutation tests, obtained by randomly shuffling concept–property assignments $1000$ times.

Table~\ref{tab:nc_llm_human_alignment} summarizes the results.
\method{} aligns significantly with human ratings across all evaluated models, with correlations above $0.6$ and MSEs significantly below permutation baselines (see Appendix~\ref{appendix:results_permutation_test} for further results).
The ensemble shows the strongest alignment, reaching $r=0.77$ and an MSE of $1.07$, indicating that aggregation improves correspondence with human semantic judgments.
The \citet{binder2016toward} norms themselves also agree closely with the out-of-context ratings we collect (Appendix~\ref{appendix:scribe_vs_ceiling}).
Together, these results support \method{} as a scalable approximation of human semantic ratings for isolated concepts.
We therefore use the ensemble throughout the remainder of the paper, with individual-model results reported in Appendix~\ref{appendix:results}.

\subsection{\method{} captures contextual shifts in human semantic judgments}
\label{subsec:results_contextual_validation}

\begin{table}[t]
    \centering
    \caption{
    Human contextual judgments and their alignment with the \method{} ensemble.
    \textbf{(a)} Mean semantic feature scores for human judgments (computed from all ratings) and \method{} across the low-target ($L_\mathrm{T}$), original ($O$), and high-target ($H_\mathrm{T}$) conditions, with one-sided Wilcoxon signed-rank tests over concept--property pairs for the predicted directional contrasts.
    \textbf{(b)} Human reliability and human--\method{} alignment in each condition. $\mathrm{ICC}(1,8)$: reliability of the mean of $8$ participants; ceiling $=\sqrt{\mathrm{ICC}(1,8)}$, the maximum correlation attainable given human noise. $r$: Pearson correlation between \method{} and the mean human rating; $r$/ceiling: correlation normalized by the noise ceiling, with $95\%$ bootstrap confidence intervals over concept--property pairs. MSE: mean squared error between \method{} and the mean human rating, with the chance level obtained by shuffling \method{} scores across pairs ($1000$ permutations) in parentheses. Values are averaged over $20$ balanced subsamples of $8$ ratings per item, as required for comparable reliability estimates across items. All ICCs, correlations and MSEs are significant ($p\leq.001$, permutation tests).
    }
    \label{tab:context_human_scribe}
    \small

    \begin{tabular}{lccc|ccc}
    \multicolumn{7}{c}{\textbf{(a) Contextual modulation}} \\
    \toprule
    \textbf{Annotator} & $\mathbf{L_T}$ & \textbf{O} & $\mathbf{H_T}$
    & $\mathbf{L_T<O}$ & $\mathbf{H_T>O}$ & $\mathbf{H_T>L_T}$ \\
    \midrule
    Human    & 2.67 & 3.23 & 4.23 & $<.001$ & $<.001$ & $<.001$ \\
    Ensemble & 1.58 & 2.18 & 3.54 & $<.001$ & $<.001$ & $<.001$ \\
    \bottomrule
    \end{tabular}

    \vspace{0.5em}

    \begin{tabular}{l|cc|ccc}
    \multicolumn{6}{c}{\textbf{(b) Human reliability and human--\method{} alignment}} \\
    \toprule
    & \multicolumn{2}{c|}{\textbf{Human reliability}} & \multicolumn{3}{c}{\textbf{\method{} alignment}} \\
    \textbf{Condition} & $\mathrm{ICC}(1,8)$ & \textbf{Ceiling} & \textbf{$r$} $\uparrow$ & \textbf{$r$/ceiling [95\% CI]} $\uparrow$ & \textbf{MSE (chance)} $\downarrow$ \\
    \midrule
    $\mathbf{L_T}$ & 0.74 & 0.86 & 0.69 & 0.81 [0.74, 0.86] & 2.64 (5.71) \\
    $\mathbf{O}$   & 0.84 & 0.92 & 0.73 & 0.80 [0.75, 0.85] & 2.59 (6.53) \\
    $\mathbf{H_T}$ & 0.72 & 0.85 & 0.64 & 0.75 [0.67, 0.82] & 2.31 (4.80) \\
    \bottomrule
    \end{tabular}
\end{table}

Existing semantic property norms~\citep{binder2016toward, mitchell2008predicting} characterize concepts in isolation, leaving unexplored how humans judgments of these properties vary with linguistic context.
To study this systematically, we construct a controlled dataset in which context is manipulated to increase or decrease specific concept–property associations.
For each concept–property pair, we generate two contextualized sentences using an instruction-tuned LLM (\texttt{llama3.1-8b}): a low-target context ($L_\mathrm{T}$) designed to decrease the association and a high-target context ($H_\mathrm{T}$) designed to increase it (see Appendix~\ref{appendix:synthetic_dataset} for details).

To examine how these contextual manipulations affect human semantic judgments, we conducted a large-scale human behavioral study, in which $921$ participants (after quality control) rated $323$ concept–property pairs under the original out-of-context condition ($O$) and the corresponding $L_\mathrm{T}$ and $H_\mathrm{T}$ contexts, yielding nearly $12{,}000$ ratings and on average $12$ ratings per item (range $8$--$20$; see Appendix~\ref{appendix:results_human_survey}).

As shown in Table~\ref{tab:context_human_scribe}(a), human judgments change systematically with context, following the expected ordering $H_\mathrm{T} > O > L_\mathrm{T}$, with all directional comparisons significant ($p<.001$).
These findings demonstrate that human semantic feature judgments are systematically modulated by linguistic context.
To our knowledge, these data constitute the first human-annotated context-dependent semantic feature norms, which we will publicly release.

\textbf{Human judgments of semantics in context are reliable.}
Agreement between individual participants is low to moderate ($\mathrm{ICC}(1,1)=0.40$, $0.27$ and $0.24$ for $O$, $L_\mathrm{T}$ and $H_\mathrm{T}$), as expected for single ratings of graded semantic properties, but the average of $n=8$ participants is reliable ($\mathrm{ICC}(1,8)=0.84$, $0.74$ and $0.72$), far above chance (permutation test, $p\leq.001$; Table~\ref{tab:context_human_scribe}(b)).
Out-of-context judgments are the most consistent, suggesting greater variability in how participants interpret concepts within linguistic contexts than in isolation.
The resulting human noise ceilings are $0.92$ ($O$), $0.86$ ($L_\mathrm{T}$) and $0.85$ ($H_\mathrm{T}$).

\textbf{\method{} captures contextual shifts and approaches human-level agreement.}
As shown in Table~\ref{tab:context_human_scribe}(a), \method{} reproduces the human ordering $H_\mathrm{T} > O > L_\mathrm{T}$, with all directional comparisons significant ($p<.001$).
Beyond capturing the direction of these shifts, \method{} also aligns with human ratings at the level of individual concept--property pairs.
\method{} scores correlate with the mean human rating in all three conditions ($r=0.73$, $0.69$ and $0.64$ for $O$, $L_\mathrm{T}$ and $H_\mathrm{T}$), reaching $80\%$, $81\%$ and $75\%$ of the human noise ceiling, respectively (Table~\ref{tab:context_human_scribe}(b)).
In other words, \method{} is as consistent with the human consensus as the average of $2.6$--$5.1$ human annotators (Appendix~\ref{appendix:scribe_vs_ceiling}).
For reference, the out-of-context norms of \citet{binder2016toward} reach $84\%$ of the ceiling on the same items, so \method{} is close to the agreement between two independently collected sets of human ratings.
\method{} also achieves significantly lower MSE than chance in all conditions (Table~\ref{tab:context_human_scribe}(b); permutation test, $p\leq.001$).
Part of this error comes from \method{} rating properties about one point lower than humans on average (Table~\ref{tab:context_human_scribe}(a); Appendix~\ref{appendix:scribe_vs_ceiling}).
The ensemble achieves the highest correlation with human ratings among all models in every condition (Appendix~\ref{appendix:scribe_vs_ceiling}).

Together, these results establish that human semantic feature judgments vary systematically with context and that \method{} captures these shifts while approaching the agreement observed among human annotators.

\subsection{Context-aware \method{} scores are good predictors of brain activity}
\label{subsec:results_brain_encoding}

\begin{figure}[t]
    \centering

    \begin{subfigure}{0.58\linewidth}
        \centering
        \includegraphics[width=\linewidth]{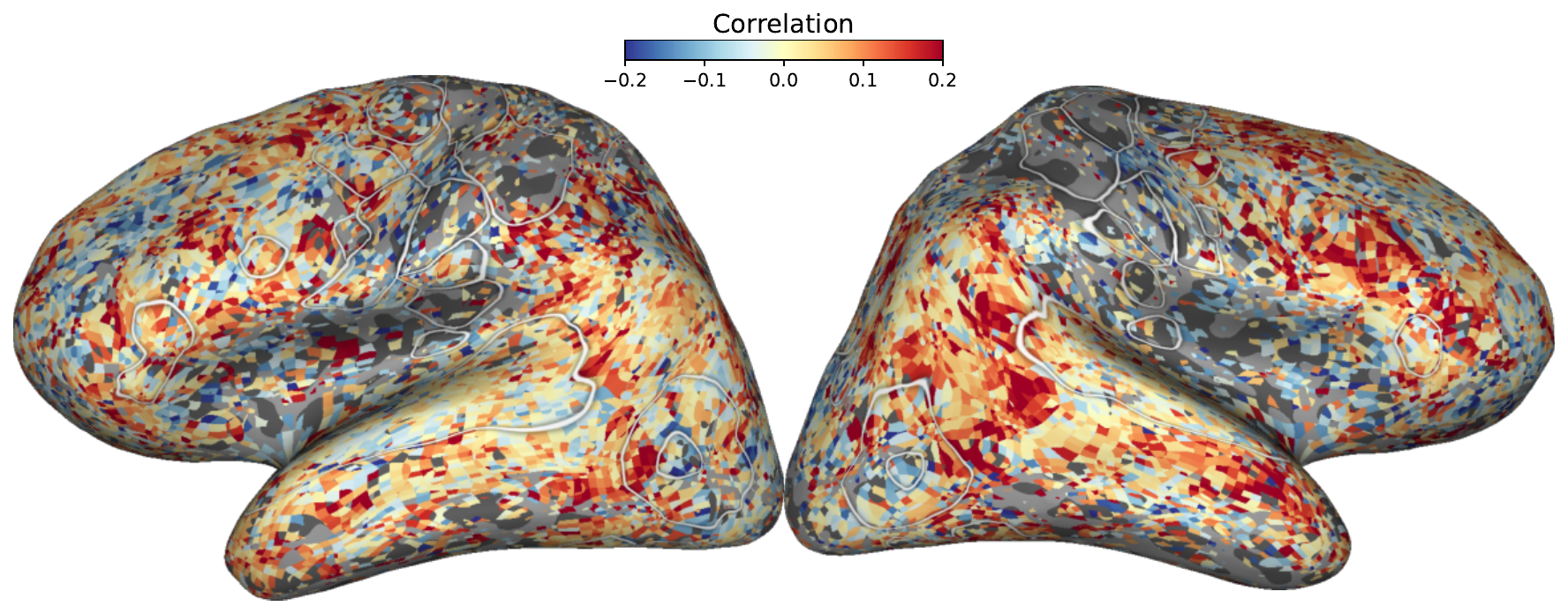}
        \caption{Difference between contextual and out-of-context alignment}
        \label{fig:diff_brainplot_subj3}
    \end{subfigure}
    \hfill
    \begin{subfigure}{0.4\linewidth}
        \centering
        \includegraphics[width=\linewidth]{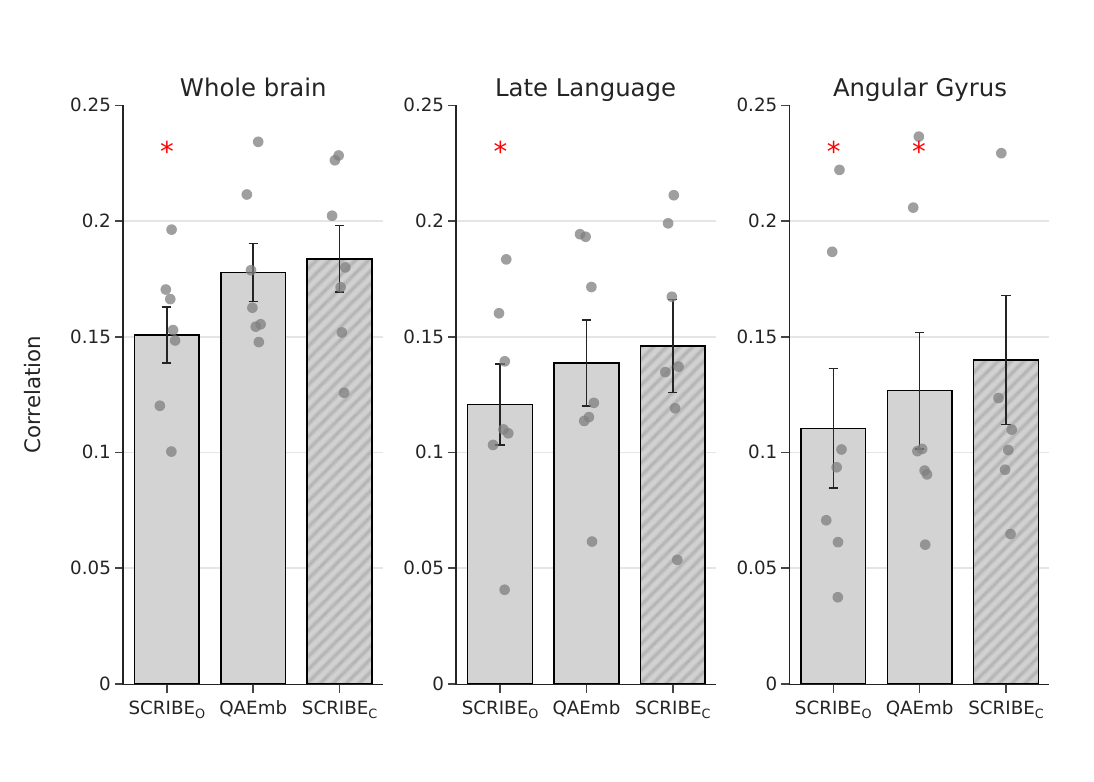}
        \caption{Brain alignment scores}
        \label{fig:barplots_overall_and_langreg}
    \end{subfigure}
    
    \caption{
    \textbf{(a)} Voxel-wise differences in brain alignment between contextual and out-of-context variants of \method{} ($\mathrm{SCRIBE}_\mathrm{C} - \mathrm{SCRIBE}_\mathrm{O}$) for a representative subject (subject 3).
    Red regions show higher predictive accuracy for the contextual model.
    Improvements due to contextualization are predominantly observed in language-related cortical regions.
    \textbf{(b)} Brain alignment for $\mathrm{SCRIBE}_\mathrm{O}$, QAEmb, and $\mathrm{SCRIBE}_\mathrm{C}$ across the whole brain and language-related ROIs.
    Bars represent mean performance across subjects, with individual subject values overlaid and error bars indicating the standard error. 
    Red asterisks (\textcolor{red}{$\ast$}) denote statistically significant differences (Wilcoxon signed-rank test). 
    Contextual \method{} ($\mathrm{SCRIBE}_\mathrm{C}$) consistently outperforms its out-of-context ($\mathrm{SCRIBE}_\mathrm{O}$) variant, while achieving performance comparable to QAEmb.
    }
    \label{fig:brain_encode_diff_binder_vs_scribe}
\end{figure}

We next assess whether the contextual semantic scores produced by \method{} capture aspects of meaning reflected in brain activity.

\textbf{Contextualization improves brain alignment of SCRIBE scores.}
We isolate the contribution of context to the neural predictivity of \method{} by comparing $\mathrm{SCRIBE}_\mathrm{O}$ (out-of-context), which estimates semantic scores from the target concept alone, with $\mathrm{SCRIBE}_\mathrm{C}$ (contextual), which additionally conditions scores on the surrounding text.
Because both variants use the same semantic feature space and annotation procedure, this comparison isolates the effect of conditioning semantic representations on linguistic context.
All results use content words and a context window of $w=10$ tokens (see Section~\ref{subsubsec:annotation_design}).

Figure~\ref{fig:diff_brainplot_subj3} shows the spatial distribution of voxel-wise differences in brain alignment for a representative subject, with higher alignment for $\mathrm{SCRIBE}_\mathrm{C}$ concentrated in language-related regions and more pronounced in the right hemisphere, consistent with the patterns observed in other subjects (Appendix~\ref{appendix:results_brain_encoding_diff_plots}).
We further quantify these observations in Figure~\ref{fig:barplots_overall_and_langreg}.
Across participants, $\mathrm{SCRIBE}_\mathrm{C}$ significantly outperforms $\mathrm{SCRIBE}_\mathrm{O}$ both across the whole brain and within the examined language-related ROIs, demonstrating that incorporating linguistic context improves the neural predictivity of \method{}.
This improvement suggests that context-dependent changes in semantic feature values capture additional information about neural responses that cannot be explained by static representations of the same concepts.

\textbf{\method{} predicts unique variance in semantic brain regions.}
We further compare the neural representations captured by $\mathrm{SCRIBE}_\mathrm{C}$ and QAEmb (Figure~\ref{fig:barplots_overall_and_langreg}).
Unlike \method{}, which represents conceptual meaning within a neuroscience-informed semantic space, QAEmb uses a broader set of neuroscience-motivated linguistic features.
While the two methods achieve comparable whole-brain alignment, $\mathrm{SCRIBE}_\mathrm{C}$ significantly outperforms QAEmb in the AG, a region strongly implicated in semantic processing~\citep{price2015converging}.
To investigate whether the two representations capture complementary brain-predictive information, we perform variance partitioning to separate their shared and uniquely explained variance.

\begin{figure}[t]
    \centering
    \begin{subfigure}{0.6\linewidth}
        \centering
        \includegraphics[width=\linewidth]{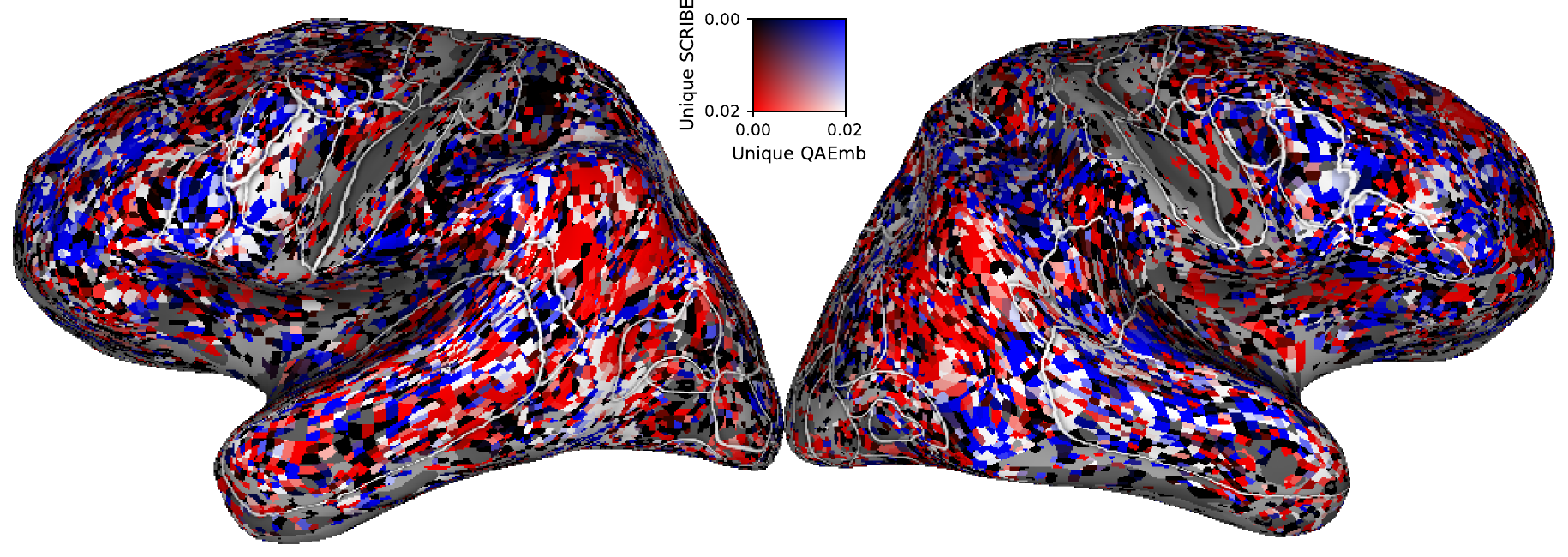}
        \caption{Whole-brain spatial distribution.}
        \label{fig:contrastive_plot_qaemb_vs_scribe_subj2}
    \end{subfigure}
    \hfill
    \begin{subfigure}{0.35\linewidth}
        \centering
        \includegraphics[width=\linewidth]{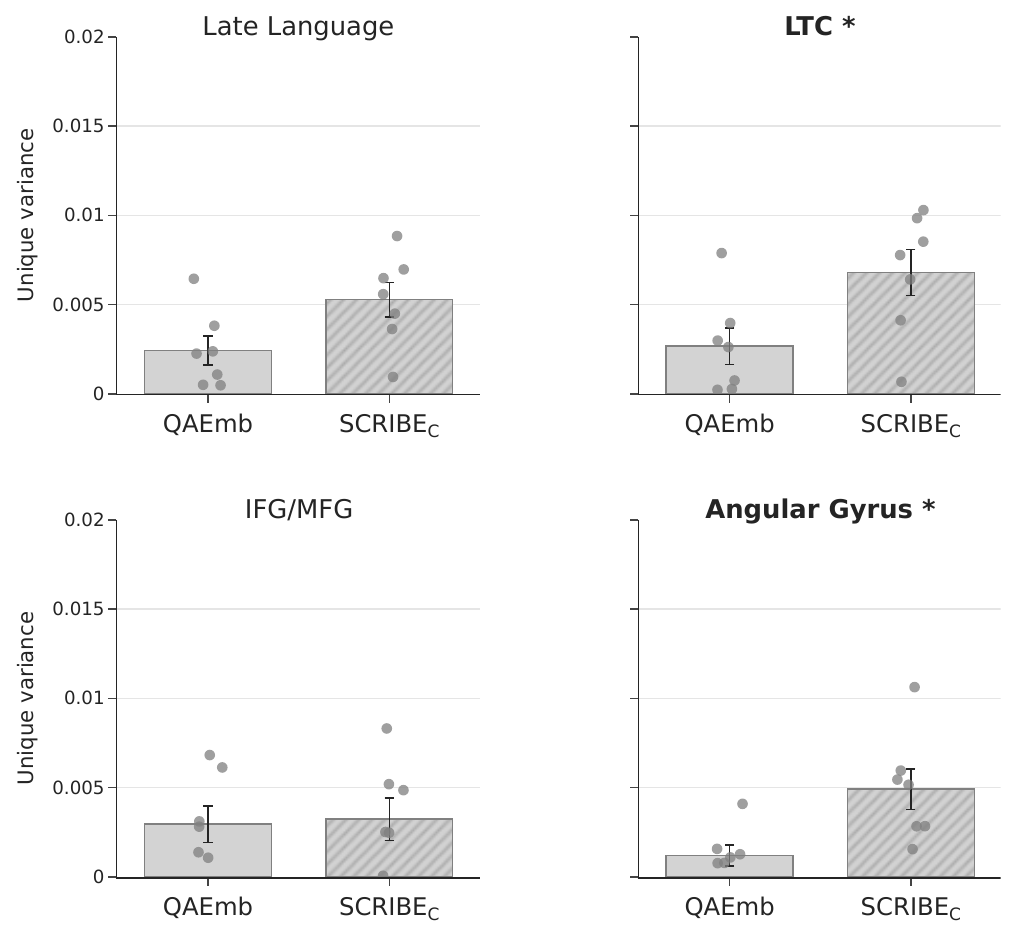}
        \caption{ROI-level aggregation.}
        \label{fig:barplots_qaemb_vs_scribe}
    \end{subfigure}
    
    \label{fig:qaemb_vs_scribe}
    \caption{
    Unique neural variance explained by $\mathrm{SCRIBE}_\mathrm{C}$ and QAEmb.
    \textbf{(a)} Spatial distribution of voxel-wise unique variance for a representative subject (subject 2).
    Red indicates greater unique variance explained by $\mathrm{SCRIBE}_\mathrm{C}$, whereas blue indicates greater unique variance explained by $\mathrm{QAEmb}$.
    \textbf{(b)} Region-wise comparison of uniquely explained variance across semantically relevant ROIs.
    Bars show mean performance across subjects, with individual values overlaid and error bars indicating the standard error. 
    $\mathrm{SCRIBE}_\mathrm{C}$ explains more unique variance than $\mathrm{QAEmb}$ on average in all four ROIs, with significant differences in LTC ($p=0.039$) and AG ($p=0.023$).
    }
\end{figure}

Figure~\ref{fig:barplots_qaemb_vs_scribe} shows that $\mathrm{SCRIBE}_\mathrm{C}$ explains more unique variance on average than QAEmb across the examined ROIs, with significant differences in the lateral temporal cortex (LTC) and the AG. The spatial distribution for a representative subject is shown in Figure~\ref{fig:contrastive_plot_qaemb_vs_scribe_subj2} (see Appendix~\ref{appendix:qaemb_and_SCRIBE} for all subjects).
Notably, both regions have been implicated in semantic processing and integration,  with the AG associated with combinatorial semantics~\citep{price2015converging} and the LTC semantic representation and integration~\citep{visser2010semantic,pallier2011cortical,pylkkanen2020neural}. 

Together, these results suggest that $\mathrm{SCRIBE}_\mathrm{C}$ captures neural variance that is not explained by QAEmb, particularly in regions associated with semantic processing, despite their comparable overall brain alignment.
Additional analyses show that the representations are complementary and that \method{}'s performance is robust after dimensionality reduction (Appendix~\ref{appendix:qaemb_and_SCRIBE} and~\ref{appendix:feature_selection}).

\subsection{\method{} captures brain-predictive information beyond syntactic features}
\label{subsec:results_vp}

Although these results demonstrate that \method{} captures unique brain-predictive information in regions associated with semantic processing, they do not establish whether this information is specifically semantic.
Because semantic information is difficult to isolate from other linguistic information in natural language, we assess the relative semantic specificity of \method{} indirectly by asking how much of its brain-predictive variance remains after accounting for independently defined syntactic features.
We focus on syntax because prior work has attributed part of the brain alignment of language-model representations to syntactic rather than semantic information, and its contribution to model--brain alignment remains debated~\citep{oota2023joint, fresen2024language, kauf2024lexical}.
It is also a plausible confound for \method{} itself, since annotation is restricted to content-word part-of-speech categories and several semantic dimensions covary with grammatical class (e.g., Motion with verbs).
We thus construct a syntactic control space comprising part-of-speech tags, dependency length, and tree depth, and use variance partitioning to quantify its unique and shared contributions with \method{} and QAEmb (more details in Appendix~\ref{appendix:results_variance_partitioning}).

\textbf{\method{} captures more unique and less syntax-related neural variance than QAEmb.}
After accounting for the measured syntactic features, $\mathrm{SCRIBE}_\mathrm{C}$ predicts more unique variance than QAEmb across language-related ROIs, reaching significance in late language regions overall, and specifically the LTC and AG (Figure \ref{fig:vp_synt_feats_qaemb_vs_scribe}b).
At the same time, $\mathrm{SCRIBE}_\mathrm{C}$ shares significantly less variance with syntax for predicting the LTC and AG (Figure~\ref{fig:vp_synt_feats_qaemb_vs_scribe}a). 
Together, these findings provide complementary evidence for greater semantic specificity: \method{} retains more predictive information after accounting for syntax, while exhibiting less overlap with the syntactic control space.
We extend the analysis to Llama2-7b representations (Appendix~\ref{appendix:llm_embeddings_vs_SCRIBE}).
$\mathrm{SCRIBE}_\mathrm{C}$ outperforms non-contextual Llama2-7b embeddings, while contextual Llama2-7b embeddings achieve higher brain predictivity
(Figure~\ref{fig:appendix_llama2-b_scribe_qaemb_corrs}).
However, contextual LLM embeddings encode a broader mixture of linguistic information. 
Notably, $\mathrm{SCRIBE}_\mathrm{C}$ shares significantly less variance with syntactic features than Llama2-7b in IFG/MFG (Figure~\ref{fig:shared_vp_synt_feats_llama}), a region often associated with syntactic processing~\citep{dapretto1999form, heim2003distributed, fiebach2005revisiting, friederici2011brain}.

\section{Discussion}
\label{sec:conclusion}

Understanding how semantic representations are constructed and modulated in the brain requires models that are both \emph{context-aware} and \emph{interpretable}.
In this work, we introduced \method{}, an approach that combines a cognitively grounded semantic feature space with LLM-based annotation to produce contextualized, interpretable semantic representations for natural language. 
By using LLMs as scalable semantic annotators, \method{} preserves the interpretability of theory-driven feature spaces while extending them beyond isolated concepts to arbitrary words in context.

Our human study first establishes that semantic feature judgments are context-dependent: across $921$ participants, controlled linguistic contexts systematically shifted ratings along the targeted semantic dimensions, and these judgments were reliable when averaged across participants.
To our knowledge, these data constitute the first human-annotated context-dependent semantic feature norms and will be publicly released. 
\method{} aligns with human semantic judgments both for isolated concepts and across contextual conditions, reaching $75$--$81\%$ of the human noise ceiling, comparable to the agreement between two independently collected sets of human ratings.

At the neural level, incorporating context significantly improves \method{}’s prediction of fMRI responses relative to its out-of-context representation, particularly in language-related cortical regions.
Contextual \method{}  explains more unique neural variance than the closest existing method, with significant differences in semantic regions, such as the LTC and AG. 
After accounting for independently defined syntactic features, \method{} retains more unique brain-predictive variance than QAEmb while sharing less variance with syntax in these regions, showing that its additional brain-predictive information persists after accounting for the measured syntactic features.

More broadly, our results show that interpretability and contextuality need not be treated as opposing goals. 
Prior feature-based models offer interpretable dimensions but rely on static out-of-context annotations, while LLM representations are contextual and scalable but difficult to attribute to specific semantic content.
\method{} bridges this gap by representing meaning along explicit, human-interpretable dimensions that dynamically vary with context, providing a scalable framework for studying contextual semantics in the brain.

\textbf{Limitations and future work.}
\method{} scores depend on the underlying LLMs and the prompting strategy.
To assess sensitivity to prompt formulation, we performed ablations, which show that performance is largely robust across the tested variations (Appendix~\ref{appendix:results_prompt_ablation}).
To reduce model-specific variability and biases, \method{} aggregates scores across multiple LLMs, analogous to aggregating human judgments; the ensemble correlates more strongly with human ratings than any individual model (Section~\ref{subsec:results_contextual_validation}).
While \method{} captures the relative ordering of human judgments, its scores are systematically lower than human ratings on average, so applications relying on absolute scores may require calibration.
Moreover, individual human judgments of context-dependent properties are noisy, and our conclusions hold at the level of averaged ratings rather than individual annotators.
In addition, our contextual stimuli are controlled, LLM-generated sentences designed to shift specific properties; validating \method{} against human judgments of naturalistic text remains an important next step.
Finally, the use of a fixed semantic space~\citep{binder2016toward} constrains representations to a predefined set of dimensions, potentially missing aspects of meaning.
However, this design choice provides a principled and interpretable basis grounded in cognitive neuroscience, enabling direct comparison with human semantic judgments and prior work.

These limitations suggest several directions for future work, including extending \method~to more diverse datasets and domains, integrating it with mechanistic interpretability methods to link semantic features to model internals, and potentially expanding the semantic space to capture a broader range of semantic dimensions.
Lastly, our syntactic control space is intentionally minimal; richer structural representations~\citep{reddy2021can, caucheteux2021disentangling} and low-level or lexical-statistical features~\citep{deniz2019representation} remain to be tested.

\section*{Acknowledgments}
\label{sec:acknowledgements}

This work was partially funded by the German Research Foundation (DFG) - DFG Research Unit FOR 5368 and by the Max Planck Institute for Software Systems graduate center.
We thank Gabriele Merlin for providing valuable feedback on earlier versions of this work.

\section*{Ethics Statement}
This work includes a human behavioral study conducted through Prolific. 
Participants provided informed consent and were compensated at a rate of USD 12 per hour (USD 2 for a study of approximately 10 minutes).
The study was approved by the Ethical Review Board at the authors' institution.
No personally identifiable information was collected or released.
The resulting contextual semantic ratings will be released in anonymized form at the level of individual ratings, with participant identifiers replaced by random codes.

\section*{Reproducibility Statement}
We provide detailed descriptions of \method{}, the semantic annotation procedure, model and prompt specifications, contextual stimulus construction, human behavioral study, brain-encoding analyses, and statistical procedures in the main text and Appendix.
We will release the contextual human annotations and code necessary to reproduce the analyses and experiments presented in this work.

\section*{AI Use Statement}
Generative AI tools were used in this work to generate controlled linguistic contexts for the synthetic dataset, as described in the paper, and to assist with the editing and refinement of the manuscript.
All AI-assisted outputs used in the research were reviewed and verified by the authors.
The authors take full responsibility for the methods, analyses, claims, and final content of the paper.

\bibliography{iclr2027_conference}
\bibliographystyle{iclr2027_conference}

\clearpage
\appendix

\clearpage
\section{Methods}
\label{appendix:methods}

\subsection{Cognitively-grounded, interpretable semantic feature space}
\label{appendix:binder_features}

\textbf{Semantic feature space.}

To ground our approach in interpretable and cognitively motivated representations, we build on the semantic feature space introduced by \citet{binder2016toward}, which provides a cognitively and neurobiologically informed characterization of conceptual meaning.
In this framework, concepts are represented as vectors over a fixed set of $65$ semantic properties spanning multiple domains, including perceptual (e.g., vision, audition, somatosensation), action-related (e.g., motor effectors and interaction), spatial, temporal, and abstract properties (e.g., social, cognitive, and affective features), as summarized in Table~\ref{tab:appendix_binder_features}. 
These features are grounded in converging evidence from cognitive neuroscience, which suggests that conceptual knowledge is distributed across modality-specific and heteromodal brain systems.
Accordingly, the semantic dimensions were shown by \citet{binder2016toward} to exhibit a structured organization, clustering into groups that align with known functional divisions of the cortex, such as sensory, motor, and affective systems.
This representation offers a principled and interpretable characterization of semantic content: each dimension corresponds to a well-defined aspect of meaning, enabling direct analysis of how specific properties contribute to downstream predictions.

In this work, we build on this semantic feature space, preserving its interpretability while extending it to contextual settings using LLM-based annotation.

\begin{figure}[t]
    \centering

    \begin{subfigure}{0.8\linewidth}
        \centering
        \includegraphics[width=\linewidth]{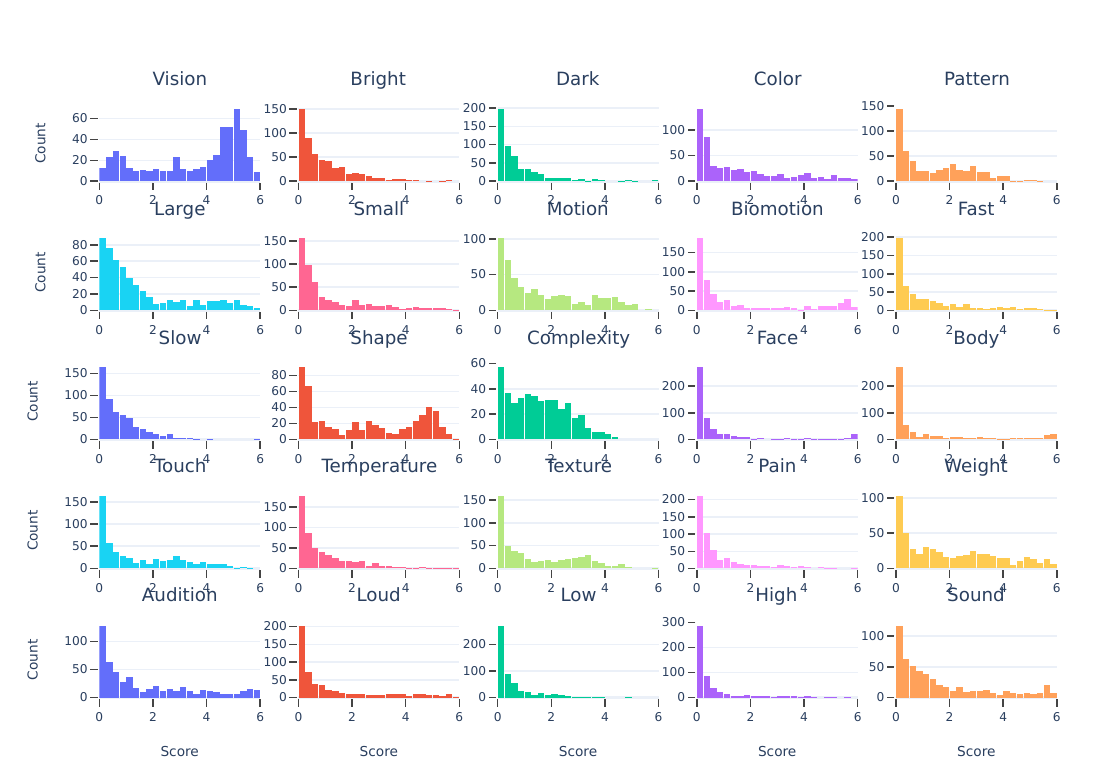}
    \end{subfigure}
    \begin{subfigure}{0.8\linewidth}
        \centering
        \includegraphics[width=\linewidth]{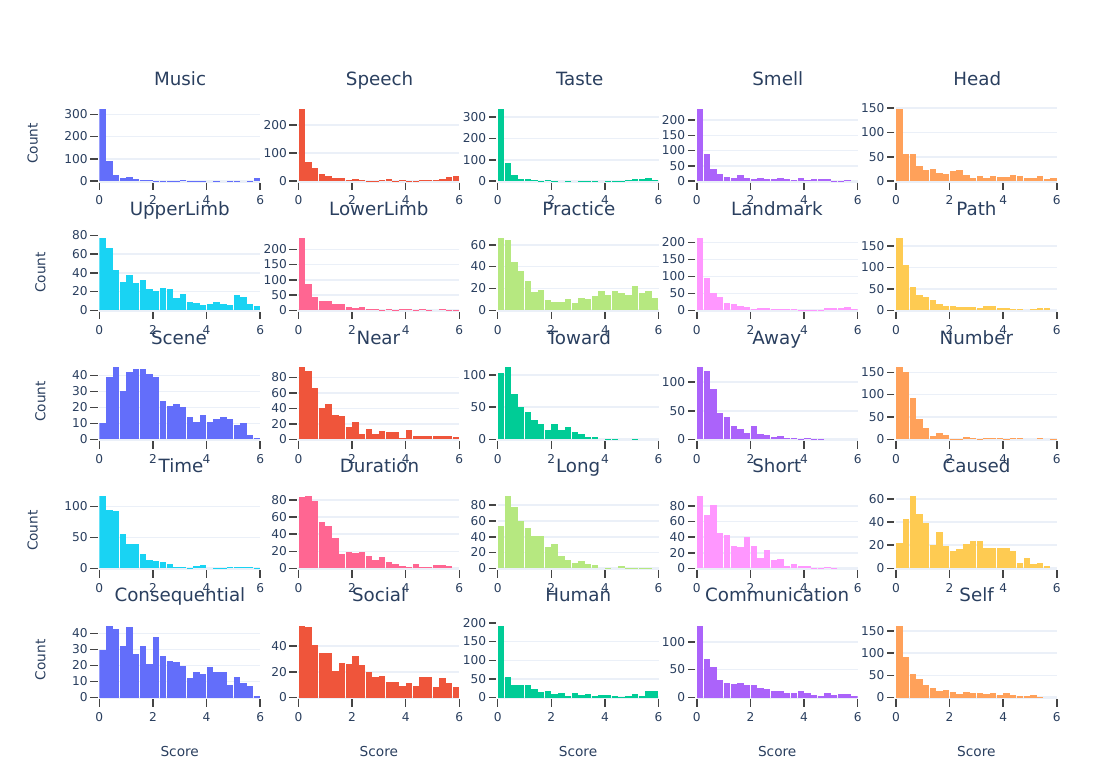}
    \end{subfigure}
    \begin{subfigure}{0.8\linewidth}
        \centering
        \includegraphics[width=\linewidth]{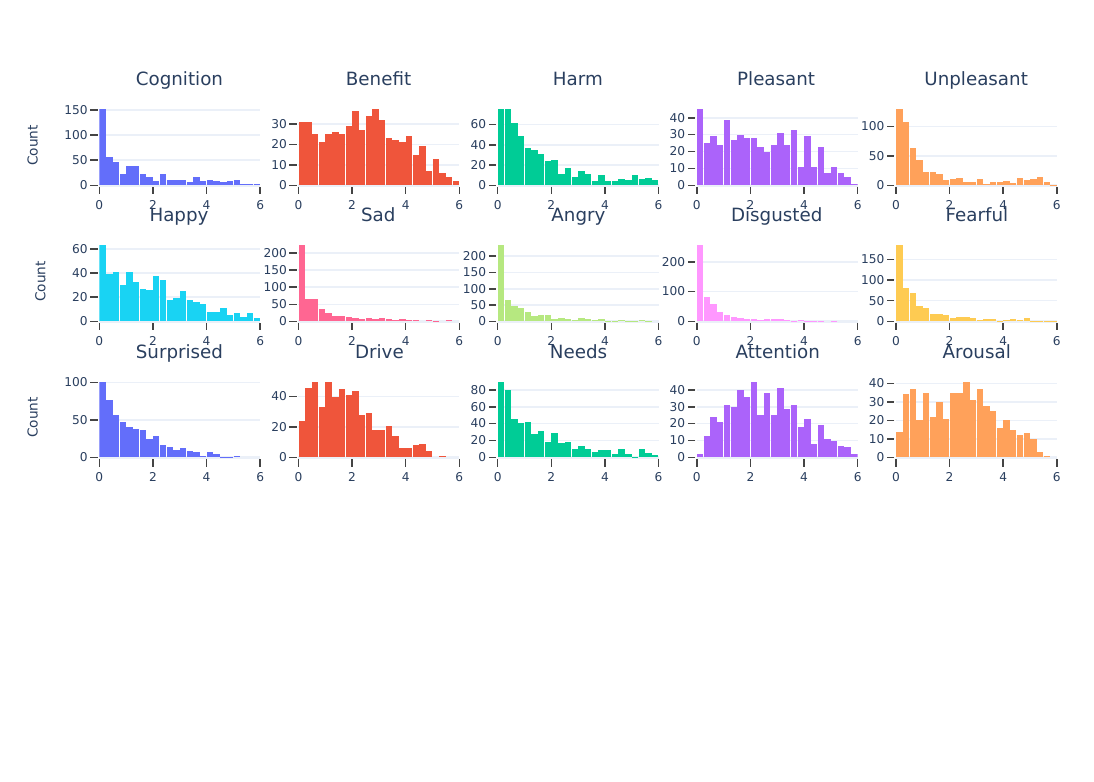}
    \end{subfigure}
    \vspace{-3cm}
    \caption{Distribution of semantic feature scores derived from human ratings in the \citet{binder2016toward} dataset (composed of $535$ words), shown for each of the $65$ semantic properties.}
    \label{fig:appendix_binder_scores_distribution_per_property}
\end{figure}

\textbf{Human semantic ratings for out-of-context concepts.}
The \citet{binder2016toward} dataset consists of $535$ out-of-context concepts ($434$ nouns, $62$ verbs, and $39$ adjectives).
Each concept is annotated along the $65$ semantic dimensions using human ratings, where participants evaluate the degree to which a concept is associated with specific properties (e.g., \textit{something that you can easily see}, \textit{something large in size}, \textit{something that makes one feel happy}, etc.).
Aggregating these ratings across participants yields continuous-valued feature vectors, forming interpretable semantic representations in the shared feature space.
Figure~\ref{fig:appendix_binder_scores_distribution_per_property} shows the distribution of semantic scores obtained from the aggregated human ratings for each property.

\begin{table*}[t]
\centering
\scriptsize
\caption{Overview of the semantic properties from the feature space of \citet{binder2016toward}, grouped by modality and cognitive domain. Each concept is represented as a vector over these interpretable semantic features. The table lists the $68$ properties that are queried individually: the three bipolar properties \emph{Temperature}, \emph{Texture}, and \emph{Weight} of the original $65$-dimensional space are each split into their two poles (Hot/Cold, Smooth/Rough, Light/Heavy), and the two resulting scores are recombined by taking their maximum, yielding the final $65$ properties (Appendix~\ref{appendix:results_human_survey}).}
\label{tab:appendix_binder_features}
\begin{tabular}{lllll}
\toprule
\textbf{Name} & \textbf{Type} & \textbf{Modality} & \textbf{Submodality} & \textbf{Description} \\
\midrule
Vision & Sensory & Vision & General & Something that you can easily see \\
Bright & Sensory & Vision & Intensity & Visually light or bright \\
Dark & Sensory & Vision & Intensity & Visually dark \\
Color & Sensory & Vision & Surface & Has a characteristic or defining color \\
Pattern & Sensory & Vision & Surface & Has a characteristic visual texture or surface pattern \\
Large & Sensory & Vision & Size & Large in size \\
Small & Sensory & Vision & Size & Small in size \\
Motion & Sensory & Vision & Motion & Shows visually observable movement \\
Biomotion & Sensory & Vision & Motion & Shows movement like that of a living thing \\
Fast & Sensory & Vision & Motion & Shows fast visible movement \\
Slow & Sensory & Vision & Motion & Shows slow visible movement \\
Shape & Sensory & Vision & Shape & Has a characteristic visual shape or form \\
Complexity & Sensory & Vision & Shape & Visually complex \\
Face & Sensory & Vision & Shape & Has a human or human-like face \\
Body & Sensory & Vision & Shape & Has human or human-like body parts \\
Touch & Sensory & Somatic & General & Can be easily recognized by touch \\
Hot & Sensory & Somatic & Surface & Hot to the touch \\
Cold & Sensory & Somatic & Surface & Cold to the touch \\
Smooth & Sensory & Somatic & Surface & Has a smooth texture to the touch \\
Rough & Sensory & Somatic & Surface & Has a rough texture to the touch \\
Light & Sensory & Somatic & Proprioception & Light in weight \\
Heavy & Sensory & Somatic & Proprioception & Heavy in weight \\
Pain & Sensory & Somatic & Nociception & Associated with pain or physical discomfort \\
Audition & Sensory & Audition & General & Something that you can easily hear \\
Loud & Sensory & Audition & Intensity & Makes a loud sound \\
Low & Sensory & Audition & Frequency & Has a low-pitched sound \\
High & Sensory & Audition & Frequency & Has a high-pitched sound \\
Sound & Sensory & Audition & Shape & Has a characteristic or recognizable sound \\
Music & Sensory & Audition & Shape & Makes a musical sound \\
Speech & Sensory & Audition & Phoneme & Someone or something that talks \\
Taste & Sensory & Gustation & Quality & Has a characteristic or defining taste \\
Smell & Sensory & Olfaction & Quality & Has a characteristic or defining smell \\
Head & Motor & Motor & Motor & Associated with actions using the face, mouth, or tongue \\
UpperLimb & Motor & Motor & Motor & Associated with actions using the arm, hand, or fingers \\
LowerLimb & Motor & Motor & Motor & Associated with actions using the leg or foot \\
Practice & Motor & Motor & Motor & A physical object one has personal experience using \\
Landmark & Spatial & Navigation & Navigation & Has a fixed location, as on a map \\
Path & Spatial & Navigation & Navigation & Shows changes in location along a direction or path \\
Scene & Spatial & Scene & Scene & Brings to mind a particular setting or physical location \\
Near & Spatial & Self Space & Proximity & Often physically near in everyday life \\
Toward & Spatial & Self Space & Toward & Associated with movement toward or into oneself \\
Away & Spatial & Self Space & Away & Associated with movement away from or out of oneself \\
Number & Number & Number & Number & Associated with a specific number or amount \\
Time & Event & Temporal & Time & Occurs at a typical or predictable time \\
Duration & Event & Temporal & Duration & Has a predictable duration, whether short or long \\
Long & Event & Temporal & Long & Lasts for a long period of time \\
Short & Event & Temporal & Short & Lasts for a short period of time \\
Caused & Event & Causal & Causal & Caused by a clear preceding event, action, or situation \\
Consequential & Event & Causal & Causal & Likely to have consequences \\
Social & Event & Social & Social & Involves an interaction between people \\
Human & Cognition & Social & Theory of Mind & Has human or human-like intentions, plans, or goals \\
Communication & Cognition & Social & Communication & Used by people to communicate \\
Self & Cognition & Self & Self & Related to one's own view of oneself \\
Cognition & Cognition & Cognition & Cognition & A form of mental activity or function of the mind \\
Benefit & Evaluation & Cognitive & Positive & Could help or benefit oneself or others \\
Harm & Evaluation & Cognitive & Negative & Could cause harm to oneself or others \\
Pleasant & Evaluation & Affective & Positive & Something one finds pleasant \\
Unpleasant & Evaluation & Affective & Negative & Something one finds unpleasant \\
Happy & Emotion & Positive & High & Makes one feel happy \\
Sad & Emotion & Negative & Low & Makes one feel sad \\
Angry & Emotion & Negative & High & Makes one feel angry \\
Disgusted & Emotion & Negative & High & Makes one feel disgusted \\
Fearful & Emotion & Negative & High & Makes one feel afraid \\
Surprised & Emotion & Neutral & High & Makes one feel surprised \\
Drive & Drive & General & General & Motivates one to do something \\
Needs & Drive & Basic & Basic & Would be hard to live without \\
Attention & Attention & General & General & Grabs one's attention \\
Arousal & Attention & General & Arousal & Makes one feel alert, activated, excited, or keyed up \\
\bottomrule
\end{tabular}
\end{table*}

\clearpage
\subsection{LLM-based semantic annotations}
\label{appendix:prompt_template_get_scores}

\method{} uses LLMs to estimate semantic feature scores by prompting them to assess how strongly a target concept is associated with each semantic property.
This procedure is applied both in an \emph{out-of-context} setting, where the word is evaluated in isolation without providing any contextual sentence, and in a \emph{contextual} setting, where the word is evaluated within a specific contextual sentence.
In both settings, scores are generated with greedy decoding (temperature $0$), making annotations deterministic, and the first number in the model output is parsed as the score.

\subsubsection{Out-of-context semantic annotation}
\label{appendix:nc_annotation_prompt}

In the out-of-context setting, \method{} obtains semantic representations using LLM-generated annotations computed independently for each concept-property pair.
For a given target concept, the model is prompted for each semantic property in the semantic feature space (described in Appendix~\ref{appendix:binder_features}), resulting in $65$ queries.
Each query asks the model to rate how strongly the word is associated with the corresponding property, returning a scalar score in the range $[0,6]$.
These scores are then aggregated to form a semantic feature vector for the concept, which is directly comparable to the human ratings of \citet{binder2016toward} when evaluated on the same set of concepts.

The prompt takes the following form:
\begin{quote}
\texttt{You are an expert in rating to what degree words are associated with semantic properties. \\
To what degree is the word ``\{word\}'' associated with \{property relation\} \{property explanation\}? \\
Provide only a single number for the rating of the word ``\{word\}'' on a scale from 0 (not at all associated or not applicable) to 6 (very strongly associated). \\
Do not include any explanation or text, just the number.\\
Output:
}
\end{quote}

An example prompt for the concept word ``bus'' and property ``motion'' is shown below:
\begin{quote}
\small{\textit{You are an expert in rating to what degree words are associated with semantic properties. \\
To what degree is the word ``bus'' associated with showing a lot of visually observable movement? \\
Provide only a single number for the rating of the word ``bus'' on a scale from 0 (not at all associated or not applicable) to 6 (very strongly associated). \\
Do not include any explanation or text, just the number.\\
Output:
}}
\end{quote}

\subsubsection{Ablation of prompt components}
\label{appendix:results_prompt_ablation}

LLMs are known to be sensitive to prompt wording, yet our main experiments use a single prompt formulation.
To assess the robustness of \method{} to prompt design choices, we performed a prompt ablation study in which we systematically modified individual components of the prompt while keeping all other instructions unchanged.

The base condition is the prompt used throughout the main experiments (Appendix~\ref{appendix:nc_annotation_prompt}).
All variants were annotated in a separate run from the one reported in Table~\ref{tab:nc_llm_human_alignment}, together with the base prompt; owing to the non-determinism of GPU inference, the base results differ slightly from Table~\ref{tab:nc_llm_human_alignment} (by at most $0.002$ in Pearson correlation and $0.007$ in MSE).

We split the prompt into four conceptual components: \textcolor{orange}{(i)} a role instruction assigning the model as an expert, \textcolor{brown}{(ii)} the task description specifying the semantic property and relation to be rated, \textcolor{purple}{(iii)} the rating scale, and \textcolor{blue}{(iv)} output instructions constraining the response format.

\begin{quote}
\small
\textcolor{orange}{\textbf{[Role]}}
You are an expert in rating to what degree words are associated with semantic properties.
\medskip
\textcolor{brown}{\textbf{[Task]}}
To what degree is the word ``\{word\}'' associated with
\{property relation\} \{property explanation\}?
\medskip
\textcolor{purple}{\textbf{[Rating scale]}}
Provide only a single number for the rating of the word ``\{word\}'' on a scale from 0 (not at all associated or not applicable)
to 6 (very strongly associated).
\medskip
\textcolor{blue}{\textbf{[Output format]}}
Do not include any explanation or text, just the number.
Output:
\end{quote}

We evaluated a series of prompt variants by modifying one component at a time relative to the base prompt. 
The tested variants are summarized in Table~\ref{tab:prompt_ablation_variants}.

\begin{table}[h]
    \centering
    \caption{Prompt variants evaluated in the prompt ablation study. Each variant modifies a single component of the base prompt while leaving all other components unchanged.}
    \label{tab:prompt_ablation_variants}
    \small
    \begin{tabular}{llp{9cm}}
    \toprule
    \textbf{Component} & \textbf{Variant} & \textbf{Modification} \\
    \midrule
    Baseline & Base prompt &
    Prompt used throughout the main experiments (Appendix~\ref{appendix:nc_annotation_prompt}). \\
    \midrule
    Role & Remove expert role &
    Remove the instruction: ``You are an expert in rating to what degree words are associated with semantic properties.'' \\
    \midrule
    Task & Property name only &
    Provide only the property name instead of its relation and definition, e.g., ``associated with Motion?'' \\
    Task & Property name + definition &
    Provide the property name followed by its relation and definition, e.g., ``associated with Motion (showing a lot of visually observable movement)?'' \\
    Task & Relation: \textit{as} &
    Replace ``associated with'' by ``as'', e.g., ``To what degree is the word \{word\} as \{property relation\} \{property explanation\}?'' \\
    Task & Relation: \textit{characterized by} &
    Replace ``associated with'' by ``characterized by'', e.g., ``To what degree is the word \{word\} characterized by \{property relation\} \{property explanation\}?'' \\
    \midrule
    Scale & Binary (0--1) &
    Replace the 0--6 rating scale with a scale ranging from 0 (not at all associated or not applicable) to 1 (very strongly associated); ratings are multiplied by 6 for evaluation. \\
    Scale & Full scale descriptions &
    Keep the 0--6 scale but provide a textual description for every rating (e.g., 1 = very weakly associated, 2 = weakly associated, etc.). \\
    \midrule
    Output & Semantic reasoning &
    Add the instruction: ``Consider the meaning of the word itself when assigning the rating.'' \\
    \bottomrule
    \end{tabular}
\end{table}

Table~\ref{tab:prompt_ablations_mse_pearson} summarizes the prompt ablation results across the individual models and their ensemble.
Performance was largely robust to how the prompt is phrased and formatted: rewording the relation, changing the rating scale, or adding a reasoning instruction changed the correlation of the ensemble with human ratings by at most $0.015$, and the original prompt achieved the best ensemble performance.
The largest drops occurred when information was removed from the prompt, namely when the property definition was replaced by the property name alone and, to a lesser extent, when the expert role was removed.
This is expected, as the property definitions specify what each semantic dimension means; as in the human rating protocol of \citet{binder2016toward}, providing them is part of the annotation procedure rather than a free design choice.

\begin{table}[t]
    \centering
    \begin{tabular}{llcc}
    \toprule
    \textbf{Model} & \textbf{Ablation} & \textbf{MSE} & \textbf{Pearson Corr.} \\
    \midrule
    \multirow{9}{*}{llama3.1-8b-instruct}
        & Base & 1.693 & 0.674 \\
        & No role & 2.615 & 0.580 \\
        & Property name only & 2.684 & 0.498 \\
        & Property name + definition & 1.867 & 0.639 \\
        & Relation ``as'' & 1.717 & 0.663 \\
        & Relation ``characterized by'' & 1.700 & 0.659 \\
        & Scale 0--1 & 1.980 & 0.662 \\
        & Long scale (0--6) & 1.942 & 0.662 \\
        & Semantic reasoning & 1.610 & 0.660 \\
    \midrule
    \multirow{9}{*}{mistral-7b-instruct}
        & Base & 2.163 & 0.616 \\
        & No role & 2.405 & 0.576 \\
        & Property name only & 2.801 & 0.445 \\
        & Property name + definition & 2.142 & 0.599 \\
        & Relation ``as'' & 2.349 & 0.557 \\
        & Relation ``characterized by'' & 2.386 & 0.581 \\
        & Scale 0--1 & 2.380 & 0.599 \\
        & Long scale (0--6) & 2.206 & 0.607 \\
        & Semantic reasoning & 2.378 & 0.589 \\
    \midrule
    \multirow{9}{*}{qwen2.5-7b-instruct}
        & Base & 1.324 & 0.693 \\
        & No role & 2.565 & 0.615 \\
        & Property name only & 2.046 & 0.507 \\
        & Property name + definition & 1.403 & 0.673 \\
        & Relation ``as'' & 1.551 & 0.643 \\
        & Relation ``characterized by'' & 1.417 & 0.664 \\
        & Scale 0--1 & 1.327 & 0.682 \\
        & Long scale (0--6) & 1.421 & 0.685 \\
        & Semantic reasoning & 1.325 & 0.692 \\
    \midrule
    \multirow{9}{*}{llama3.1-70b-instruct}
        & Base & 1.530 & 0.742 \\
        & No role & 2.033 & 0.656 \\
        & Property name only & 2.628 & 0.587 \\
        & Property name + definition & 1.517 & 0.751 \\
        & Relation ``as'' & 1.508 & 0.747 \\
        & Relation ``characterized by'' & 1.539 & 0.741 \\
        & Scale 0--1 & 1.489 & 0.735 \\
        & Long scale (0--6) & 1.722 & 0.734 \\
        & Semantic reasoning & 1.578 & 0.737 \\
    \midrule
    \multirow{9}{*}{Ensemble}
        & Base & \textbf{1.076} & \textbf{0.767} \\
        & No role & 1.496 & 0.698 \\
        & Property name only & 1.785 & 0.599 \\
        & Property name + definition & 1.147 & 0.756 \\
        & Relation ``as'' & 1.084 & 0.752 \\
        & Relation ``characterized by'' & 1.150 & 0.757 \\
        & Scale 0--1 & 1.115 & 0.756 \\
        & Long scale (0--6) & 1.140 & 0.765 \\
        & Semantic reasoning & 1.086 & 0.765 \\
    \bottomrule
    \end{tabular}
    \caption{
    Prompt ablation results across individual models and their ensemble on the $535$ words of \citet{binder2016toward}.
    Pearson correlation and MSE are computed per word across the $65$ properties and averaged over words.
    The base prompt is the main prompt; its values differ slightly from Table~\ref{tab:nc_llm_human_alignment} because of non-deterministic GPU inference across runs.
    }
    \label{tab:prompt_ablations_mse_pearson}
\end{table}

\clearpage
\subsubsection{Contextual semantic annotation}
\label{appendix:contextual_annotation_prompt}

In the contextual setting, \method{} follows the same procedure as in the out-of-context case, using LLM-generated annotations independently for each concept-property pair.
The key difference is that semantic scores are now conditioned on the sentence in which the concept appears, enabling context-dependent representations.

The prompt is modified as follows:
\begin{quote}
\texttt{You are an expert in rating to what degree words are associated with semantic properties. \\}
\textbf{Given the context ``\{contextual\_sentence\}'',}
\texttt{to what degree is the word ``\{word\}'' associated with \{property relation\} \{property explanation\}? \\
Provide only a single number for the rating of the word ``\{word\}'' on a scale from 0 (not at all associated or not applicable) to 6 (very strongly associated). \\
Do not include any explanation or text, just the number. \\
Output:
}
\end{quote}

This formulation allows the model to adjust semantic feature values based on how the meaning of the target word is modulated by context.

An example prompt for the concept word ``bus'', the property ``motion'', and the contextual sentence of ``The bus was parked at the station'' is shown below:
\begin{quote}
\small{\textit{You are an expert in rating to what degree words are associated with semantic properties. \\
Given the context ``The bus was parked at the station'', to what degree is the word ``bus'' associated with showing a lot of visually observable movement? \\
Provide only a single number for the rating of the word ``bus'' on a scale from 0 (not at all associated or not applicable) to 6 (very strongly associated). \\
Do not include any explanation or text, just the number.\\
Output:
}}
\end{quote}

\paragraph{Examples of context window sizes.} Table~\ref{tab:context_window_examples} illustrates the different context window sizes used for contextual semantic annotation.
For each target word, the context consists of the $k$ tokens preceding and including the target, with larger values of $k$ providing progressively broader linguistic context.

\begin{table}[h]
\centering
\small
\caption{Examples of the context provided to SCRIBE for different context
window sizes. The target word is shown in bold.}
\label{tab:context_window_examples}
\begin{tabularx}{\linewidth}{c l X}
\toprule
\textbf{Example} & \textbf{Context size} & \textbf{Context} \\
\midrule

\multirow{4}{*}{1}
& $k=3$  & of those \textbf{people} \\
& $k=5$  & I'm one of those \textbf{people} \\
& $k=10$ & raised with a religion and I'm one of those \textbf{people} \\
& $k=50$ & so a lot of people are raised with a religion and I'm one of those \textbf{people} \\

\addlinespace

\multirow{4}{*}{2}
& $k=3$  & the word \textbf{pretty} \\
& $k=5$  & extension of the word \textbf{pretty} \\
& $k=10$ & word beautiful was just an extension of the word \textbf{pretty} \\
& $k=50$ & when I was younger the word beautiful was just an extension of the word \textbf{pretty} \\

\addlinespace

\multirow{4}{*}{3}
& $k=3$  & date and \textbf{i} \\
& $k=5$  & a christian date and \textbf{i} \\
& $k=10$ & time i was going on a christian date and \textbf{i} \\
& $k=50$ & it was the first time i was going on a christian date and \textbf{i} \\

\addlinespace

\multirow{4}{*}{4}
& $k=3$  & learn a \textbf{lesson} \\
& $k=5$  & romance not learn a \textbf{lesson} \\
& $k=10$ & go to morocco to escape romance not learn a \textbf{lesson} \\
& $k=50$ & my plan was to go to morocco to escape romance not learn a \textbf{lesson} \\

\bottomrule
\end{tabularx}
\end{table}

\clearpage
\subsection{Synthetic dataset for controlled contextual evaluation}
\label{appendix:synthetic_dataset}


To evaluate whether contextual information induces systematic changes in semantic representations, we construct a synthetic dataset designed to isolate the effect of context on individual concept–property associations. 
A key requirement for this evaluation is to control for the baseline strength of each association, as it constrains how much it can change under contextual manipulation. 
We therefore first select a subset of concept–property pairs that span the full range of baseline association strengths (Section~\ref{appendix:synthetic_dataset_select_concepts_and_properties}), and only then generate contexts that induce opposite semantic effects: one that aims to decrease the association (``low'' condition) and one that aims to increase it (``high'' condition) (Section~\ref{appendix:synthetic_dataset_context_generation}). 
This design enables controlled testing of whether semantic features respond predictably to targeted contextual manipulations.

\subsubsection{Concept and property selection}
\label{appendix:synthetic_dataset_select_concepts_and_properties}

A key challenge in evaluating contextual effects is that the baseline out-of-context strength of a concept–property association may constrain how much it can change. 
For example, pairs with very high initial ratings have limited room for further increase, while pairs with very low ratings may be difficult to decrease.
To control for this effect, we leverage the $535$ concepts from the \citet{binder2016toward} dataset, which provides human ratings for each concept–property pair and allows us to explicitly characterize baseline out-of-context association strength.

\begin{figure}[h]
    \centering
    \begin{subfigure}{0.48\linewidth}
        \centering
        \includegraphics[width=\linewidth]{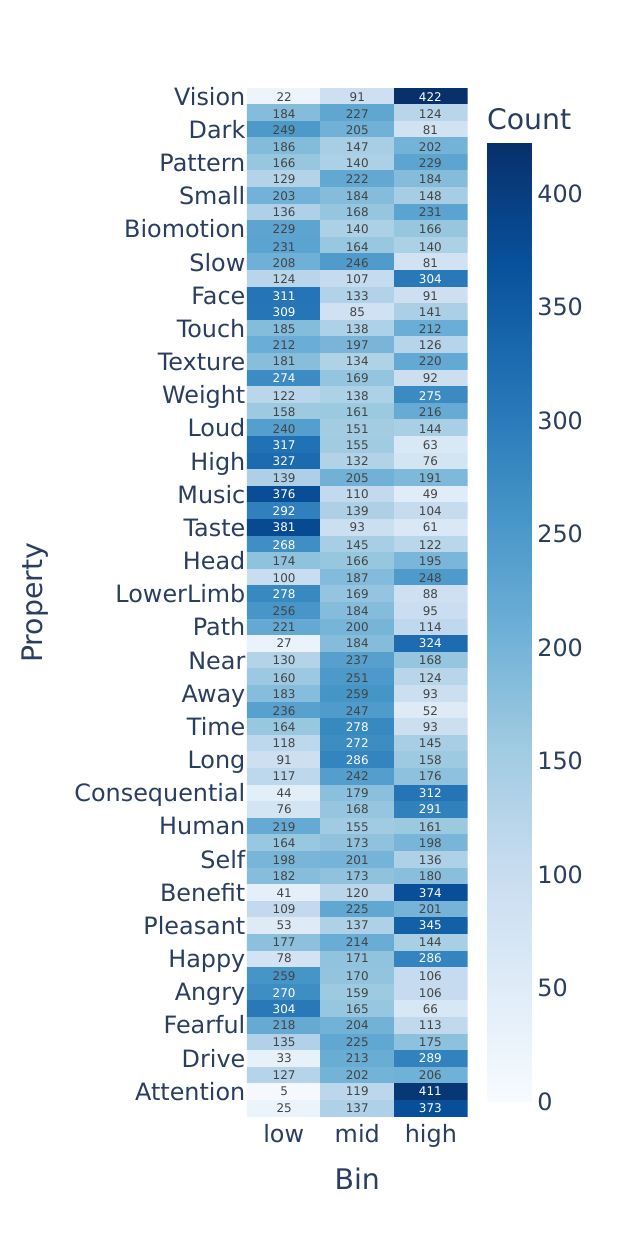}
        \caption{Quantiles thresholds.}
    \end{subfigure}
    \begin{subfigure}{0.48\linewidth}
        \centering
        \includegraphics[width=\linewidth]{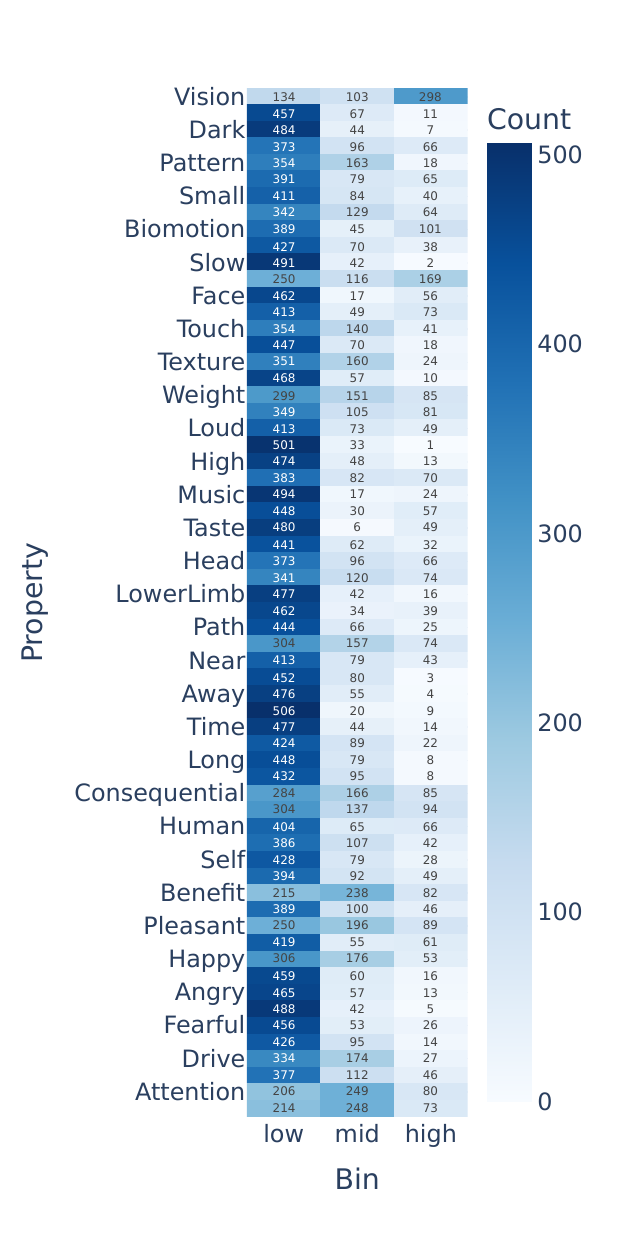}
        \caption{Manual thresholds.} 
    \end{subfigure}
\caption{
    Distribution of (concept, property) pairs across discretized baseline bins under two binning strategies. 
    \textbf{(a)} Quantile-based thresholds ($0.34$, $1.5$) yield more uniform distributions across bins; however, the threshold for the high bin is low relative to the full rating scale ($0$–$6$), leading to less interpretable groupings. 
    \textbf{(b)} Manual thresholds aligned with the original rating scale ($[0,2)$, $[2,4)$, $[4,6]$) produce more interpretable assignments of low, mid, and high association levels across properties. 
    These observations motivate the use of fixed thresholds in our synthetic dataset construction.
    }
    \label{fig:apendix_concept_property_bin_synthetic_data}
\end{figure}

We discretize these ratings into three levels (\emph{low}, \emph{mid}, and \emph{high}) using thresholds aligned with the original rating scale:
\[
\text{low}: [0,2), \quad \text{mid}: [2,4), \quad \text{high}: [4,6].
\]
This yields a categorical matrix that assigns each concept–property pair to a baseline association bin (see Figure~\ref{fig:apendix_concept_property_bin_synthetic_data} for bin coverage).

To ensure controlled comparisons across these levels, we construct a balanced subset of concepts. 
Not all properties exhibit sufficient variability across concepts (Section~\ref{appendix:binder_features}). 
We therefore retain only properties for which each bin contains at least $8$ concepts, resulting in $57$ properties after filtering, and excluding \textit{Low, Slow, Toward, Away, Disgusted, Complexity, Taste, Dark}.
For each retained property, we then select a subset of concepts that spans the full range of baseline associations using a greedy selection procedure:
\begin{enumerate}
    \item Initialize coverage counts for each (property, bin) cell.
    \item At each step, consider all remaining candidate concepts.
    \item For each candidate, compute how many under-covered (property, bin) cells it would help fill.
    \item Select the concept that maximizes this coverage gain.
    \item Update coverage counts and repeat until all cells reach the target.
\end{enumerate}

At convergence, we then select $8$ concepts per property bin (low, mid, high), yielding $24$ concepts per property. 
This balanced design ensures that contextual effects are evaluated under comparable baseline conditions, rather than being driven by saturation or floor effects.
Consistent with this concern, human ratings for low-baseline pairs were not further decreased by low-target contexts (Appendix~\ref{appendix:results_human_survey}).

The final set of $144$ concepts is:
\texttt{deceit, snub, gasp, infinity, truth, reality, advice, era, grew, evening, number, fed, terrorist, quantity, handshake, angry, jaw, street, dinner, hospital, baseball, bonfire, foot, accordion, animosity, crossed, gunshot, dropped, cough, ricochet, raspberry, egg, morning, dime, tulip, festival, apricot, sun, cloud, summer, shiny, explosion, dandelion, lightning, election, empty, ate, chime, banjo, carnival, fireworks, bus, beach, bagpipe, avalanche, toe, drew, cash, coffee, dolphin, belief, lived, protest, fiddle, leg, volcano, elm, snake, alligator, clang, girl, mouth, bicycle, sick, loud, built, apartment, ambulance, battle, college, audience, school, riot, jury, couple, thunder, pineapple, drank, finger, banana, bell, bee, cellphone, harmonica, beer, long, belch, delivered, activist, victim, man, matinee, parade, army, sum, council, lip, arm, walked, marched, approached, ran, hailstorm, expensive, airport, bugle, celebrated, old, tired, shouted, doctor, dead, left, hiked, friendly, insult, car, treaty, wonder, problem, funeral, injured, torment, semester, year, listened (to), screech, scream, excuse, aggressive, restaurant, worked, day, intellect}.

\subsubsection{Contextual sentence generation}
\label{appendix:synthetic_dataset_context_generation}

For each selected concept-property pair in the synthetic dataset, we then use \texttt{llama3.1-8b-instruct} to generate two sentences: one intended to decrease the target property score (low condition) and one intended to increase it (high condition).
These sentences are designed to manipulate the semantic interpretation of the target concept in a controlled manner.

\paragraph{Prompt template to generate the contextual sentences.}
The prompt template used to construct the synthetic dataset is shown below:

\begin{quote}
\texttt{You are a sentence generation assistant. Your task is to write ONE simple and short sentence using the target word ``\{concept\}''. The sentence should evoke a **\{level\}** value for the semantic property ``\{property name\}'', defined as \{property relation\} \{property explanation\}. \\
Constraints: \\
- Use the word ``\{concept\}'' exactly as given in the sentence. \\
- Do \textbf{not} use the property name ``\{property name\}'' directly in the sentence. \\
- Do \textbf{not} include explanations or extra formatting, only the sentence. \\
Output format: \\
Return the sentence in the following format:
\textless ONE sentence with ``\{concept\}'' \textgreater.\\
No additional text, keys or punctuation.\\
Context: \\
The word ``\{concept\}'' belongs to the super category ``\{super category concept\}'' and category ``\{category concept\}'' \\
Output:
}
\end{quote}

An example to generate a contextual sentence for the concept word ``bus'', the property ``motion'', and the target level ``low'' is shown below:
\begin{quote}
\small{\textit{
You are a sentence generation assistant.\\
Your task is to write ONE simple and short sentence using the target word ``bus''.\\
The sentence should evoke a ``low'' value for the semantic property ``Motion'', defined as showing a lot of visually observable movement. \\
Constraints:\\
- Use the word ``bus'' exactly as given in the sentence. \\
- Do not use the property name ``Motion'' as given directly in the sentence.\\
- Do not include explanations or extra formatting, just the sentence.\\
Output format:\\
Return the sentence in the following format:
\textless ONE sentence with ``bus'' \textgreater.\\
No additional text, keys or punctuation.\\
Context:\\
The word ``bus'' belongs to the super category ``artifact'' and category ``vehicle''.\\
Output:
}}
\end{quote}





\paragraph{Example of contextual sentences generated.}
We manually inspect a subset of generated examples to ensure quality and coherence.
Some examples are presented in Table~\ref{tab:appendix_contextual_examples}.

\begin{table}[h]
    \centering
    \small
    \caption{
    Examples of controlled contextual manipulations for selected property-concept pairs. 
    For each pair, the low-context sentence is designed to reduce the target semantic property, while the high-context sentence increases it, demonstrating how context systematically modulates semantic feature annotations.
    }
    \begin{tabular}{llp{5.5cm}p{5.5cm}}
    \toprule
    \textbf{Property} & \textbf{Concept} & \textbf{Low-context sentence} & \textbf{High-context sentence} \\
    \midrule
    Angry & thunder 
    & The thunderstorm was a beautiful display of nature.
    & The loud thunder made me furious. \\
    
    Motion & bus 
    & The bus was parked at the station. 
    & The bus sped down the highway. \\
    
    Social & dinner 
    & I ate dinner alone at home. 
    & We had dinner together with friends at the restaurant. \\
    
    Pleasant & fireworks 
    & The loud fireworks startled the baby. 
    & Watching fireworks on the 4th of July was a joyful experience. \\

    Fast & bus
    & The bus moved slowly through traffic.
    & The bus sped down the highway. \\

    Motion & activist
    & The activist sat quietly in the corner.
    & The activist marched through the streets. \\

    Number & leg
    & She has a broken leg.
    & The dog has four legs. \\

    Small & egg
    & The ostrich egg is enormous.
    & The tiny egg was barely visible to the naked eye. \\
    
    \bottomrule
    \end{tabular}
    \label{tab:appendix_contextual_examples}
\end{table}

\clearpage
\subsection{Brain encoding setup}
\label{appendix:brain_encoding}

\subsubsection{fMRI dataset}
\label{appendix:fmri_data}

We use a subset of the fMRI dataset originally described and made publicly available in previous work~\citep{lebel2023natural}.
The subset used consists of $11$ stories and $8$ participants, all of whom listened to all the stories.
The speech stimuli consisted of 10 to 15 minute stories taken from The \emph{Moth Radio Hour} podcast and used previously~\citep{moussa2024improving, antonello2023scaling, oota2024speech}.
The selected stories cover a wide range of topics and are highly engaging~\citep{oota2024speech}.
The model validation dataset consisted of one 10 min story.
In this work, we include 7 of the 8 subjects, excluding subject 4 due to incomplete data (a missing story).
To enable comparison with QAEmb~\citep{singh2025evaluating}, an interpretability representation used as baseline, we further exclude one story (``myfirstdaywiththeyankees'') that is not available in its feature set, resulting in a total of 9 stories used for training the encoding models.

\subsubsection{ROIs Details}
\label{appendix:rois_details}

The human cerebral cortex multi-modal parcellation (Glasser atlas) defines 180 regions of interest (ROIs) per hemisphere~\citep{glasser2016multi}.
Following prior work~\citep{moussa2024improving, oota2024speech}, we define language regions based on a subset of these ROIs, as listed in Table~\ref{tab:brain_regions}.

\renewcommand{\arraystretch}{1.5}  

\begin{table}[h]
    \centering
    \caption{Brain regions and corresponding ROI labels.}
    \label{tab:brain_regions}
    \begin{tabular}{p{4cm} p{7cm}}  
    \toprule
    \textbf{Region} & \textbf{Labels} \\
    \midrule
    Angular gyrus (AG) & PFm, PGs, PGi, TPOJ2, TPOJ3 \\
    Lateral temporal cortex (LTC) & STSda, STSva, STGa, TE1a, TE2a, TGv, TGd, A5, \newline
                              STSdp, STSvp, PSL, STV, TPOJ1 \\
    Inferior frontal gyrus (IFG) & 44, 45, IFJa, IFSp \\
    Middle frontal gyrus (MFG) & 55b \\
    Primary auditory cortex (A1) & A1 \\
    Early auditory regions & A1, PBelt, MBelt, LBelt, RI, A4 \\
    \bottomrule
    \end{tabular}
    \vspace{0.5em}
\end{table}

\subsubsection{Estimated noise ceiling}
\label{appendix:noise_ceiling}

Noise in fMRI data can impair evaluations, so it is important to estimate the ``noise ceiling'' of each voxel in the fMRI recordings.
The voxel-wise noise ceiling is estimated for all participants based on the preferred method by the fMRI dataset paper~\citep{lebel2023natural}, which leverages within-participant repetitions of the same story to compute the maximum explainable variance for each voxel.
This noise ceiling value estimates the amount of explainable variance in the brain signal, ranging from $0$ to $1$.
We use this estimated noise ceiling to filter noisy voxels during brain alignment evaluation.
We use a filtration threshold of $0.4$, in line with the findings of \citet{antonello2023scaling}.
After filtering voxels with low noise ceiling, there remain $30,000$ to $50,000$ voxels per participant.

\clearpage
\section{Results}
\label{appendix:results}

\subsection{\method~generates semantic features that align with human judgments}
\label{appendix:results_nc}

\subsubsection{Permutation Test}
\label{appendix:results_permutation_test}

\begin{figure}[h]
    \centering
    \begin{subfigure}{0.3\linewidth}
        \centering
        \includegraphics[width=\linewidth]{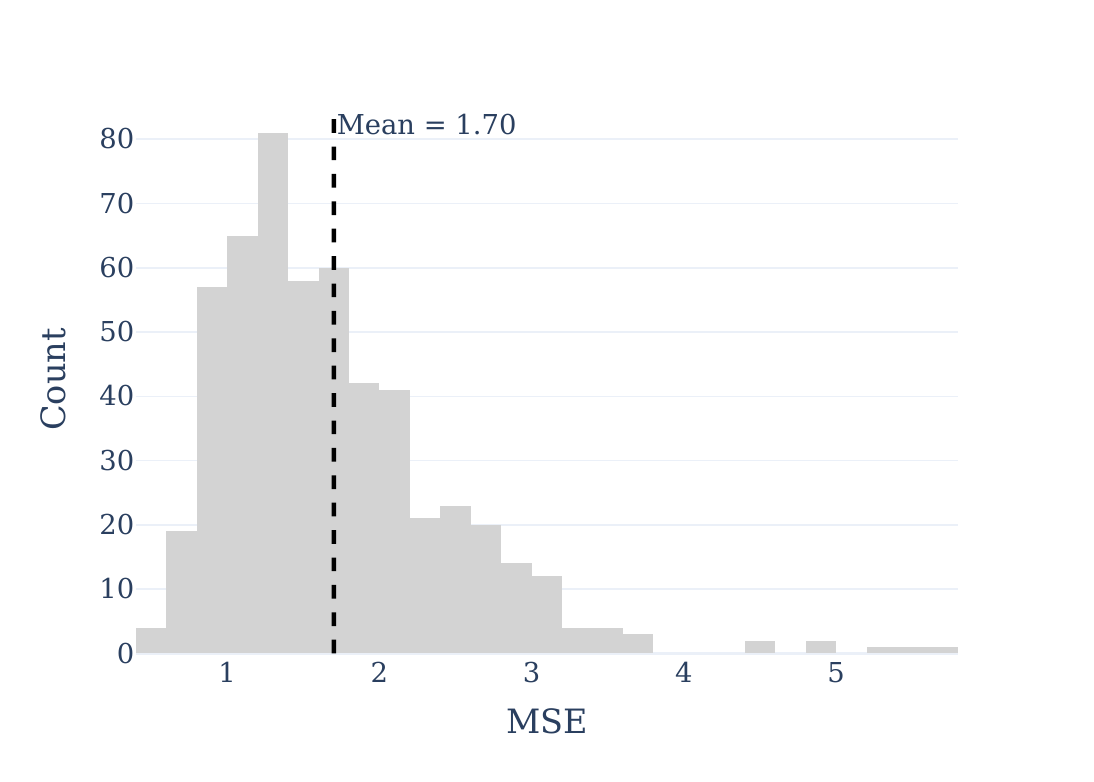}
        \caption{MSE real (llama3.1-8b).}
    \end{subfigure}
    \begin{subfigure}{0.3\linewidth}
        \centering
        \includegraphics[width=\linewidth]{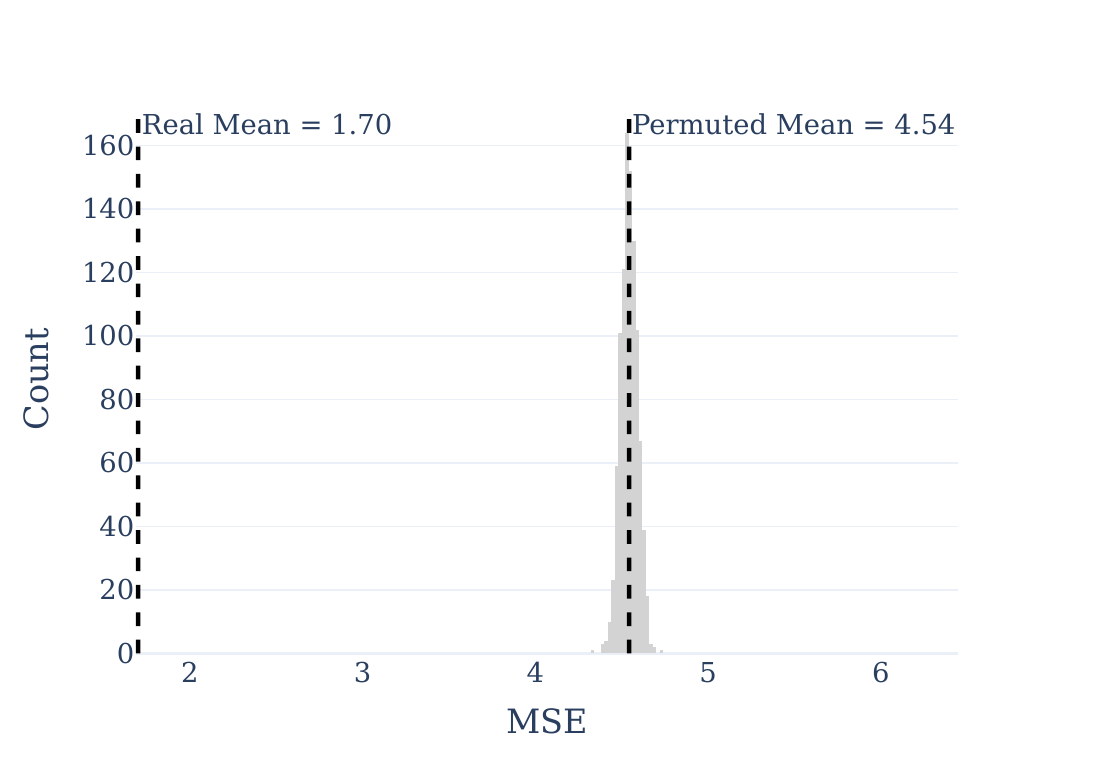}
        \caption{MSE perm. (llama3.1-8b).}
    \end{subfigure}

    \begin{subfigure}{0.3\linewidth}
        \centering
        \includegraphics[width=\linewidth]{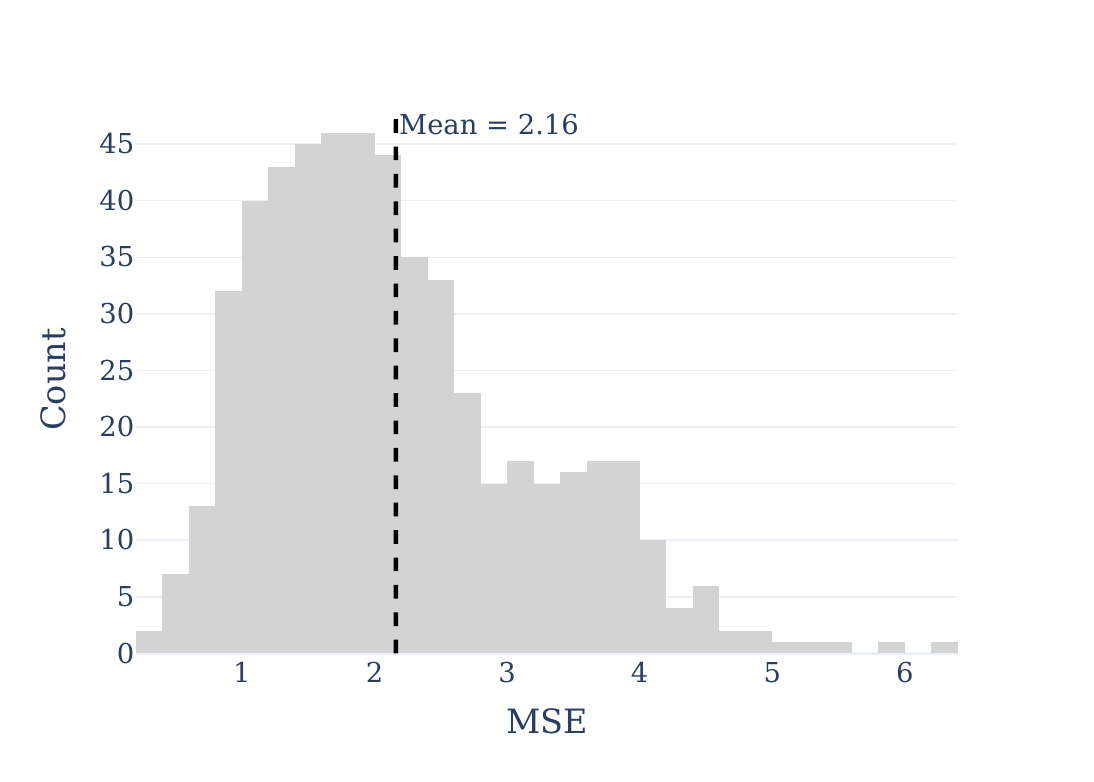}
        \caption{MSE real (mistral-7b).}
    \end{subfigure}
    \begin{subfigure}{0.3\linewidth}
        \centering
        \includegraphics[width=\linewidth]{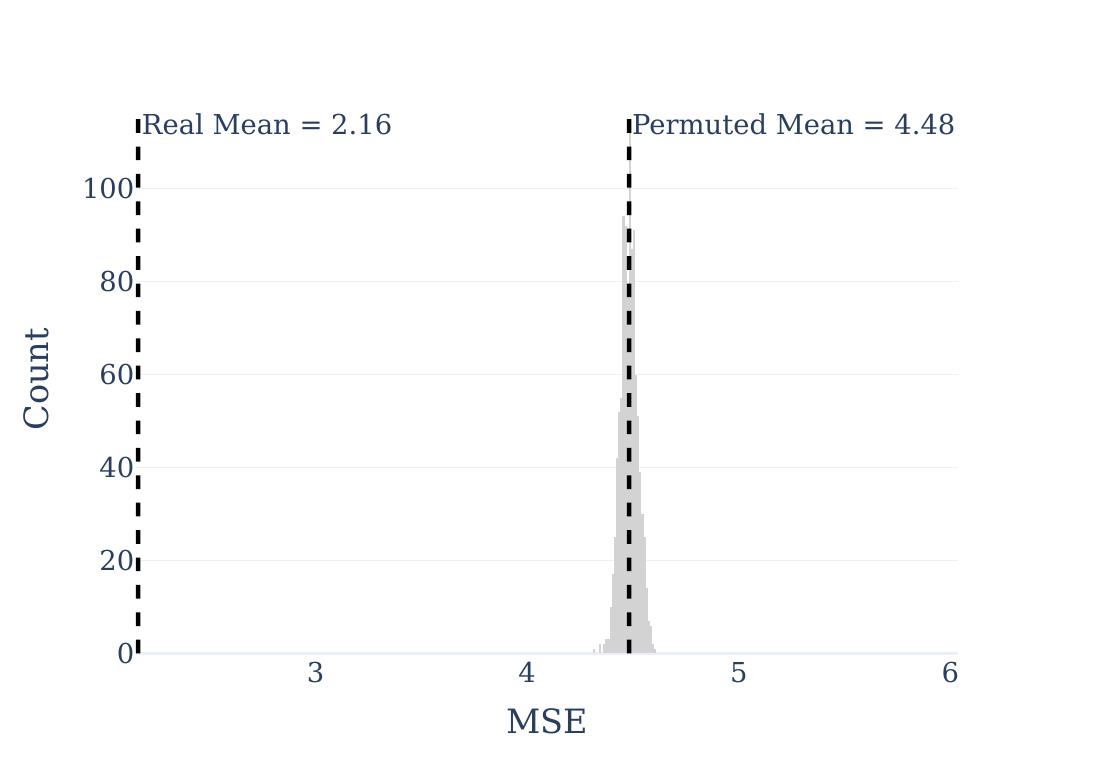}
        \caption{MSE perm. (mistral-7b).}
    \end{subfigure}

    \begin{subfigure}{0.3\linewidth}
        \centering
        \includegraphics[width=\linewidth]{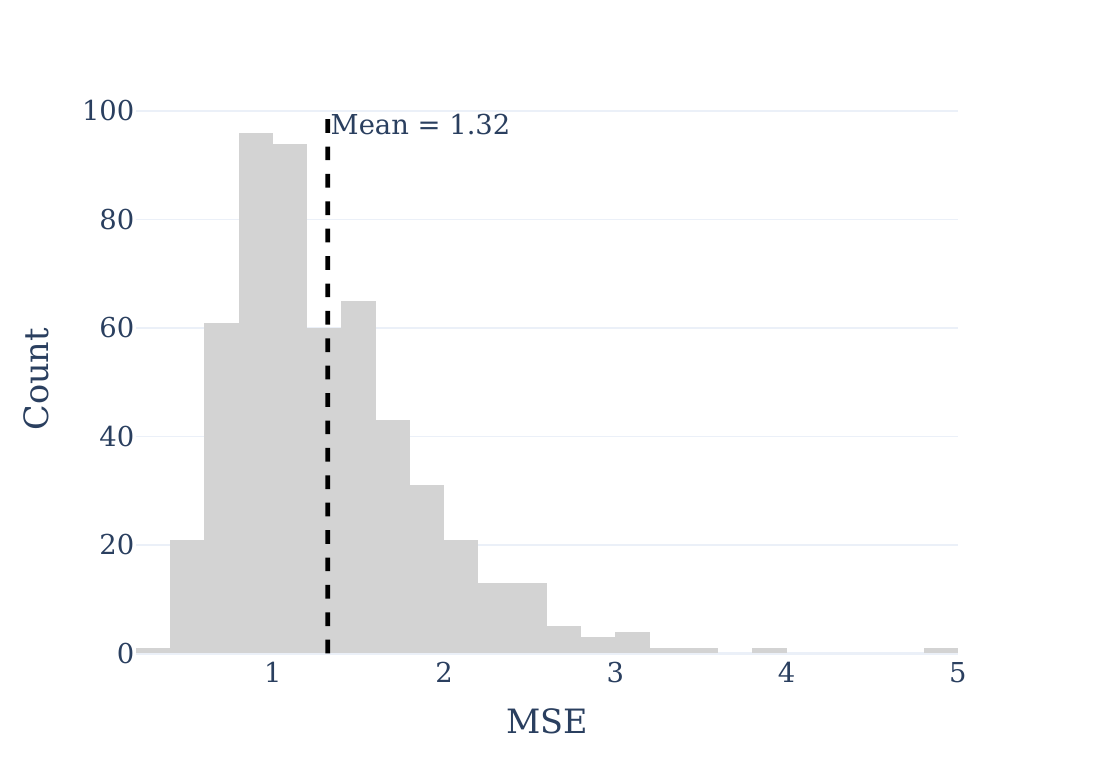}
        \caption{MSE real (qwen2.5-7b).}
    \end{subfigure}
    \begin{subfigure}{0.3\linewidth}
        \centering
        \includegraphics[width=\linewidth]{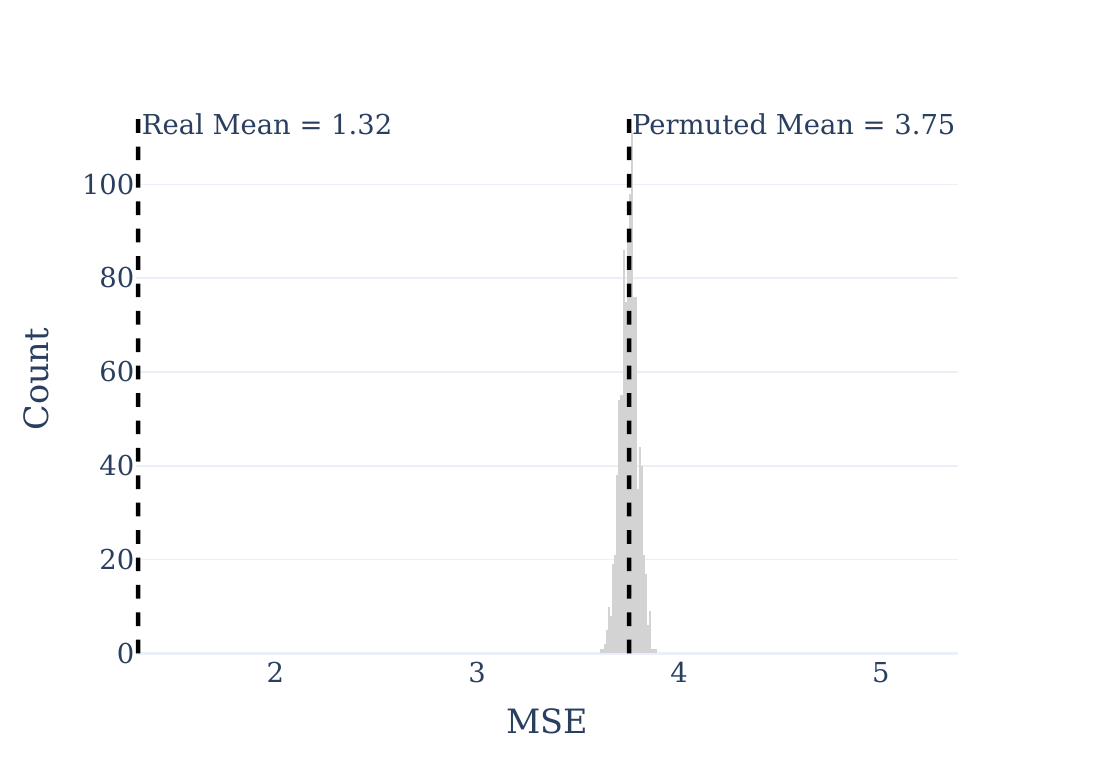}
        \caption{MSE perm. (qwen2.5-7b).}
    \end{subfigure}

    \begin{subfigure}{0.3\linewidth}
        \centering
        \includegraphics[width=\linewidth]{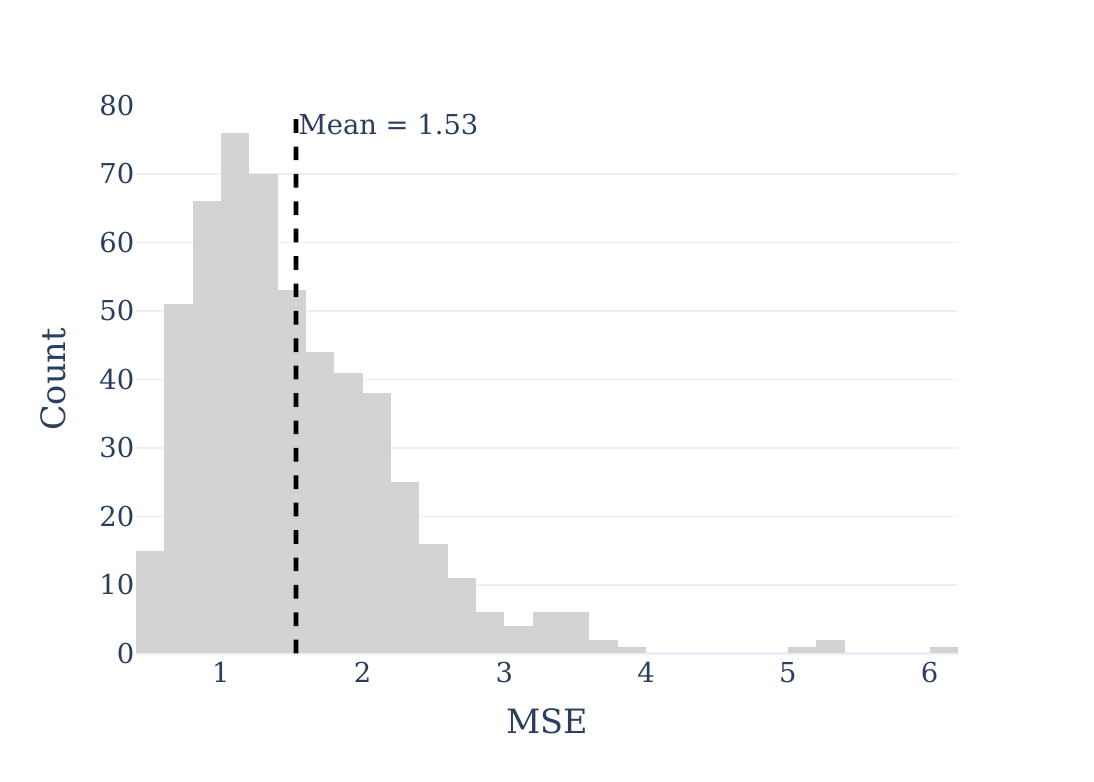}
        \caption{MSE real (llama3.1-70b).}
    \end{subfigure}
    \begin{subfigure}{0.3\linewidth}
        \centering
        \includegraphics[width=\linewidth]{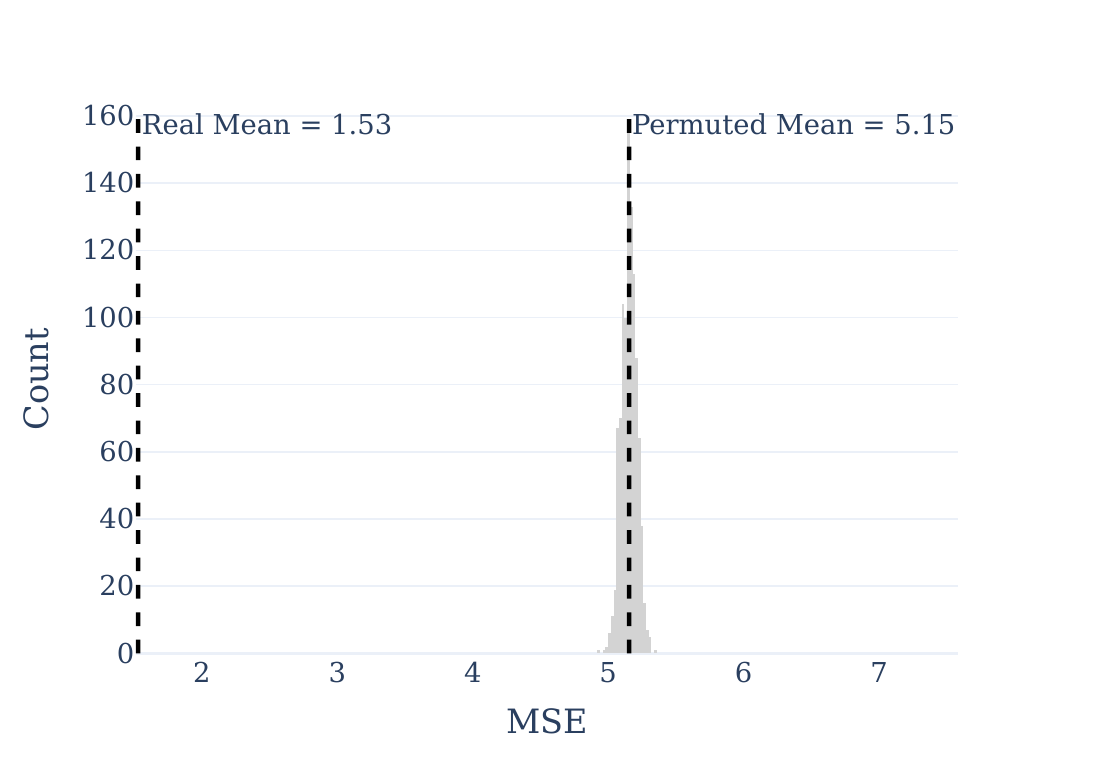}
        \caption{MSE perm. (llama3.1-70b).}
    \end{subfigure}
    
    \begin{subfigure}{0.3\linewidth}
        \centering
        \includegraphics[width=\linewidth]{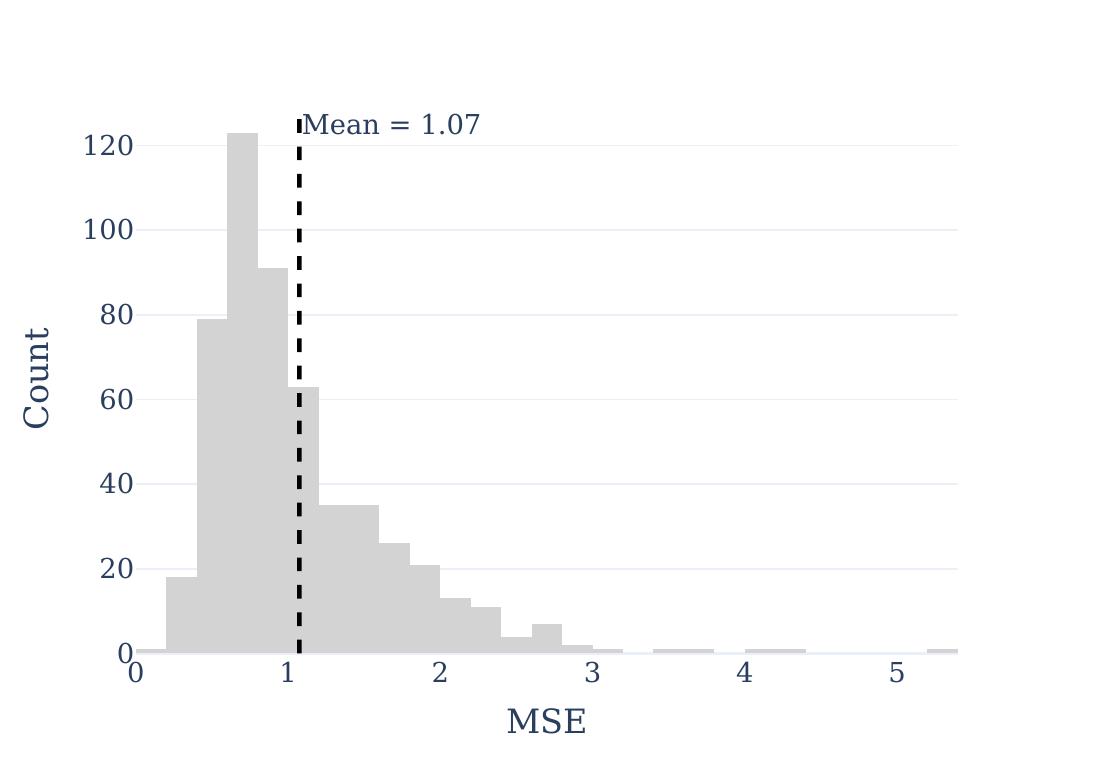}
        \caption{MSE real (ensemble).}
    \end{subfigure}
    \begin{subfigure}{0.3\linewidth}
        \centering
        \includegraphics[width=\linewidth]{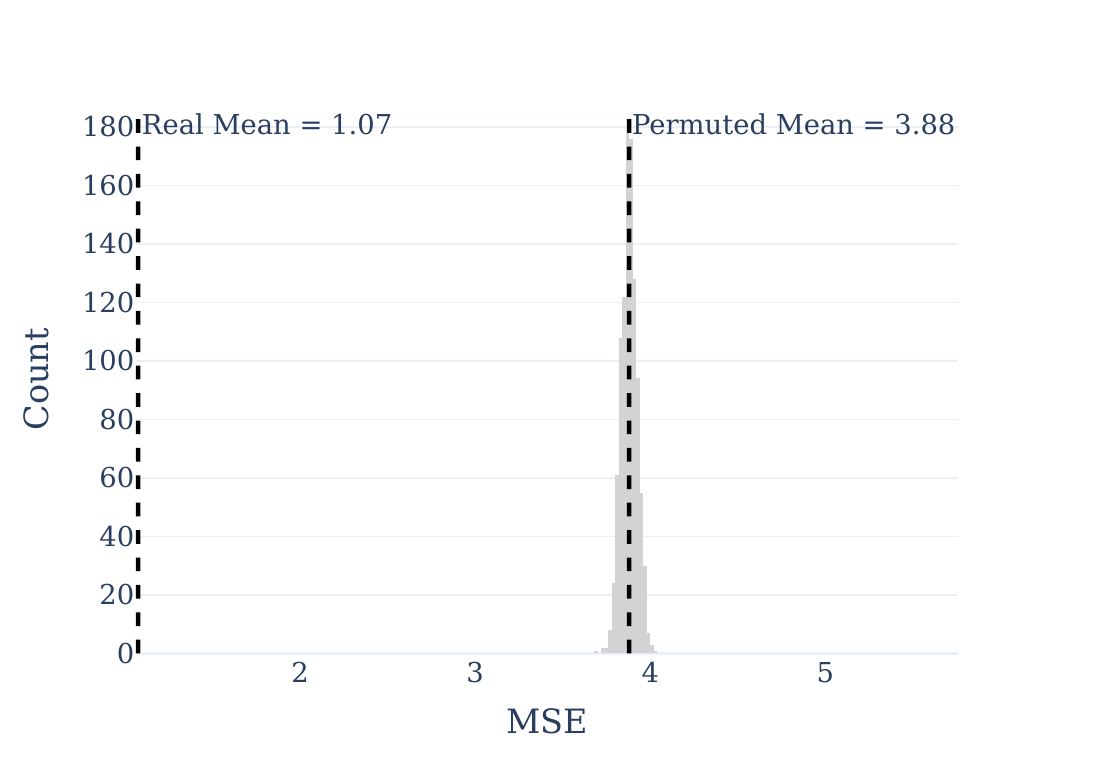}
        \caption{MSE perm. (ensemble).}
    \end{subfigure}
\caption{
    Distribution of mean squared error (MSE) for out-of-context semantic annotation across models. 
    For each model, we show the distribution of the MSE values computed on the true concept–property assignments (left) and under a permutation baseline obtained by randomly shuffling concept–property pairs (right). 
    Across all models, the true MSE distribution is substantially lower than the permutation baseline, indicating that LLM-based annotations capture meaningful semantic structure beyond chance. 
    Results are shown for individual models and the ensemble.
    }
    \label{fig:appendix_permutation_test_mse_distributions}
\end{figure}

\begin{figure}[h]
    \centering
    \begin{subfigure}{0.3\linewidth}
        \centering
        \includegraphics[width=\linewidth]{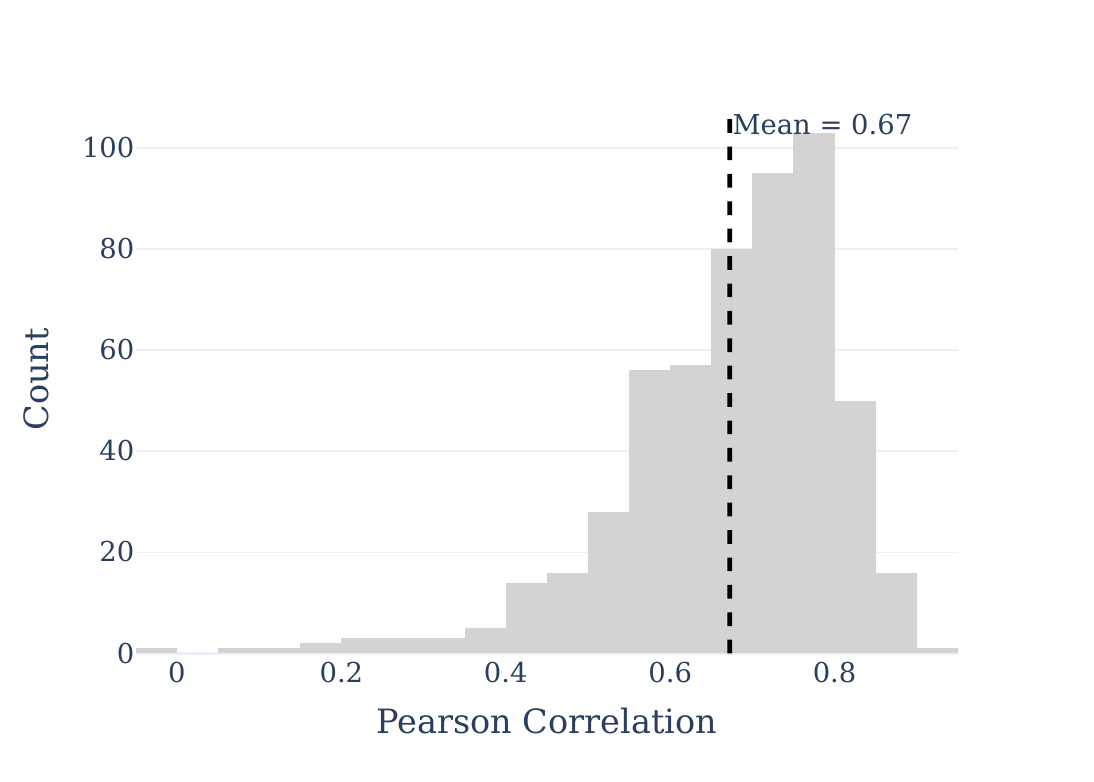}
        \caption{Pearson real (llama3.1-8b).}
    \end{subfigure}
    \begin{subfigure}{0.3\linewidth}
        \centering
        \includegraphics[width=\linewidth]{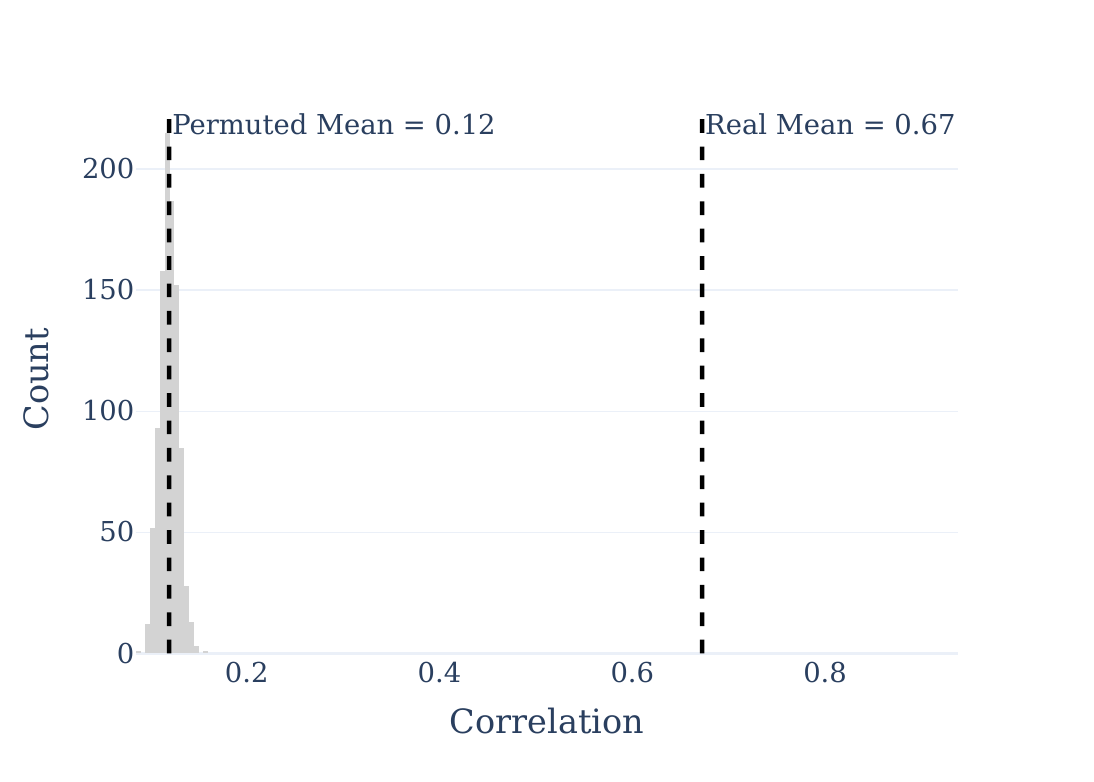}
        \caption{Pearson perm. (llama3.1-8b).}
    \end{subfigure}

    \begin{subfigure}{0.3\linewidth}
        \centering
        \includegraphics[width=\linewidth]{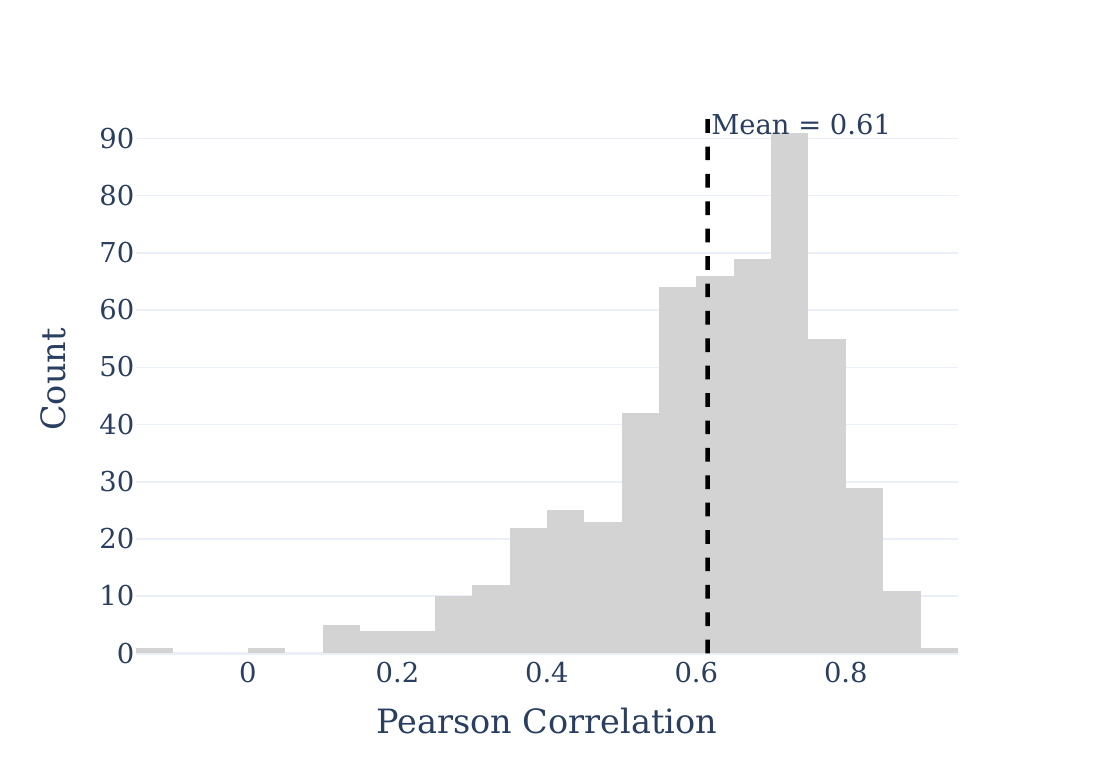}
        \caption{Pearson real (mistral-7b).}
    \end{subfigure}
    \begin{subfigure}{0.3\linewidth}
        \centering
        \includegraphics[width=\linewidth]{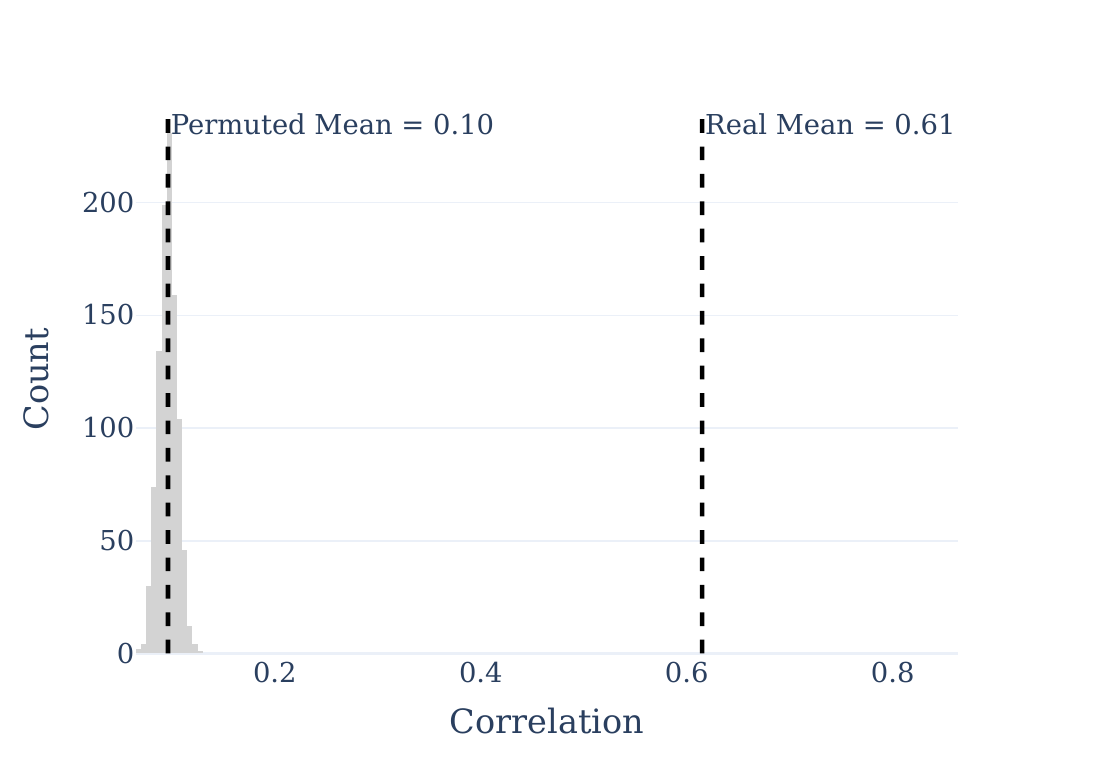}
        \caption{Pearson perm. (mistral-7b).}
    \end{subfigure}

    \begin{subfigure}{0.3\linewidth}
        \centering
        \includegraphics[width=\linewidth]{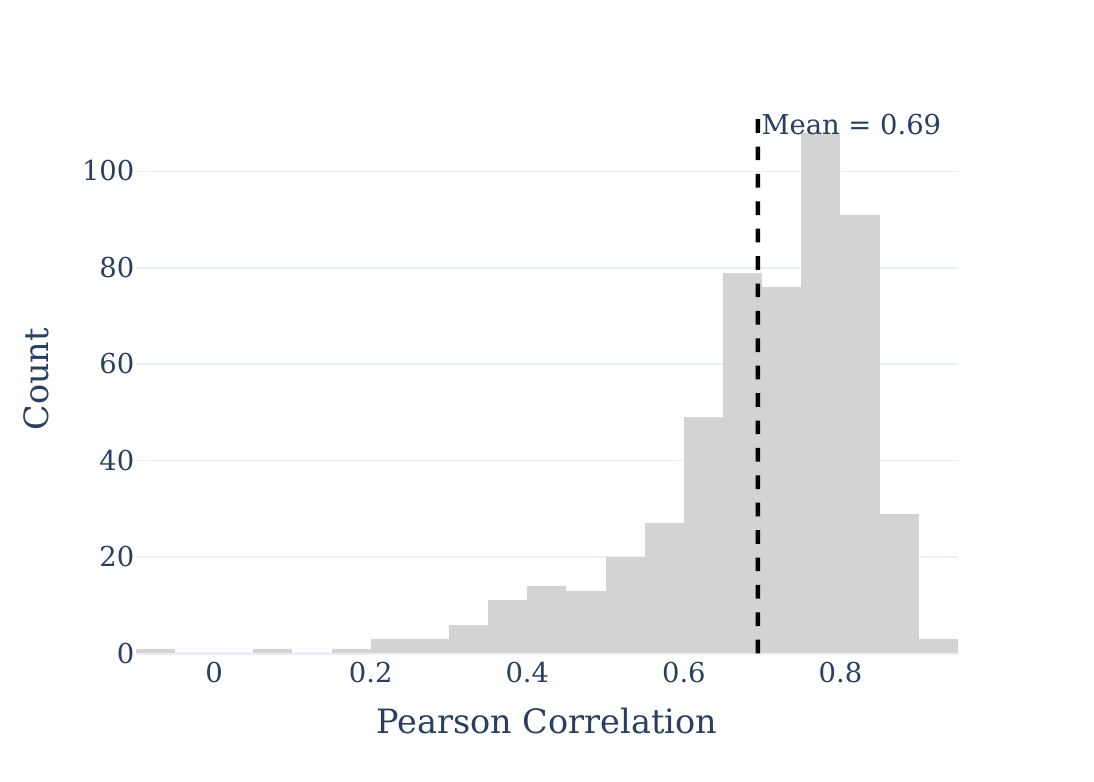}
        \caption{Pearson real (qwen2.5-7b).}
    \end{subfigure}
    \begin{subfigure}{0.3\linewidth}
        \centering
        \includegraphics[width=\linewidth]{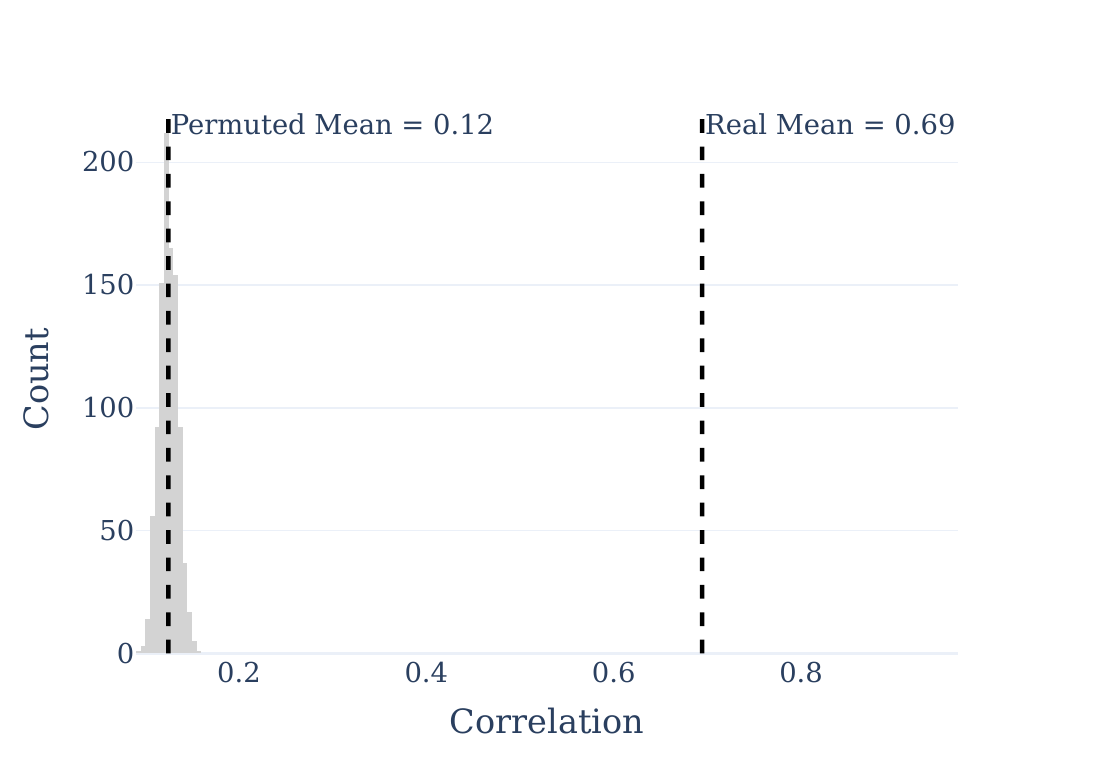}
        \caption{Pearson perm. (qwen2.5-7b).}
    \end{subfigure}

    \begin{subfigure}{0.3\linewidth}
        \centering
        \includegraphics[width=\linewidth]{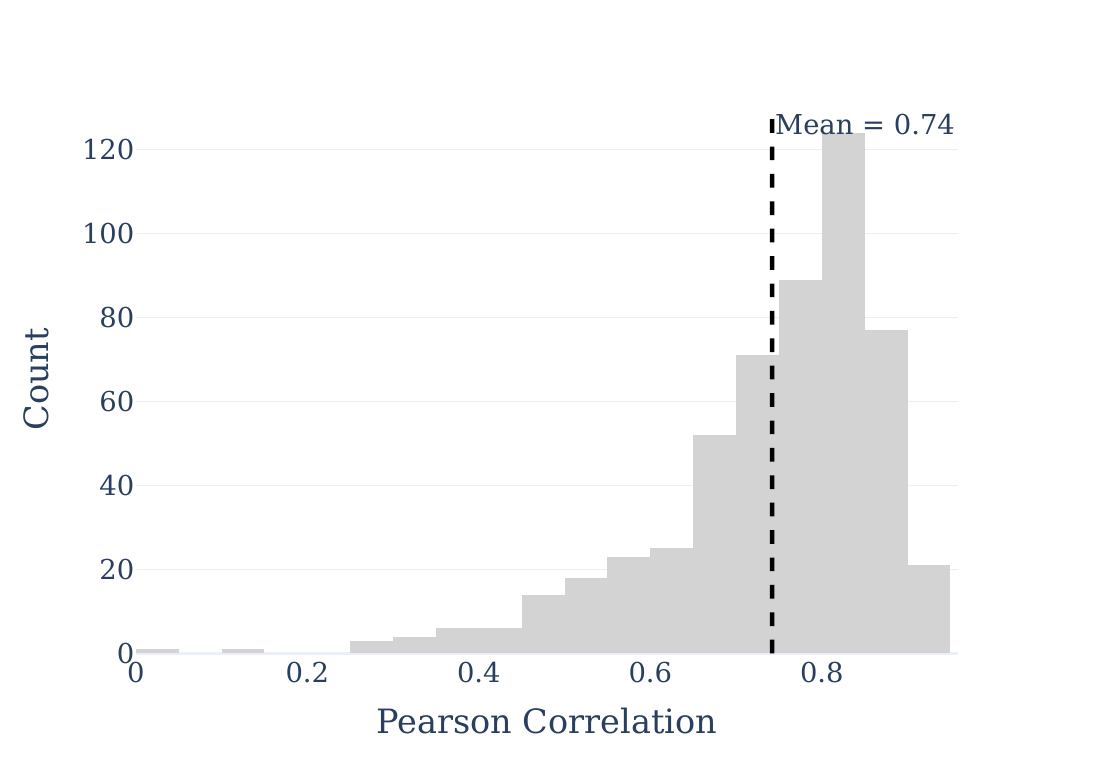}
        \caption{Pearson real (llama3.1-70b).}
    \end{subfigure}
    \begin{subfigure}{0.3\linewidth}
        \centering
        \includegraphics[width=\linewidth]{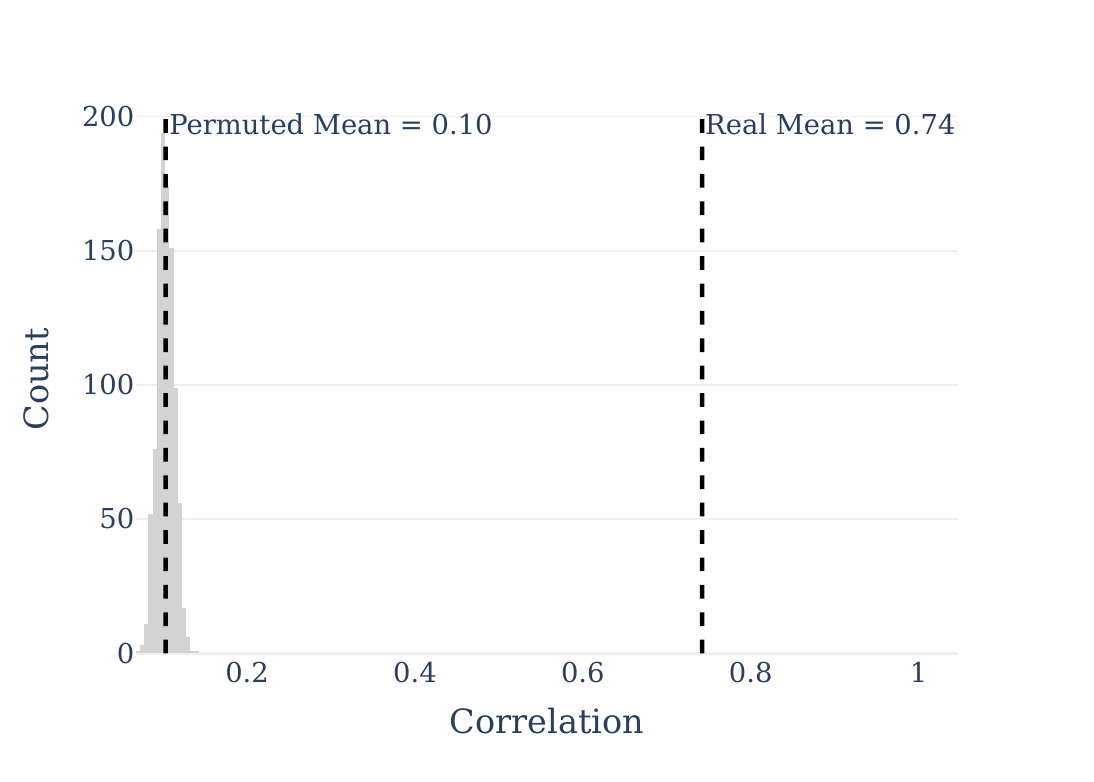}
        \caption{Pearson perm. (llama3.1-70b).}
    \end{subfigure}

    \begin{subfigure}{0.3\linewidth}
        \centering
        \includegraphics[width=\linewidth]{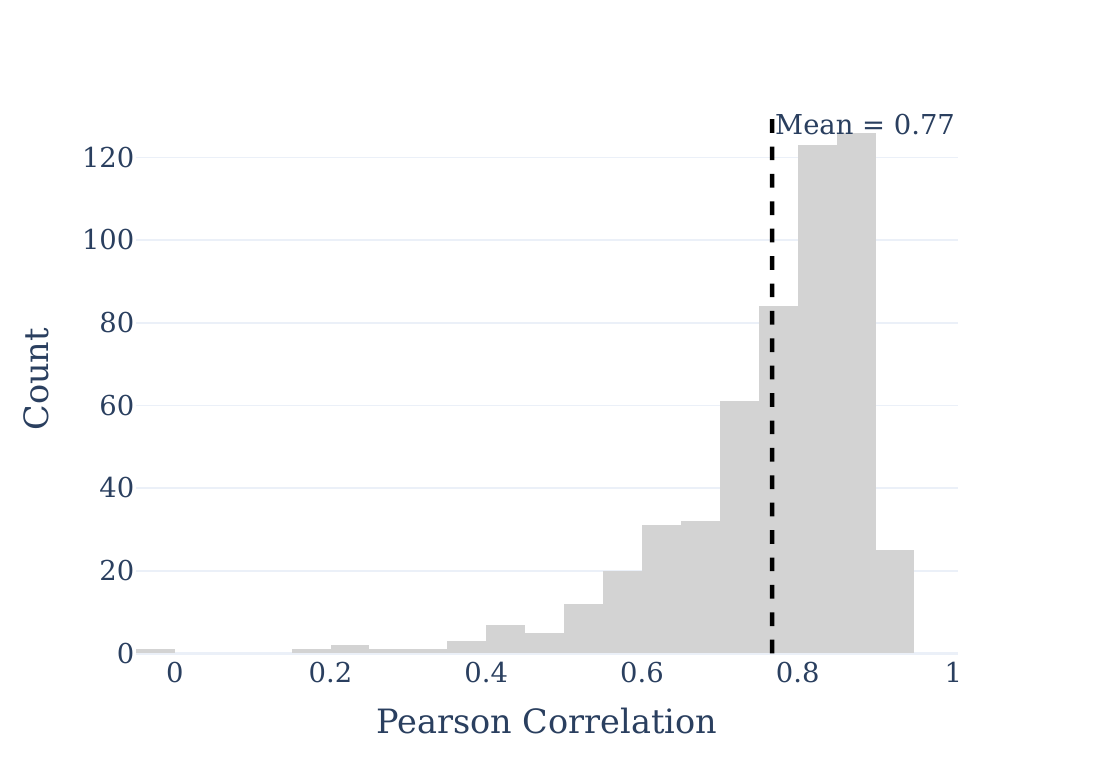}
        \caption{Pearson real (ensemble).}
    \end{subfigure}
    \begin{subfigure}{0.3\linewidth}
        \centering
        \includegraphics[width=\linewidth]{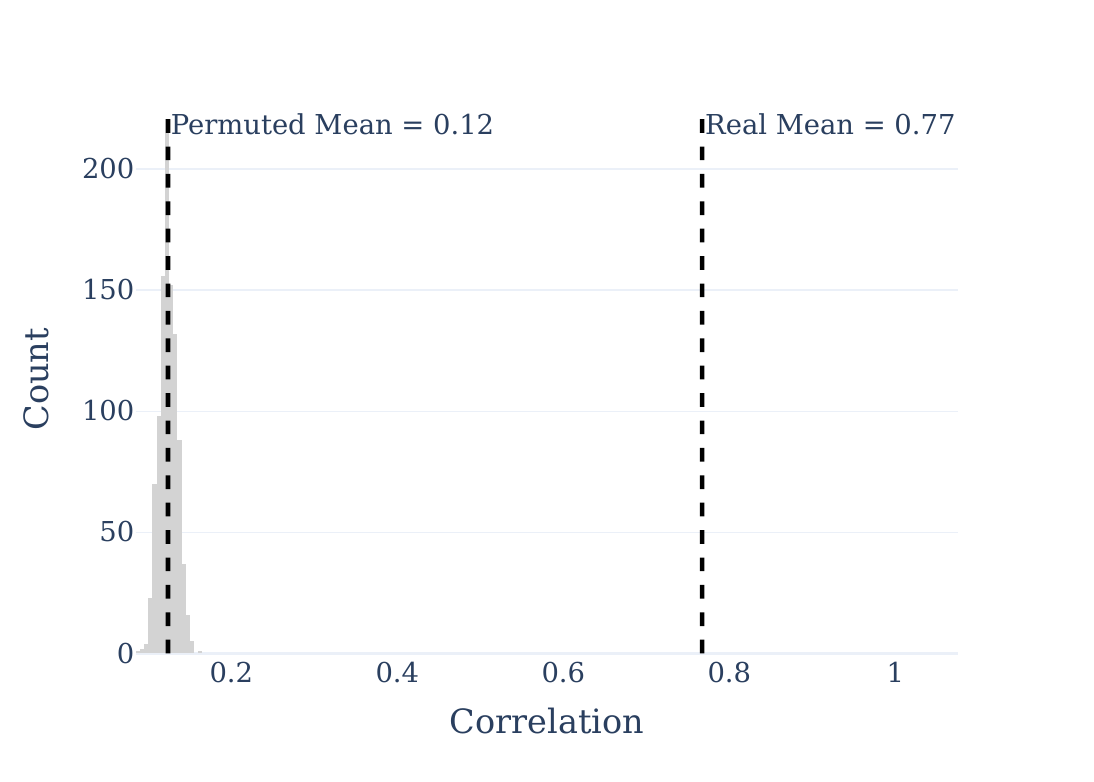}
        \caption{Pearson perm. (ensemble).}
    \end{subfigure}
    \caption{
    Distribution of Pearson correlation between LLM-generated semantic annotations and human ratings across models. 
    For each model, we show the distribution of correlation values computed on the true concept–property assignments (left) and under a permutation baseline obtained by randomly shuffling concept–property pairs (right). 
    Across all models, the true correlation distribution is substantially higher than the permutation baseline, indicating that LLM-based annotations capture meaningful semantic structure beyond chance. 
    Results are shown for individual models and the ensemble.
    }
    \label{fig:appendix_permutation_test_pearson_distributions}
\end{figure}

To assess whether the observed alignment between LLM-generated annotations and human semantic ratings reflects meaningful structure, we run a permutation test. 
Specifically, for each model, we randomly shuffle concept–property assignments and recompute evaluation metrics (MSE and Pearson correlation), thereby destroying the true semantic correspondence while preserving the marginal distributions of annotations.
We repeat this procedure $1{,}000$ times to obtain empirical null distributions for each metric. 
Figures~\ref{fig:appendix_permutation_test_mse_distributions} and~\ref{fig:appendix_permutation_test_pearson_distributions} show the distributions of scores obtained from the true annotations against those from the permutation baseline. 
Across all models, the real distributions are clearly separated from the permuted ones, with lower MSE and higher correlation values, showing that the annotations obtained from the LLM models capture systematic relationships between concepts and semantic properties.

\clearpage
\subsection{\method{} captures contextual shifts in human semantic judgments}
\label{appendix:results_c}

\subsubsection{Large-scale human behavioral study}
\label{appendix:results_human_survey}

Semantic feature norms provide interpretable representations of conceptual meaning along dimensions such as sensory, motor, affective, social, and cognitive properties.
However, existing human-annotated semantic spaces typically characterize concepts in isolation, assigning each concept a fixed set of property values regardless of the linguistic context in which it occurs.
\method{} extends the neurocognitively grounded semantic feature space of \citet{binder2016toward} to contextual settings by using LLMs to estimate semantic properties of concepts conditioned on their linguistic context.
In its controlled evaluation, \method{} used LLM-generated contexts designed to systematically increase or decrease individual concept–property associations and showed corresponding changes in its semantic annotations (see Appendix~\ref{appendix:synthetic_dataset} for more information).
To determine whether these contextual manipulations produce analogous changes in human semantic judgments, we conducted a human annotation study using the same controlled stimuli.

We selected $54$ of the $57$ properties included in the original synthetic dataset, excluding Temperature, Weight, and Texture. 
These three properties are derived by taking the maximum rating across pairs of opposing dimensions (Hot/Cold, Heavy/Light, and Rough/Smooth, respectively).
Extending this procedure to contextual annotations would require both constituent dimensions to be evaluated within the same sentence, which is difficult to guarantee when generating contexts that independently target each property.
For each of the $54$ retained properties, we selected six concepts based on their original out-of-context ratings: two with low, two with mid-range, and two with high concept–property association strengths according to the original out-of-context human ratings from \citet{binder2016toward}.
This stratification ensured coverage across the range of baseline concept–property association strengths.
This resulted in a total of $324$ concept–property pairs ($54 \times 6$).

Each concept–property pair was evaluated under three conditions: a high-target context (H$_\mathrm{T}$) designed to increase the association with the target property, a low-target context (L$_\mathrm{T}$) designed to decrease it, and an out-of-context condition ($O$) in which the concept was presented without a sentence.
This resulted in $324\times3=972$ unique annotation items. 
Participants rated the strength of each concept–property association on a $0$--$6$ scale, with contextual judgments made specifically with respect to the meaning of the concept in the provided sentence.

\textbf{Participants and data collection.}
We recruited participants through Prolific, requiring English fluency, a $98$--$100$\% approval rate, and at least $100$ previous submissions.
The study took approximately 10 minutes, and participants were paid USD 2 (USD 12/hour).
The $324$ concept--property pairs were split into two questionnaire versions, each covering half of the pairs, and data were collected in two rounds using the same stimuli.
In the first round, each participant rated $14$ semantic annotation items plus one attention check; in the second round, each participant rated $12$ items plus two attention checks.
Each participant saw only one condition for any given concept--property pair, and condition assignment was balanced across participants.
Each participant took part only once, in a single round and a single questionnaire version.
We applied the same quality-control criteria to all data: participants were excluded if they failed an attention check, had a mean response time below $7$ seconds per item, or gave the same score to all items (Table~\ref{tab:survey_funnel}).
In the first round, participants failing the attention check were already rejected during recruitment.
We further excluded one pair (Large/\emph{mouth}) for which an incorrect high-target sentence was shown in the first round, leaving $323$ pairs ($969$ items).
The final sample comprises $921$ participants and $11{,}983$ ratings, with on average $12.4$ ratings per item ($SD=1.4$, median $=12$, range $=8$--$20$).
The two rounds rank items consistently: the correlation between the item means of the two rounds is $0.81$ ($O$), $0.71$ ($L_\mathrm{T}$) and $0.66$ ($H_\mathrm{T}$), which, after correcting for the noise of each round's means (Spearman--Brown at the harmonic-mean number of ratings per round), corresponds to $1.02$, $1.03$ and $1.02$, i.e., as high as their reliability allows (values slightly above $1$ reflect estimation noise).
Second-round ratings were on average $0.14$--$0.20$ points lower in all conditions; since this offset is shared across items, it does not affect correlation-based analyses and, if anything, makes the pooled reliability estimates conservative.

\textbf{Contextual modulation.}
Human judgments showed a clear and systematic effect of linguistic context.
After averaging all ratings for each item, ratings were highest in the high-target condition ($M=4.23$, $SD=1.07$), intermediate in the out-of-context condition ($M=3.23$, $SD=1.47$), and lowest in the low-target condition ($M=2.67$, $SD=1.28$).
One-sided Wilcoxon signed-rank tests over concept--property pairs were significant for all three directional comparisons ($H_\mathrm{T} > O$, $O > L_\mathrm{T}$, and $H_\mathrm{T} > L_\mathrm{T}$; all $p<.001$; Table~\ref{tab:contextual_synthetic_data_results}).
Stratifying by the original Binder association strength shows that high-target contexts increased ratings in all bins, whereas low-target contexts decreased ratings only for pairs with mid or high baseline association: for pairs already rated low out of context ($L_\mathrm{O}$; $O=1.78$), low-target contexts had no further effect ($L_\mathrm{T}=1.73$, $p=.33$), consistent with a floor effect.
The \method{} ensemble shows the expected ordering in all bins.
The effect is broadly distributed across the semantic feature space: averaging over the concepts of each property, $H_\mathrm{T}$ ratings exceeded $O$ ratings for all $54$ properties and $L_\mathrm{T}$ ratings for $53$ of $54$ properties, and the complete ordering $H_\mathrm{T} > O > L_\mathrm{T}$ held for $46$ of $54$ properties (ensemble: $53$, $53$ and $43$).
All individual models also reproduce the human ordering (Table~\ref{tab:human_llm_context_manipulation}), including an ensemble that excludes \texttt{llama3.1-8b}, the model used to generate the sentences, showing that the effect does not depend on the generator scoring its own sentences.

\textbf{Together, these data provide direct human validation that interpretable semantic property judgments can be systematically modulated by linguistic context and support the central assumption underlying \method{}'s context-sensitive semantic annotations.}
They also provide a human-annotated resource for evaluating scalable approaches to contextual semantic annotation.
Importantly, obtaining about twelve independent judgments for only $972$ controlled annotation items already required more than $900$ participants, \emph{illustrating the practical difficulty of extending human annotation to arbitrary concepts and naturalistic contexts at scale}.

\begin{table}[t]
    \centering
    \caption{Participant flow per data-collection round and questionnaire version. Each version covers half of the concept--property pairs.}
    \small
    \begin{tabular}{lcccc}
    & \multicolumn{2}{c}{\textbf{Round 1}} & \multicolumn{2}{c}{\textbf{Round 2}} \\
    & Version 1 & Version 2 & Version 1 & Version 2 \\
    \midrule
    Survey responses              & 270 & 276 & 256 & 258 \\
    Completed and matched to Prolific & 249 & 255 & 250 & 250 \\
    After removing duplicates     & 245 & 255 & 250 & 250 \\
    Failed attention check(s)     & --$^\dagger$ & --$^\dagger$ & 25 & 29 \\
    Mean response time $<7$\,s    & 2 & 13 & 4 & 5 \\
    Same score on all items       & 1 & 0 & 0 & 0 \\
    \midrule
    \textbf{Final participants}   & \textbf{242} & \textbf{242} & \textbf{221} & \textbf{216} \\
    \bottomrule
    \multicolumn{5}{l}{\footnotesize $^\dagger$Rejected during recruitment.}
    \end{tabular}
    \label{tab:survey_funnel}
\end{table}

\begin{table}[t]
    \centering
    \caption{
    Contextual modulation of semantic feature scores for the human-evaluated subset.
    We report mean scores for human judgments and the \method{} ensemble across the original ($O$), low-target ($L_\mathrm{T}$), and high-target ($H_\mathrm{T}$) conditions, together with two- and one-sided Wilcoxon signed-rank tests.
    Results are shown across all concept–property pairs ($\mathrm{All}_\mathrm{O}$) and stratified by original out-of-context association strength ($L_\mathrm{O}$, $M_\mathrm{O}$, $H_\mathrm{O}$).
    Human means are computed from all ratings of each item. For pairs with low baseline association ($L_\mathrm{O}$), low-target contexts do not further decrease human ratings, consistent with a floor effect.
    }
    \resizebox{\textwidth}{!}{%
    \begin{tabular}{llccc ccc ccc}
    \textbf{Annotator} & \textbf{Concept}
    & \multicolumn{3}{c}{\textbf{Average scores}}
    & \multicolumn{3}{c}{\textbf{Wilcoxon $p$-value (two-sided)}}
    & \multicolumn{3}{c}{\textbf{Wilcoxon $p$-value (one-sided)}} \\
    & \textbf{Bin}
    & $\mathbf{L_T}$ & \textbf{O} & $\mathbf{H_T}$
    & $\mathbf{L_T}$ vs O
    & $\mathbf{H_T}$ vs O
    & $\mathbf{H_T}$ vs $\mathbf{L_T}$
    & $\mathbf{L_T<O}$
    & $\mathbf{H_T>O}$
    & $\mathbf{H_T>L_T}$ \\
    \midrule

    Human & All$_O$ & 2.67 & 3.23 & 4.23 & $<.001$ & $<.001$ & $<.001$ & $<.001$ & $<.001$ & $<.001$ \\
     & L$_O$ & 1.73 & 1.78 & 3.36 & $0.666$ & $<.001$ & $<.001$ & $0.333$ & $<.001$ & $<.001$ \\
     & M$_O$ & 2.75 & 3.43 & 4.33 & $<.001$ & $<.001$ & $<.001$ & $<.001$ & $<.001$ & $<.001$ \\
     & H$_O$ & 3.52 & 4.47 & 4.98 & $<.001$ & $<.001$ & $<.001$ & $<.001$ & $<.001$ & $<.001$ \\
    \midrule
    Ensemble & All$_O$ & 1.58 & 2.18 & 3.54 & $<.001$ & $<.001$ & $<.001$ & $<.001$ & $<.001$ & $<.001$ \\
     & L$_O$ & 0.57 & 0.69 & 2.21 & $<.001$ & $<.001$ & $<.001$ & $<.001$ & $<.001$ & $<.001$ \\
     & M$_O$ & 1.41 & 2.08 & 3.60 & $<.001$ & $<.001$ & $<.001$ & $<.001$ & $<.001$ & $<.001$ \\
     & H$_O$ & 2.74 & 3.76 & 4.80 & $<.001$ & $<.001$ & $<.001$ & $<.001$ & $<.001$ & $<.001$ \\

    \bottomrule
    \end{tabular}%
    }

    \label{tab:contextual_synthetic_data_results}
\end{table}

\begin{table}[t]
    \centering
     \caption{
     Comparison of low-target ($L_T$), original/no-context (O), and high-target ($H_T$) scores assigned by human annotators, individual LLMs, and LLM ensembles to the evaluated subset of the synthetic dataset.
     O denotes each annotator's own out-of-context score.
     All annotators exhibit the intended ordering $L_T < O < H_T$, with significant differences in all three expected directions ($L_T < O$, $H_T > O$, and $H_T > L_T$; all $p<.05$, one-sided Wilcoxon signed-rank tests over concept--property pairs).
     Two-sided tests are also reported for completeness.
     The ensemble excluding Llama3.1-8B provides a control excluding the model used to generate the synthetic sentences.
     }
    \scriptsize{
    \begin{tabular}{lccc ccc ccc}
    \textbf{Model}
    & \multicolumn{3}{c}{\textbf{Average scores}}
    & \multicolumn{3}{c}{\textbf{Wilcoxon $p$-value (two-sided)}}
    & \multicolumn{3}{c}{\textbf{Wilcoxon $p$-value (one-sided)}} \\
    & $\mathbf{L_T}$ & \textbf{O} & $\mathbf{H_T}$
    & $\mathbf{L_T}$ vs O
    & $\mathbf{H_T}$ vs O
    & $\mathbf{H_T}$ vs $\mathbf{L_T}$
    & $\mathbf{L_T < O}$
    & $\mathbf{H_T > O}$
    & $\mathbf{H_T > L_T}$ \\
    \midrule

    Human & 2.67 & 3.23 & 4.23 & $<.001$ & $<.001$ & $<.001$ & $<.001$ & $<.001$ & $<.001$ \\
    Llama3.1-70B & 1.71 & 2.53 & 4.09 & $<.001$ & $<.001$ & $<.001$ & $<.001$ & $<.001$ & $<.001$ \\
    Llama3.1-8B & 1.77 & 2.27 & 3.22 & $<.001$ & $<.001$ & $<.001$ & $<.001$ & $<.001$ & $<.001$ \\
    Mistral-7B & 1.39 & 1.60 & 3.50 & $0.026$ & $<.001$ & $<.001$ & $0.013$ & $<.001$ & $<.001$ \\
    Qwen2.5-7B & 1.43 & 2.32 & 3.36 & $<.001$ & $<.001$ & $<.001$ & $<.001$ & $<.001$ & $<.001$ \\
    \midrule
    Ensemble & 1.58 & 2.18 & 3.54 & $<.001$ & $<.001$ & $<.001$ & $<.001$ & $<.001$ & $<.001$ \\
    Ensemble w/o Llama3.1-8B & 1.51 & 2.15 & 3.65 & $<.001$ & $<.001$ & $<.001$ & $<.001$ & $<.001$ & $<.001$ \\
    \bottomrule
    \end{tabular}
    }
    \label{tab:human_llm_context_manipulation}
\end{table}

\clearpage
\subsubsection{Reliability of human judgments and noise ceiling}
\label{appendix:human_reliability}

\textbf{Balanced subsampling.}
Items received between $8$ and $20$ ratings, but the intraclass correlation assumes the same number of ratings per item.
We therefore draw $n=8$ ratings per item, from $8$ distinct participants, such that each participant contributes at most once to each of the $8$ rating columns, and repeat this for $20$ random draws.
All quantities are computed within each draw and then averaged across draws; ratings themselves are never averaged across draws.

\textbf{Intraclass correlation.}
For each condition, let $\mathrm{MS}_B$ and $\mathrm{MS}_W$ be the between-item and within-item mean squares of a one-way ANOVA over the $323 \times n$ rating matrix.
The one-way random-effects intraclass correlations are
\begin{equation}
\mathrm{ICC}(1,1) = \frac{\mathrm{MS}_B - \mathrm{MS}_W}{\mathrm{MS}_B + (n-1)\,\mathrm{MS}_W},
\qquad
\mathrm{ICC}(1,n) = \frac{\mathrm{MS}_B - \mathrm{MS}_W}{\mathrm{MS}_B}.
\end{equation}
$\mathrm{ICC}(1,1)$ is the expected correlation between two single raters, and $\mathrm{ICC}(1,n)$ the reliability of the mean of $n$ raters, i.e., the expected correlation between the means of two independent groups of $n$ raters.
The two are related by the Spearman--Brown formula, $\mathrm{ICC}(1,n) = n\,\mathrm{ICC}(1,1) / (1 + (n-1)\,\mathrm{ICC}(1,1))$, which we use to extrapolate reliability to other numbers of raters.
The one-way model treats raters as interchangeable, which matches our design: each item is rated by a different random subset of participants.

\textbf{Noise ceiling.}
A model whose scores equal the true item values correlates with the $n$-rater mean at $\sqrt{\mathrm{ICC}(1,n)}$, which we use as the human noise ceiling.
An independent split-half estimate (correlating two halves of $4$ raters and applying Spearman--Brown) gives the same ceilings to within $0.005$.

\begin{figure}[t]
    \centering
    \includegraphics[width=0.7\linewidth]{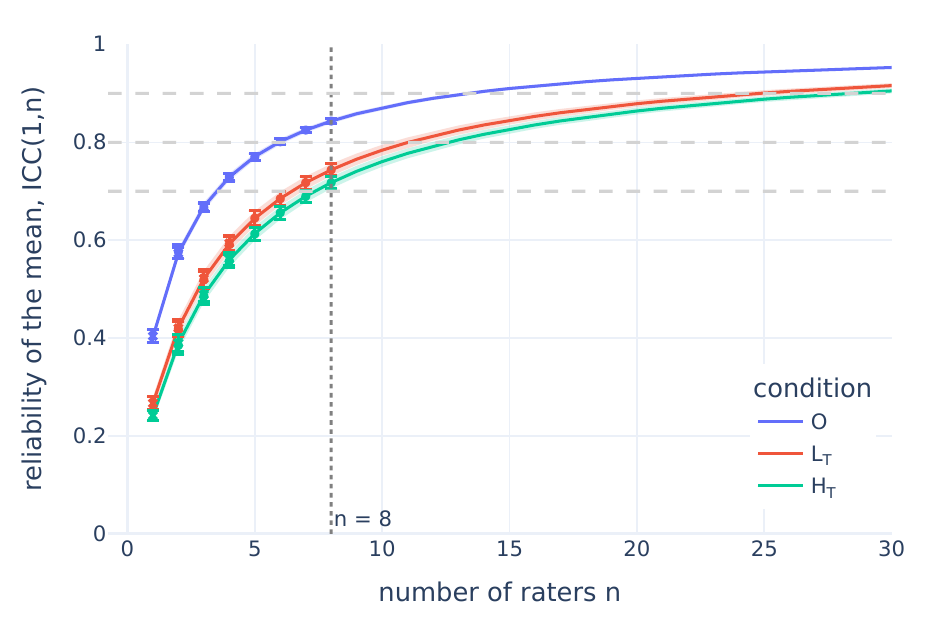}
    \caption{Reliability of the mean human rating as a function of the number of raters per item. Lines show the Spearman--Brown prediction from $\mathrm{ICC}(1,1)$ (shaded: $\pm$ SD over $20$ subsamples). Markers show empirical estimates from random subsets of raters: circles, $\mathrm{ICC}(1,n)$ computed on $n$ rater columns; crosses, the correlation between the means of two disjoint groups of $n$ raters (error bars: SD over subsamples). The dotted line marks $n=8$, the number of raters used in the main analyses; dashed grey lines mark reliabilities of $0.7$, $0.8$ and $0.9$.}
    \label{fig:reliability_curve}
\end{figure}

\textbf{Results.}
Table~\ref{tab:human_icc} summarizes human reliability.
Reliability is far above chance in all conditions: shuffling ratings across items yields null ICC$(1,8)$ values below $0.13$ (95th percentile; $1000$ permutations), compared to observed values of $0.72$--$0.84$ (all $p\leq.001$; $F$-tests $p<10^{-59}$).
Out-of-context judgments are more reliable than contextual ones (bootstrap over items: $O - L_\mathrm{T} = 0.10$ [$0.06$, $0.15$], $O - H_\mathrm{T} = 0.13$ [$0.08$, $0.18$]), while the two contextual conditions do not differ ($0.03$ [$-0.03$, $0.09$]).
Computing ICC(1,1) on all ratings without subsampling (ANOVA estimator for unequal group sizes) gives the same values ($0.400$, $0.271$, $0.245$).
Figure~\ref{fig:reliability_curve} shows how reliability grows with the number of raters: reaching $0.90$ would require about $14$ raters per item out of context and $25$--$29$ in context.

\begin{table}[t]
    \centering
    \caption{Reliability of human judgments per condition (mean over $20$ balanced subsamples of $n=8$ ratings per item; $95\%$ bootstrap CI over items). The last column reports the number of raters needed for a reliability of $0.90$ (Spearman--Brown).}
    \small
    \begin{tabular}{lccccc}
    \textbf{Condition} & $\mathrm{ICC}(1,1)$ & $\mathrm{ICC}(1,8)$ [95\% CI] & \textbf{Ceiling} & $F$ & \textbf{Raters for 0.90} \\
    \midrule
    $O$            & 0.40 & 0.84 [0.82, 0.87] & 0.92 & 6.39 & 14 \\
    $L_\mathrm{T}$ & 0.27 & 0.74 [0.70, 0.78] & 0.86 & 3.91 & 25 \\
    $H_\mathrm{T}$ & 0.24 & 0.72 [0.66, 0.76] & 0.85 & 3.55 & 29 \\
    \bottomrule
    \end{tabular}
    \label{tab:human_icc}
\end{table}

\clearpage
\subsubsection{Alignment of \method{} with human judgments relative to the noise ceiling}
\label{appendix:scribe_vs_ceiling}

\textbf{Per-model comparison.}
Table~\ref{tab:per_model_ceiling} reports the alignment of each model and the ensemble with the mean of $8$ human ratings.
The ensemble achieves the highest correlation in all three conditions; in terms of MSE, it is best in both contextual conditions, while \texttt{qwen2.5-7b} has a slightly lower MSE out of context, reflecting a scale closer to that of human ratings.

\begin{table}[t]
    \centering
    \caption{Alignment of individual models and the ensemble with human ratings (mean over $20$ balanced subsamples of $8$ ratings per item). Ceilings: $0.92$ ($O$), $0.86$ ($L_\mathrm{T}$), $0.85$ ($H_\mathrm{T}$).}
    \small
    \begin{tabular}{l ccc ccc}
    & \multicolumn{3}{c}{\textbf{$r$/ceiling} $\uparrow$} & \multicolumn{3}{c}{\textbf{MSE} $\downarrow$} \\
    \textbf{Model} & $O$ & $L_\mathrm{T}$ & $H_\mathrm{T}$ & $O$ & $L_\mathrm{T}$ & $H_\mathrm{T}$ \\
    \midrule
    Llama3.1-70B & 0.79 & 0.76 & 0.64 & 2.98 & 3.97 & 3.30 \\
    Llama3.1-8B  & 0.70 & 0.68 & 0.66 & 3.49 & 2.77 & 4.27 \\
    Mistral-7B   & 0.57 & 0.65 & 0.67 & 5.82 & 4.41 & 3.69 \\
    Qwen2.5-7B   & 0.76 & 0.70 & 0.63 & \textbf{2.35} & 3.41 & 2.59 \\
    \midrule
    Ensemble     & \textbf{0.80} & \textbf{0.81} & \textbf{0.75} & 2.59 & \textbf{2.64} & \textbf{2.31} \\
    \bottomrule
    \end{tabular}
    \label{tab:per_model_ceiling}
\end{table}

\textbf{Equivalent number of annotators.}
To express alignment in human units, we ask how many human raters $m$ would be needed for their mean to correlate with the $n$-rater mean as strongly as \method{} does.
For two independent groups of $m$ and $n$ raters, this correlation is $\sqrt{\mathrm{ICC}(1,m)\,\mathrm{ICC}(1,n)}$; solving $\sqrt{\mathrm{ICC}(1,m)\,\mathrm{ICC}(1,n)} = r$ with Spearman--Brown gives $m = 2.6$ ($O$), $5.1$ ($L_\mathrm{T}$) and $4.1$ ($H_\mathrm{T}$).
We confirm this empirically by splitting the $8$ raters into a reference group of $4$ and correlating the reference mean with (i) \method{} and (ii) the mean of $m=1,\dots,4$ raters from the other half (Table~\ref{tab:equiv_annotators}).
\method{} lies between $2$ and $3$ raters out of context, matches $4$ raters in the high-target condition, and exceeds $4$ raters in the low-target condition.

\begin{table}[t]
    \centering
    \caption{Correlation with the mean of $4$ held-out raters, for the mean of $m$ other raters and for the \method{} ensemble ($50$ random splits $\times$ $20$ subsamples).}
    \small
    \begin{tabular}{lccc}
    & $O$ & $L_\mathrm{T}$ & $H_\mathrm{T}$ \\
    \midrule
    $m=1$ rater  & 0.54 & 0.39 & 0.37 \\
    $m=2$ raters & 0.65 & 0.50 & 0.47 \\
    $m=3$ raters & 0.70 & 0.56 & 0.52 \\
    $m=4$ raters & 0.73 & 0.59 & 0.56 \\
    \midrule
    \method{} ensemble & 0.68 & 0.62 & 0.56 \\
    \bottomrule
    \end{tabular}
    \label{tab:equiv_annotators}
\end{table}

\textbf{MSE and systematic offset.}
Because the mean of $n$ raters is itself noisy, even a perfect model would not reach an MSE of zero: its expected MSE is the noise variance of the mean, $\mathrm{MS}_W/n$ (\emph{MSE floor}).
We further decompose the MSE as $\mathrm{MSE} = (\bar{y}-\bar{h})^2 + (s_y - s_h)^2 + 2 s_y s_h (1-r)$, where $\bar{y}, s_y$ and $\bar{h}, s_h$ are the mean and standard deviation of \method{} and human scores: the three terms capture a systematic offset, a difference in scale, and imperfect correlation (Table~\ref{tab:mse_decomposition}).
\method{} rates properties about one point lower than humans in $O$ and $L_\mathrm{T}$, and this offset accounts for $42$--$45\%$ of the MSE in these conditions ($20\%$ in $H_\mathrm{T}$).
For reference, the Binder norms also rate slightly lower than our participants on the same items (offset $-0.38$).

\begin{table}[t]
    \centering
    \caption{MSE between \method{} (or the Binder norms) and the mean of $8$ human ratings, with $95\%$ bootstrap CIs, the permutation baseline, the MSE floor, and the decomposition into offset, scale and correlation terms.}
    \small
    \resizebox{\textwidth}{!}{%
    \begin{tabular}{ll ccc c ccc}
    \textbf{Compared} & \textbf{Cond.} & \textbf{MSE} [95\% CI] & \textbf{Chance} & \textbf{Floor} & \textbf{Offset} $\bar{y}-\bar{h}$ & \textbf{Offset term} & \textbf{Scale term} & \textbf{Corr. term} \\
    \midrule
    \method{} & $O$            & 2.59 [2.22, 2.98] & 6.53 & 0.36 & $-1.04$ & 1.09 & 0.07 & 1.43 \\
    \method{} & $L_\mathrm{T}$ & 2.64 [2.31, 3.00] & 5.71 & 0.45 & $-1.09$ & 1.19 & 0.11 & 1.35 \\
    \method{} & $H_\mathrm{T}$ & 2.31 [1.91, 2.74] & 4.80 & 0.35 & $-0.68$ & 0.47 & 0.41 & 1.43 \\
    \midrule
    Binder norms & $O$         & 1.45 [1.22, 1.69] & 5.54 & 0.36 & $-0.38$ & 0.15 & 0.07 & 1.24 \\
    \bottomrule
    \end{tabular}}
    \label{tab:mse_decomposition}
\end{table}

\textbf{Human reference and robustness.}
The Binder norms correlate with our out-of-context ratings at $r=0.77$ ($84\%$ of the ceiling), close to \method{}'s $80\%$.
Using all ratings instead of balanced subsamples gives higher correlations, since the human means are less noisy ($r=0.75$, $0.73$, $0.67$ for $O$, $L_\mathrm{T}$, $H_\mathrm{T}$), but the same ceiling-normalized alignment ($0.80$, $0.80$, $0.75$, using approximate ceilings from Spearman--Brown at the harmonic-mean number of ratings).

\clearpage
\subsubsection{Results on the full synthetic dataset}
\label{appendix:results_full_synthetic_data}

\begin{table}[h]
    \centering
    \caption{
    Per-model results for the controlled contextual manipulation experiment (see Section~\ref{subsec:results_contextual_validation} and Appendix~\ref{appendix:synthetic_dataset}) using all the contextual generated sentences.
    We report average semantic scores and Wilcoxon signed-rank tests for each individual model and their ensemble. 
    $O$ denotes each model's own out-of-context score.
    For all models and the ensemble, semantic feature values shift in the intended direction ($H_\mathrm{T} > O > L_\mathrm{T}$) across all pairs and within the mid and high bins.
    For pairs with low baseline association ($L_\mathrm{O}$), \texttt{llama3.1-8b} and \texttt{mistral-7b} do not lower scores further in the low-target condition, mirroring the floor effect observed in human judgments (Table~\ref{tab:contextual_synthetic_data_results}); the ensemble shows significant effects in all bins.
    }
    \scriptsize{
    \begin{tabular}{l lccc ccc ccc}
    \textbf{Model} & \textbf{Concept} & \multicolumn{3}{c}{\textbf{Average scores}} & \multicolumn{3}{c}{\textbf{Wilcoxon $p$-value (two-sided)}} & \multicolumn{3}{c}{\textbf{Wilcoxon $p$-value (one-sided)}} \\
    \textbf{} & \textbf{Bin} & \textbf{L$_\mathrm{T}$} & \textbf{O} & \textbf{H$_\mathrm{T}$} & \textbf{L$_\mathrm{T}$ vs O} & \textbf{H$_\mathrm{T}$ vs O} & \textbf{H$_\mathrm{T}$ vs L$_\mathrm{T}$} & \textbf{L$_\mathrm{T}$ $<$ O} & \textbf{H$_\mathrm{T}$ $>$ O} & \textbf{H$_\mathrm{T}$ $>$ L$_\mathrm{T}$} \\
    \midrule
    \multirow{4}{*}{Llama3.1-8B}
        & All$_\mathrm{O}$ & 1.86 & 2.16 & 3.04 & $<.001$ & $<.001$ & $<.001$ & $<.001$ & $<.001$ & $<.001$ \\
        \cmidrule{2-11}
        & L$_\mathrm{O}$ & 0.67 & 0.62 & 1.39 & $0.168$ & $<.001$ & $<.001$ & $0.916$ & $<.001$ & $<.001$ \\
        & M$_\mathrm{O}$ & 1.86 & 2.06 & 3.08 & $0.009$ & $<.001$ & $<.001$ & $0.005$ & $<.001$ & $<.001$ \\
        & H$_\mathrm{O}$ & 3.06 & 3.81 & 4.64 & $<.001$ & $<.001$ & $<.001$ & $<.001$ & $<.001$ & $<.001$ \\
    \midrule
    \multirow{4}{*}{Mistral-7B}
        & All$_\mathrm{O}$ & 1.37 & 1.55 & 3.11 & $<.001$ & $<.001$ & $<.001$ & $<.001$ & $<.001$ & $<.001$ \\
        \cmidrule{2-11}
        & L$_\mathrm{O}$ & 0.35 & 0.28 & 1.59 & $0.048$ & $<.001$ & $<.001$ & $0.976$ & $<.001$ & $<.001$ \\
        & M$_\mathrm{O}$ & 1.23 & 1.40 & 3.30 & $0.042$ & $<.001$ & $<.001$ & $0.021$ & $<.001$ & $<.001$ \\
        & H$_\mathrm{O}$ & 2.52 & 2.97 & 4.43 & $<.001$ & $<.001$ & $<.001$ & $<.001$ & $<.001$ & $<.001$ \\
    \midrule
    \multirow{4}{*}{Qwen2.5-7B}
        & All$_\mathrm{O}$ & 1.57 & 2.25 & 3.04 & $<.001$ & $<.001$ & $<.001$ & $<.001$ & $<.001$ & $<.001$ \\
        \cmidrule{2-11}
        & L$_\mathrm{O}$ & 0.60 & 0.95 & 1.85 & $<.001$ & $<.001$ & $<.001$ & $<.001$ & $<.001$ & $<.001$ \\
        & M$_\mathrm{O}$ & 1.47 & 2.21 & 3.16 & $<.001$ & $<.001$ & $<.001$ & $<.001$ & $<.001$ & $<.001$ \\
        & H$_\mathrm{O}$ & 2.63 & 3.59 & 4.12 & $<.001$ & $<.001$ & $<.001$ & $<.001$ & $<.001$ & $<.001$ \\
    \midrule
    \multirow{4}{*}{Llama3.1-70B}
        & All$_\mathrm{O}$ & 1.81 & 2.45 & 3.68 & $<.001$ & $<.001$ & $<.001$ & $<.001$ & $<.001$ & $<.001$ \\
        \cmidrule{2-11}
        & L$_\mathrm{O}$ & 0.33 & 0.47 & 2.04 & $0.043$ & $<.001$ & $<.001$ & $0.021$ & $<.001$ & $<.001$ \\
        & M$_\mathrm{O}$ & 1.81 & 2.51 & 3.95 & $<.001$ & $<.001$ & $<.001$ & $<.001$ & $<.001$ & $<.001$ \\
        & H$_\mathrm{O}$ & 3.29 & 4.38 & 5.05 & $<.001$ & $<.001$ & $<.001$ & $<.001$ & $<.001$ & $<.001$ \\
    \midrule
    \multirow{4}{*}{Ensemble}
        & All$_\mathrm{O}$ & 1.65 & 2.10 & 3.22 & $<.001$ & $<.001$ & $<.001$ & $<.001$ & $<.001$ & $<.001$ \\
        \cmidrule{2-11}
        & L$_\mathrm{O}$ & 0.49 & 0.58 & 1.72 & $<.001$ & $<.001$ & $<.001$ & $<.001$ & $<.001$ & $<.001$ \\
        & M$_\mathrm{O}$ & 1.59 & 2.05 & 3.37 & $<.001$ & $<.001$ & $<.001$ & $<.001$ & $<.001$ & $<.001$ \\
        & H$_\mathrm{O}$ & 2.87 & 3.69 & 4.56 & $<.001$ & $<.001$ & $<.001$ & $<.001$ & $<.001$ & $<.001$ \\
    \bottomrule
    \end{tabular}
    }
    \label{tab:all_contextual_synthetic_data_results_individual_models}
\end{table}

\clearpage
\subsection{\method{} predicts brain activity}
\label{appendix:results_brainenc}

\subsubsection{Annotation design: word types and context size}
\label{appendix:results_brain_wordtypes_contextlen}

In this section, we report correlation values for different word type annotations and context sizes, evaluated separately for each model on voxels with a noise ceiling greater than $0.4$~(see Appendix~\ref{appendix:noise_ceiling} for details on noise ceiling estimation).
Tables~\ref{tab:annotation_context_results_llama3.1-8b-instruct}, \ref{tab:annotation_context_results_mistral-7b-instruct}, \ref{tab:annotation_context_results_qwen2.5-7b-instruct}, and \ref{tab:annotation_context_results_llama3.1-70b-instruct} present the results for the instruct-based \texttt{llama3.1-8B}, \texttt{mistral-7b}, \texttt{qwen2.5-7b}, and \texttt{llama3.1-70b} models, respectively.

Across most models, we observe a consistent pattern: performance significantly improves with contextual information, with a context window of $k=10$ yielding the best results.
This context size aligns with prior work~\cite{jain2018incorporating, toneva2019interpreting}, which shows that increasing context improves encoding performance up to a point, after which the gains saturate.
Moreover, using the full set of semantic POS tags (\textit{All Semantic POS}) leads to the highest overall correlations.

\begin{table}[h]
    \centering
    \caption{
    Brain encoding correlations across context sizes using \texttt{llama3.1-8b-instruct}.
    We compare out-of-context annotations, fixed-size context windows ($k \in \{3,5,10,50\}$), and an \emph{up to concept} condition including all preceding stimulus context.
    Results are reported for three POS subsets: \textit{Nouns},
    \textit{Nouns, Verbs, Adjectives}, and \textit{All Semantic POS}.
    Values correspond to noise-ceiling-normalized correlations.
    Bold indicates the best context within each POS group, and underlining indicates the overall best configuration.
    $\ast$ denotes significant improvements over the out-of-context baseline (one-sided Wilcoxon signed-rank test, $p<0.05$).
    }
    \scriptsize{
    \begin{tabular}{llcccccccc}
    \textbf{Word type} & \textbf{Context Size}
    & \textbf{S1} & \textbf{S2} & \textbf{S3}
    & \textbf{S5} & \textbf{S6} & \textbf{S7} & \textbf{S8}
    & \textbf{Avg $\pm$ STE} \\
    \midrule

    \multirow{6}{*}{Nouns}
    & out-of-context & 0.1372 & 0.1122 & 0.1190 & 0.1066 & 0.0933 & 0.0896 & 0.0931 & 0.1073 $\pm$ 0.0065 \\
    & window/k3      & 0.1420 & 0.1237 & 0.1301 & 0.1217 & 0.0789 & 0.0881 & 0.0807 & 0.1093 $\pm$ 0.0098 \\
    & window/k5      & 0.1442 & 0.1202 & 0.1330 & 0.1129 & 0.1051 & 0.0873 & 0.0884 & 0.1130 $\pm$ 0.0081$^\ast$ \\
    & window/k10     & 0.1621 & 0.1362 & 0.1384 & 0.1200 & 0.1101 & 0.0869 & 0.0858 & 0.1199 $\pm$ 0.0106$^\ast$ \\
    & \textbf{window/k50}
                        & 0.1760 & 0.1485 & 0.1207 & 0.1381 & 0.1116 & 0.0831 & 0.0928
                        & \textbf{0.1244 $\pm$ 0.0123} \\
    & up to concept  & 0.1566 & 0.1191 & 0.1001 & 0.1048 & 0.0732 & 0.0777 & 0.0681 & 0.1000 $\pm$ 0.0118 \\

    \midrule

    \multirow{6}{*}{Nouns, Verbs, Adj}
    & out-of-context & 0.1854 & 0.1479 & 0.1379 & 0.1534 & 0.1392 & 0.1222 & 0.1129 & 0.1427 $\pm$ 0.0089 \\
    & window/k3      & 0.1841 & 0.1713 & 0.1446 & 0.1574 & 0.1500 & 0.1258 & 0.1062 & 0.1485 $\pm$ 0.0100 \\
    & window/k5      & 0.1985 & 0.1741 & 0.1483 & 0.1666 & 0.1704 & 0.1293 & 0.1124 & 0.1571 $\pm$ 0.0110$^\ast$ \\
    & window/k10     & 0.2030 & 0.1795 & 0.1463 & 0.1654 & 0.1710 & 0.1189 & 0.1032 & 0.1553 $\pm$ 0.0132 \\
    & \textbf{window/k50}
                        & 0.2228 & 0.1822 & 0.1330 & 0.1669 & 0.1854 & 0.1236 & 0.1051
                        & \textbf{0.1599 $\pm$ 0.0156} \\
    & up to concept  & 0.2137 & 0.1749 & 0.1105 & 0.1465 & 0.1516 & 0.1186 & 0.1020 & 0.1454 $\pm$ 0.0150 \\

    \midrule

    \multirow{6}{*}{All Semantic POS}
    & out-of-context & 0.1869 & 0.1516 & 0.1515 & 0.1665 & 0.1142 & 0.1251 & 0.1099 & 0.1437 $\pm$ 0.0108 \\
    & window/k3      & 0.1972 & 0.1800 & 0.1561 & 0.1723 & 0.1292 & 0.1359 & 0.1085 & 0.1542 $\pm$ 0.0118$^\ast$ \\
    & window/k5      & 0.2042 & 0.1909 & 0.1519 & 0.1723 & 0.1458 & 0.1346 & 0.1154 & 0.1593 $\pm$ 0.0119$^\ast$ \\
    & \underline{\textbf{window/k10}}
                        & 0.2107 & 0.1996 & 0.1546 & 0.1849 & 0.1543 & 0.1300 & 0.1036
                        & \underline{\textbf{0.1625 $\pm$ 0.0147}}$^\ast$ \\
    & window/k50     & 0.2215 & 0.1881 & 0.1483 & 0.1634 & 0.1666 & 0.1181 & 0.0912 & 0.1567 $\pm$ 0.0163 \\
    & up to concept  & 0.2153 & 0.1633 & 0.1259 & 0.1546 & 0.1072 & 0.1146 & 0.0957 & 0.1395 $\pm$ 0.0157 \\

    \bottomrule
    \end{tabular}
    }
    \label{tab:annotation_context_results_llama3.1-8b-instruct}
\end{table}

\begin{table}[h]
    \centering
    \caption{
    Brain encoding correlations across context sizes using \texttt{mistral-7b-instruct}.
    We compare out-of-context annotations, fixed-size context windows ($k \in \{3,5,10,50\}$), and an \emph{up to concept} condition including all preceding stimulus context.
    Results are reported for three POS subsets: \textit{Nouns},
    \textit{Nouns, Verbs, Adjectives}, and \textit{All Semantic POS}.
    Values correspond to noise-ceiling-normalized correlations.
    Bold indicates the best context within each POS group, and underlining indicates the overall best configuration.
    $\ast$ denotes significant improvements over the out-of-context baseline (one-sided Wilcoxon signed-rank test, $p<0.05$).
    }
    \scriptsize{
    \begin{tabular}{llcccccccc}
    \textbf{Word type} & \textbf{Context Size}
    & \textbf{S1} & \textbf{S2} & \textbf{S3}
    & \textbf{S5} & \textbf{S6} & \textbf{S7} & \textbf{S8}
    & \textbf{Avg $\pm$ STE} \\
    \midrule

    \multirow{6}{*}{Nouns}
    & out-of-context & 0.1368 & 0.0917 & 0.1012 & 0.0848 & 0.0821 & 0.0871 & 0.0693 & 0.0933 $\pm$ 0.0081 \\
    & window/k3      & 0.1449 & 0.1248 & 0.1000 & 0.0938 & 0.0445 & 0.0939 & 0.0721 & 0.0963 $\pm$ 0.0124 \\
    & window/k5      & 0.1506 & 0.1337 & 0.1156 & 0.0975 & 0.0794 & 0.1048 & 0.0898 & 0.1102 $\pm$ 0.0095$^\ast$ \\
    & \textbf{window/k10}
                        & 0.1729 & 0.1521 & 0.1340 & 0.1300 & 0.0897 & 0.1103 & 0.0944
                        & \textbf{0.1262 $\pm$ 0.0115}$^\ast$ \\
    & window/k50     & 0.1622 & 0.1441 & 0.1248 & 0.1206 & 0.0871 & 0.0975 & 0.0880 & 0.1177 $\pm$ 0.0109$^\ast$ \\
    & up to concept  & 0.1318 & 0.1171 & 0.1016 & 0.1051 & 0.0790 & 0.0832 & 0.0720 & 0.0985 $\pm$ 0.0082 \\

    \midrule

    \multirow{6}{*}{Nouns, Verbs, Adj}
    & out-of-context & 0.1828 & 0.1414 & 0.1138 & 0.1405 & 0.1124 & 0.1306 & 0.0918 & 0.1305 $\pm$ 0.0110 \\
    & window/k3      & 0.2036 & 0.1859 & 0.1055 & 0.1539 & 0.1180 & 0.1350 & 0.0984 & 0.1429 $\pm$ 0.0152 \\
    & window/k5      & 0.2041 & 0.1829 & 0.1279 & 0.1650 & 0.1195 & 0.1476 & 0.1154 & 0.1518 $\pm$ 0.0128$^\ast$ \\
    & \textbf{window/k10}
                        & 0.2184 & 0.2023 & 0.1299 & 0.1836 & 0.1452 & 0.1511 & 0.1309
                        & \textbf{0.1659 $\pm$ 0.0134}$^\ast$ \\
    & window/k50     & 0.2136 & 0.1757 & 0.1294 & 0.1661 & 0.1409 & 0.1532 & 0.1032 & 0.1546 $\pm$ 0.0134$^\ast$ \\
    & up to concept  & 0.1809 & 0.1781 & 0.0948 & 0.1475 & 0.1312 & 0.1417 & 0.0982 & 0.1389 $\pm$ 0.0130 \\

    \midrule

    \multirow{6}{*}{All Semantic POS}
    & out-of-context & 0.1892 & 0.1623 & 0.1463 & 0.1600 & 0.0923 & 0.1382 & 0.0958 & 0.1406 $\pm$ 0.0134 \\
    & window/k3      & 0.2104 & 0.1972 & 0.1322 & 0.1787 & 0.1184 & 0.1289 & 0.1034 & 0.1527 $\pm$ 0.0159 \\
    & window/k5      & 0.2141 & 0.2016 & 0.1506 & 0.1766 & 0.1117 & 0.1438 & 0.1171 & 0.1594 $\pm$ 0.0150$^\ast$ \\
    & \underline{\textbf{window/k10}}
                        & 0.2240 & 0.2159 & 0.1638 & 0.1897 & 0.1463 & 0.1493 & 0.1321
                        & \underline{\textbf{0.1744 $\pm$ 0.0136}}$^\ast$ \\
    & window/k50     & 0.2175 & 0.1879 & 0.1636 & 0.1685 & 0.1421 & 0.1562 & 0.1102 & 0.1637 $\pm$ 0.0128$^\ast$ \\
    & up to concept  & 0.1792 & 0.1810 & 0.1141 & 0.1565 & 0.0933 & 0.1314 & 0.0865 & 0.1346 $\pm$ 0.0147 \\

    \bottomrule
    \end{tabular}
    }
    \label{tab:annotation_context_results_mistral-7b-instruct}
\end{table}

\begin{table}[h]
    \centering
    \caption{
    Brain encoding correlations across context sizes using \texttt{qwen2.5-7b-instruct}.
    We compare out-of-context annotations, fixed-size context windows ($k \in \{3,5,10,50\}$), and an \emph{up to concept} condition including all preceding stimulus context.
    Results are reported for three POS subsets: \textit{Nouns},
    \textit{Nouns, Verbs, Adjectives}, and \textit{All Semantic POS}.
    Values correspond to noise-ceiling-normalized correlations.
    Bold indicates the best context within each POS group, and underlining indicates the overall best configuration.
    $\ast$ denotes significant improvements over the out-of-context baseline (one-sided Wilcoxon signed-rank test, $p<0.05$).
    }
    \scriptsize{
    \begin{tabular}{llcccccccc}
    \toprule
    \textbf{Word type} & \textbf{Context Size}
    & \textbf{S1} & \textbf{S2} & \textbf{S3}
    & \textbf{S5} & \textbf{S6} & \textbf{S7} & \textbf{S8}
    & \textbf{Avg $\pm$ STE} \\
    \midrule

    \multirow{6}{*}{Nouns}
    & out-of-context & 0.1420 & 0.1083 & 0.1209 & 0.1181 & 0.0919 & 0.0883 & 0.0811 & 0.1072 $\pm$ 0.0082 \\
    & window/k3      & 0.1537 & 0.1455 & 0.1360 & 0.1236 & 0.0880 & 0.0857 & 0.0906 & 0.1176 $\pm$ 0.0110$^\ast$ \\
    & window/k5      & 0.1618 & 0.1482 & 0.1321 & 0.1304 & 0.1061 & 0.0927 & 0.0917 & 0.1233 $\pm$ 0.0103$^\ast$ \\
    & \textbf{window/k10}
                        & 0.1809 & 0.1581 & 0.1449 & 0.1396 & 0.1097 & 0.0958 & 0.1036
                        & \textbf{0.1332 $\pm$ 0.0119}$^\ast$ \\
    & window/k50     & 0.1823 & 0.1363 & 0.1358 & 0.1358 & 0.0987 & 0.0878 & 0.0883 & 0.1236 $\pm$ 0.0130$^\ast$ \\
    & up to concept  & 0.1606 & 0.1162 & 0.1415 & 0.1211 & 0.0917 & 0.0856 & 0.0876 & 0.1149 $\pm$ 0.0109$^\ast$ \\

    \midrule

    \multirow{6}{*}{Nouns, Verbs, Adj}
    & out-of-context & 0.1842 & 0.1583 & 0.1170 & 0.1543 & 0.1338 & 0.1363 & 0.0960 & 0.1400 $\pm$ 0.0109 \\
    & window/k3      & 0.2016 & 0.1888 & 0.1362 & 0.1716 & 0.1535 & 0.1292 & 0.1142 & 0.1564 $\pm$ 0.0122$^\ast$ \\
    & window/k5      & 0.2084 & 0.1880 & 0.1335 & 0.1839 & 0.1647 & 0.1357 & 0.1166 & 0.1616 $\pm$ 0.0128$^\ast$ \\
    & \textbf{window/k10}
                        & 0.2180 & 0.2032 & 0.1433 & 0.1922 & 0.1806 & 0.1404 & 0.1324
                        & \textbf{0.1729 $\pm$ 0.0129}$^\ast$ \\
    & window/k50     & 0.2256 & 0.1821 & 0.1362 & 0.1891 & 0.1696 & 0.1284 & 0.1158 & 0.1638 $\pm$ 0.0148$^\ast$ \\
    & up to concept  & 0.2045 & 0.1595 & 0.1434 & 0.1582 & 0.1318 & 0.1412 & 0.1117 & 0.1500 $\pm$ 0.0110$^\ast$ \\

    \midrule

    \multirow{6}{*}{All Semantic POS}
    & out-of-context & 0.1934 & 0.1655 & 0.1324 & 0.1619 & 0.1199 & 0.1583 & 0.0963 & 0.1468 $\pm$ 0.0123 \\
    & window/k3      & 0.2073 & 0.2022 & 0.1585 & 0.1821 & 0.1585 & 0.1420 & 0.1078 & 0.1655 $\pm$ 0.0132$^\ast$ \\
    & window/k5      & 0.2164 & 0.2025 & 0.1593 & 0.1995 & 0.1660 & 0.1432 & 0.1196 & 0.1724 $\pm$ 0.0133$^\ast$ \\
    & \underline{\textbf{window/k10}}
                        & 0.2242 & 0.2227 & 0.1701 & 0.1967 & 0.1740 & 0.1516 & 0.1318
                        & \underline{\textbf{0.1816 $\pm$ 0.0132}}$^\ast$ \\
    & window/k50     & 0.2342 & 0.1888 & 0.1423 & 0.1930 & 0.1759 & 0.1383 & 0.1249 & 0.1711 $\pm$ 0.0145$^\ast$ \\
    & up to concept  & 0.1964 & 0.1721 & 0.1416 & 0.1634 & 0.1028 & 0.1363 & 0.1015 & 0.1449 $\pm$ 0.0133 \\

    \bottomrule
    \end{tabular}
    }
    \label{tab:annotation_context_results_qwen2.5-7b-instruct}
\end{table}

\begin{table}[h]
    \centering
    \caption{
    Brain encoding correlations across context sizes using \texttt{llama3.1-70b-instruct}.
    We compare out-of-context annotations, fixed-size context windows ($k \in \{3,5,10,50\}$), and an \emph{up to concept} condition including all preceding stimulus context.
    Results are reported for three POS subsets: \textit{Nouns},
    \textit{Nouns, Verbs, Adjectives}, and \textit{All Semantic POS}.
    Values correspond to noise-ceiling-normalized correlations.
    Bold indicates the best context within each POS group, and underlining indicates the overall best configuration.
    $\ast$ denotes significant improvements over the out-of-context baseline (one-sided Wilcoxon signed-rank test, $p<0.05$).
    }
    \scriptsize{
    \begin{tabular}{llcccccccc}
    \toprule
    \textbf{Word type} & \textbf{Context Size}
    & \textbf{S1} & \textbf{S2} & \textbf{S3}
    & \textbf{S5} & \textbf{S6} & \textbf{S7} & \textbf{S8}
    & \textbf{Avg $\pm$ STE} \\
    \midrule

    \multirow{6}{*}{Nouns}
    & out-of-context & 0.1347 & 0.1239 & 0.1135 & 0.1104 & 0.0942 & 0.0943 & 0.0872 & 0.1083 $\pm$ 0.0066 \\
    & window/k3      & 0.1454 & 0.1430 & 0.1136 & 0.1143 & 0.0832 & 0.0971 & 0.0838 & 0.1115 $\pm$ 0.0097 \\
    & window/k5      & 0.1502 & 0.1382 & 0.1241 & 0.1247 & 0.0791 & 0.1026 & 0.0926 & 0.1160 $\pm$ 0.0096 \\
    & \textbf{window/k10}
                        & 0.1711 & 0.1519 & 0.1295 & 0.1300 & 0.0927 & 0.1094 & 0.0961
                        & \textbf{0.1258 $\pm$ 0.0109}$^\ast$ \\
    & window/k50     & 0.1789 & 0.1372 & 0.1069 & 0.1370 & 0.0686 & 0.0948 & 0.0947 & 0.1169 $\pm$ 0.0139 \\
    & up to concept  & 0.1509 & 0.1383 & 0.1080 & 0.1200 & 0.0524 & 0.0905 & 0.1028 & 0.1090 $\pm$ 0.0123 \\

    \midrule

    \multirow{6}{*}{Nouns, Verbs, Adj}
    & out-of-context & 0.1710 & 0.1649 & 0.1275 & 0.1379 & 0.1420 & 0.1194 & 0.1020 & 0.1378 $\pm$ 0.0092 \\
    & window/k3      & 0.1952 & 0.2025 & 0.1409 & 0.1590 & 0.1495 & 0.1286 & 0.1015 & 0.1539 $\pm$ 0.0135$^\ast$ \\
    & window/k5      & 0.2048 & 0.2045 & 0.1474 & 0.1772 & 0.1600 & 0.1381 & 0.1202 & 0.1646 $\pm$ 0.0123$^\ast$ \\
    & \textbf{window/k10}
                        & 0.2174 & 0.2152 & 0.1570 & 0.1885 & 0.1821 & 0.1418 & 0.1357
                        & \textbf{0.1768 $\pm$ 0.0125}$^\ast$ \\
    & window/k50     & 0.2196 & 0.1881 & 0.1415 & 0.1802 & 0.1455 & 0.1329 & 0.1184 & 0.1609 $\pm$ 0.0136$^\ast$ \\
    & up to concept  & 0.2083 & 0.1806 & 0.1177 & 0.1542 & 0.1434 & 0.1364 & 0.1449 & 0.1551 $\pm$ 0.0114$^\ast$ \\

    \midrule

    \multirow{6}{*}{All Semantic POS}
    & out-of-context & 0.1783 & 0.1628 & 0.1482 & 0.1525 & 0.1179 & 0.1306 & 0.0874 & 0.1397 $\pm$ 0.0115 \\
    & window/k3      & 0.1992 & 0.1972 & 0.1569 & 0.1686 & 0.1422 & 0.1413 & 0.0950 & 0.1572 $\pm$ 0.0137$^\ast$ \\
    & window/k5      & 0.2005 & 0.2197 & 0.1693 & 0.1880 & 0.1642 & 0.1432 & 0.1171 & 0.1693 $\pm$ 0.0143$^\ast$ \\
    & \underline{\textbf{window/k10}}
                        & 0.2182 & 0.2181 & 0.1731 & 0.1993 & 0.1758 & 0.1522 & 0.1235
                        & \underline{\textbf{0.1800 $\pm$ 0.0132}}$^\ast$ \\
    & window/k50     & 0.2282 & 0.2017 & 0.1696 & 0.1747 & 0.1459 & 0.1381 & 0.1237 & 0.1688 $\pm$ 0.0139$^\ast$ \\
    & up to concept  & 0.2141 & 0.1887 & 0.1324 & 0.1669 & 0.1197 & 0.1386 & 0.1402 & 0.1572 $\pm$ 0.0129 \\

    \bottomrule
    \end{tabular}
    }
    \label{tab:annotation_context_results_llama3.1-70b-instruct}
\end{table}

\clearpage
\subsubsection{Voxel-wise difference brain maps of out-of-context and contextual \method{}}
\label{appendix:results_brain_encoding_diff_plots}

\begin{figure}[h]
    \centering
    \begin{subfigure}{0.45\linewidth}
        \centering
        \includegraphics[width=\linewidth]{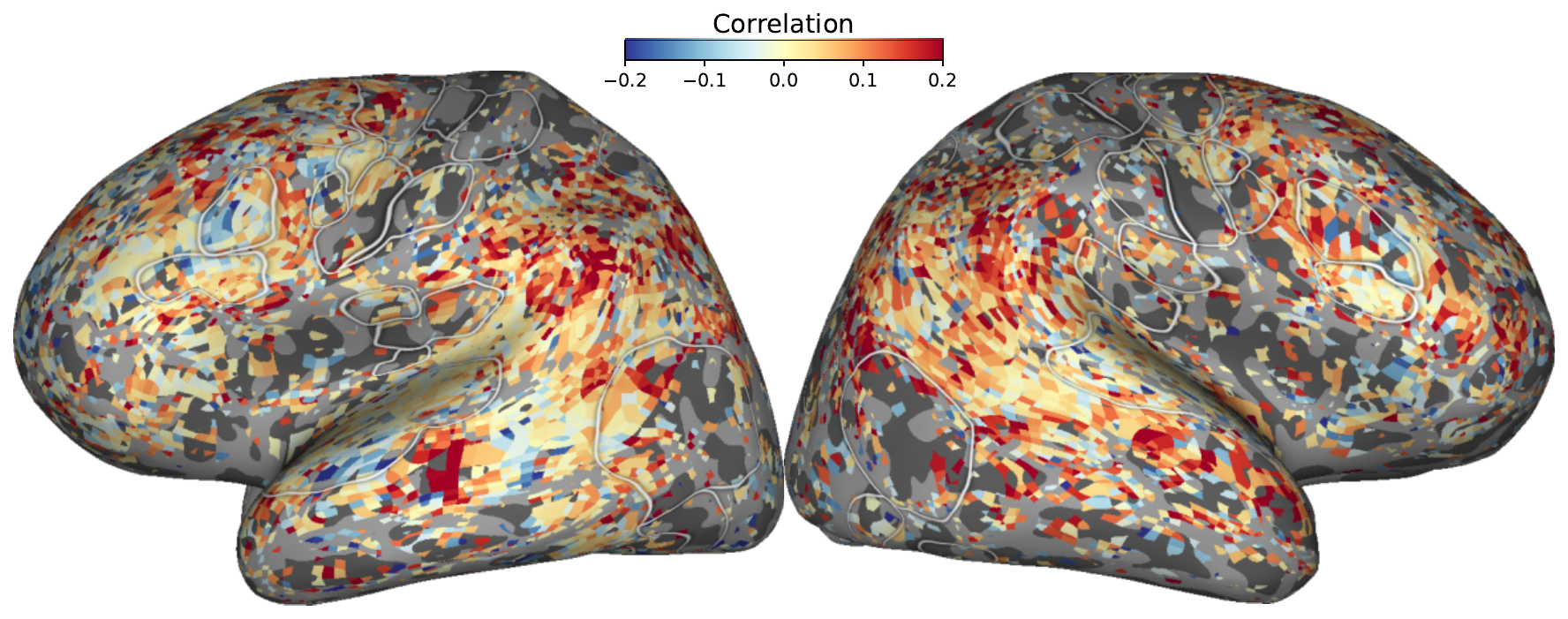}
        \caption{Subject 1.}
    \end{subfigure}
    \hfill
    \begin{subfigure}{0.45\linewidth}
        \centering
        \includegraphics[width=\linewidth]{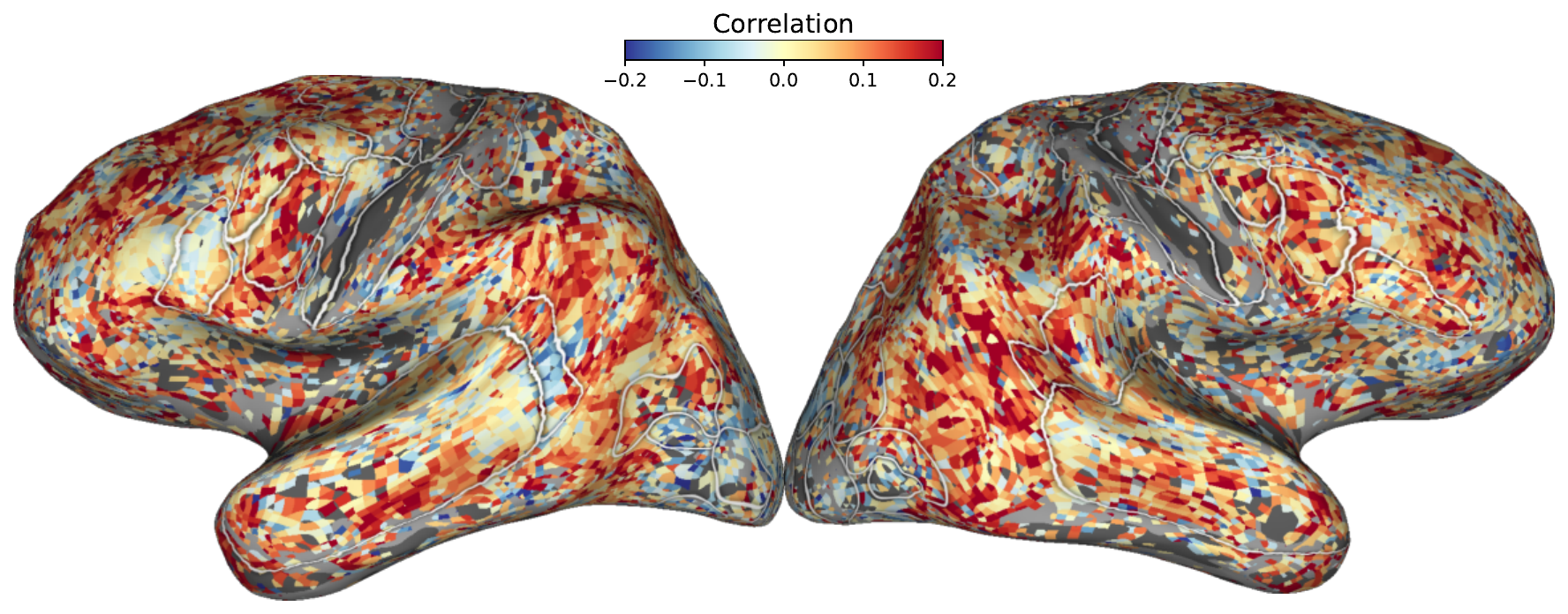}
        \caption{Subject 2.}
    \end{subfigure}

    \begin{subfigure}{0.45\linewidth}
        \centering
        \includegraphics[width=\linewidth]{data/images/appendix/B_results/3_brain/SCRIBE_c_vs_o/diff_inflated_SCRIBE-c_minus_SCRIBE-o_subject_3.pdf}
        \caption{Subject 3.}
    \end{subfigure}
    \hfill
    \begin{subfigure}{0.45\linewidth}
        \centering
        \includegraphics[width=\linewidth]{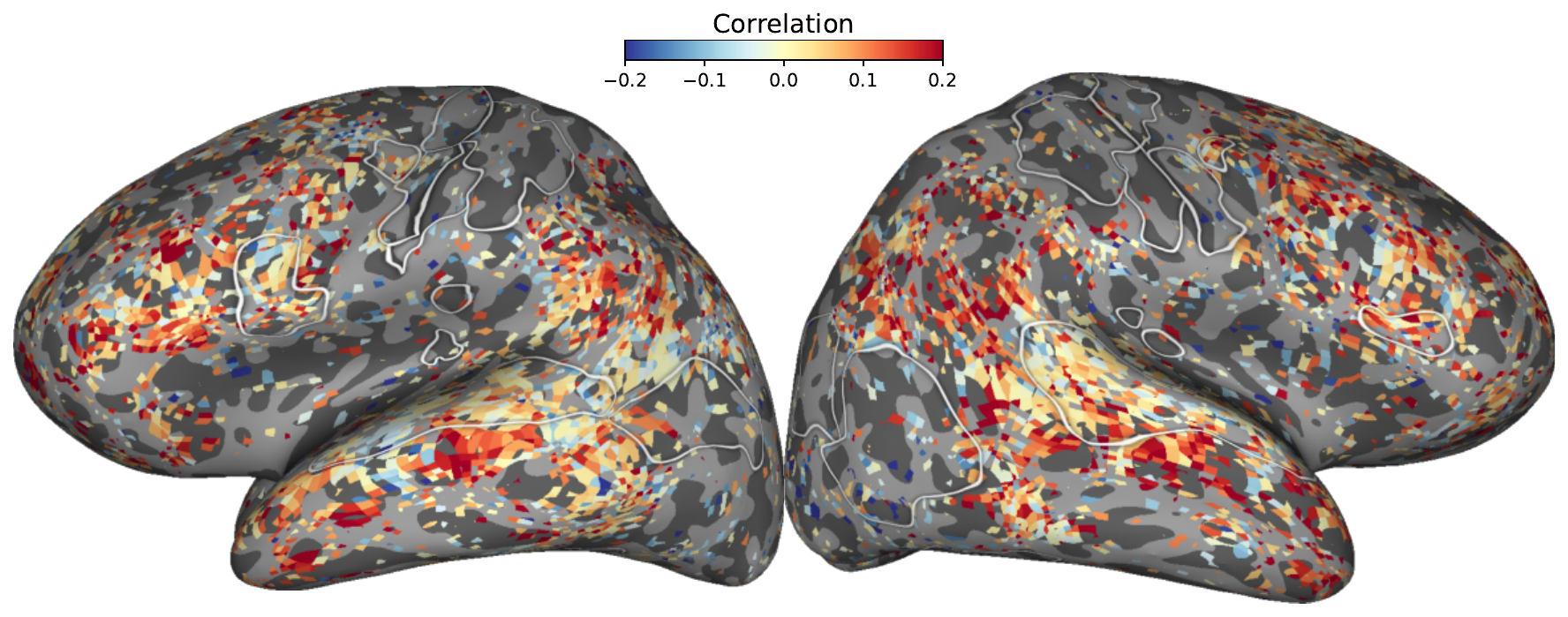}
        \caption{Subject 5.}
    \end{subfigure}

    \begin{subfigure}{0.45\linewidth}
        \centering
        \includegraphics[width=\linewidth]{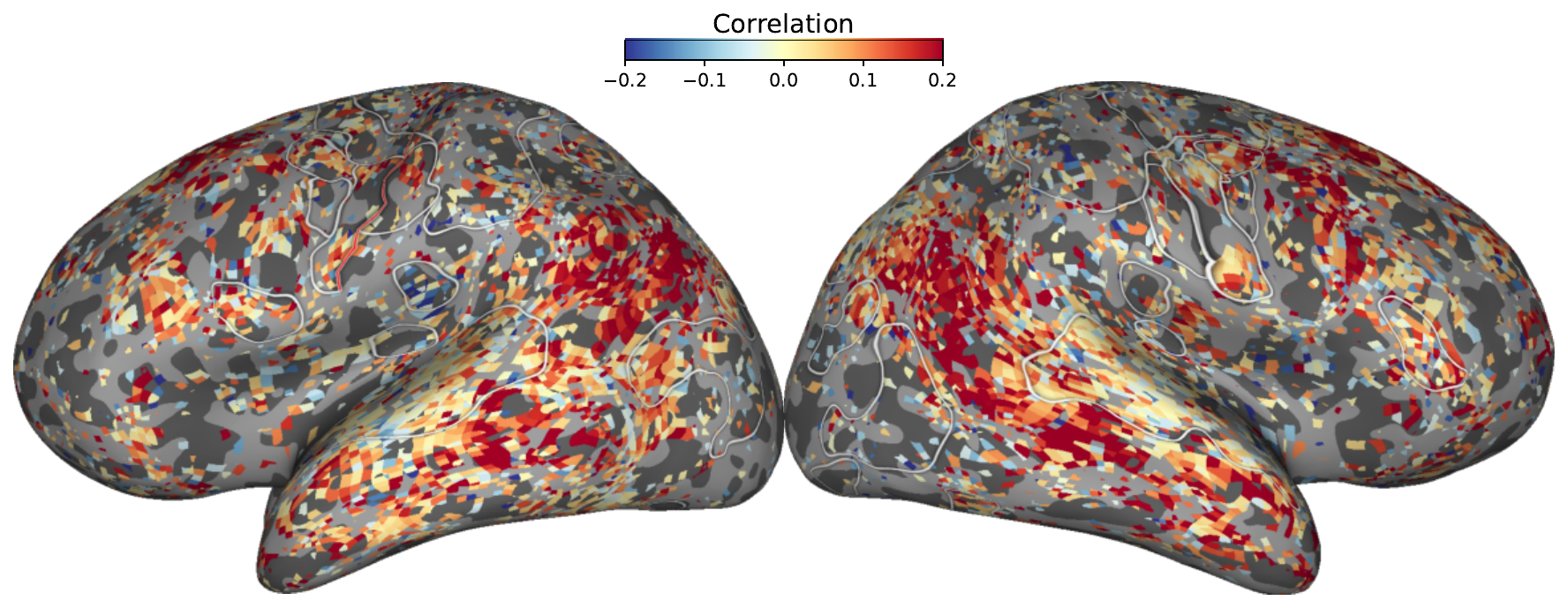}
        \caption{Subject 6.}
    \end{subfigure}
    \hfill
    \begin{subfigure}{0.45\linewidth}
        \centering
        \includegraphics[width=\linewidth]{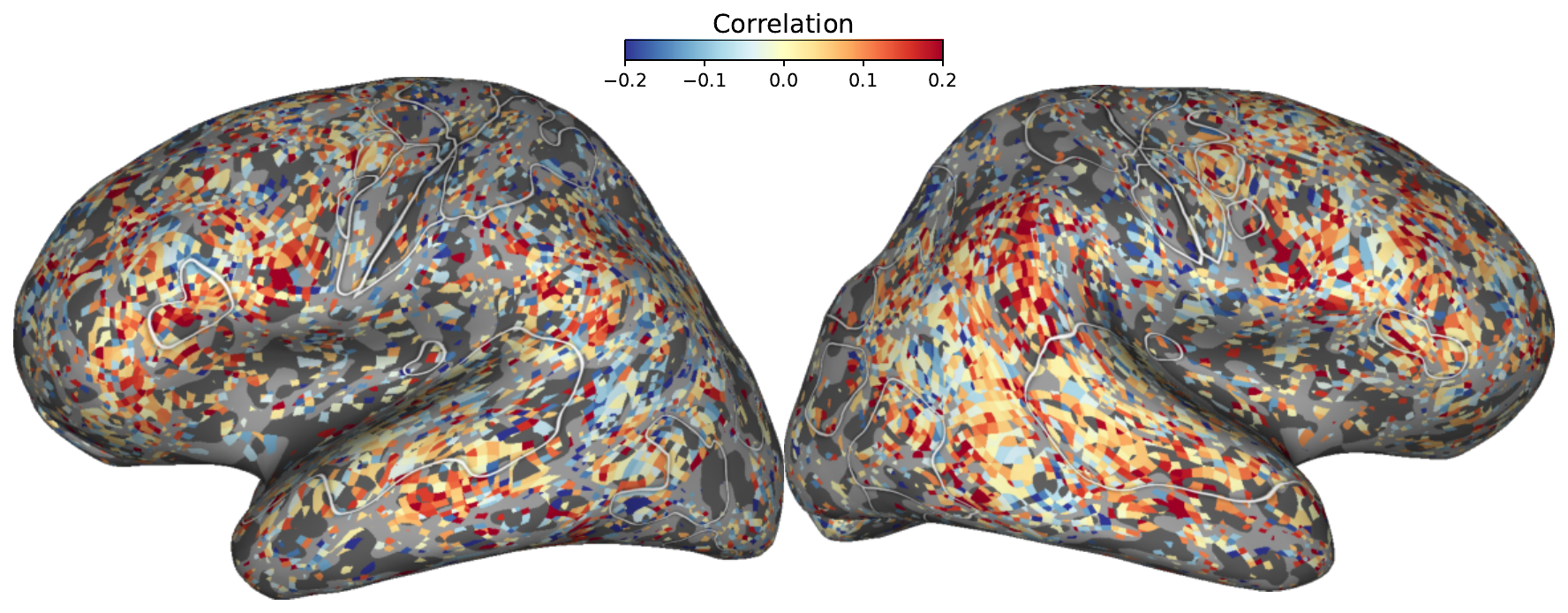}
        \caption{Subject 7.}
    \end{subfigure}

    \begin{subfigure}{0.45\linewidth}
        \centering
        \includegraphics[width=\linewidth]{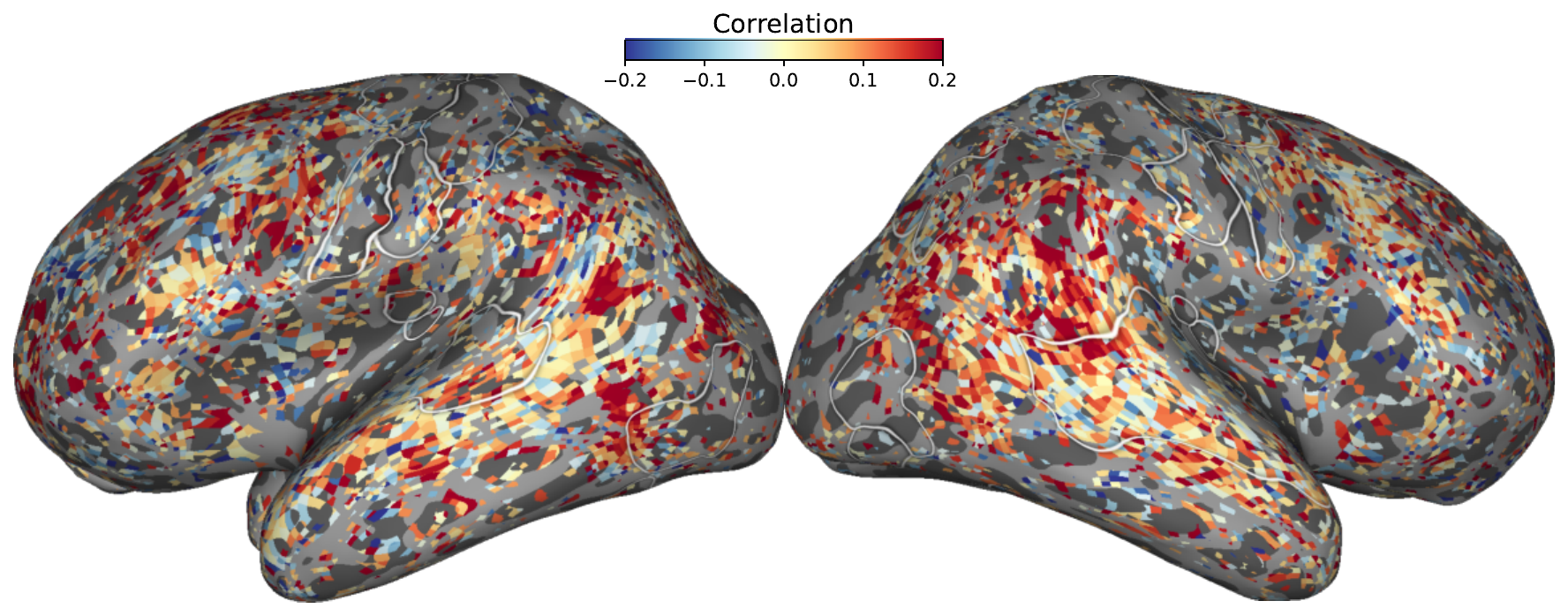}
        \caption{Subject 8.}
    \end{subfigure}

    \caption{
    Voxel-wise differences in brain-prediction accuracy between contextual and out-of-context \method{} representations ($\mathrm{SCRIBE}_{C} - \mathrm{SCRIBE}_{O}$) across subjects.
    Red regions indicate greater prediction accuracy for $\mathrm{SCRIBE}_{C}$, whereas blue regions indicate greater accuracy for $\mathrm{SCRIBE}_{O}$.
    Contextualization-related improvements are predominantly observed in language-related cortical regions, with stronger effects in the right hemisphere.
    }
    \label{fig:appendix_diff_brain_plots_nc_vs_c_allsubjects}
\end{figure}

To visualize the spatial effect of contextualization, we computed voxel-wise differences in brain-prediction accuracy between $\mathrm{SCRIBE}_{C}$ and $\mathrm{SCRIBE}_{O}$.
Figure~\ref{fig:appendix_diff_brain_plots_nc_vs_c_allsubjects} shows these difference maps for each subject, with positive values indicating greater prediction accuracy for the contextual representation.

\clearpage
\subsubsection{Complementarity between QAEmb and \method{}}
\label{appendix:qaemb_and_SCRIBE}

Although $\mathrm{SCRIBE}_\mathrm{C}$ and QAEmb achieve comparable brain-encoding performance individually, an important question is whether the two representations capture redundant semantic information or provide complementary information about neural responses.
We investigated this in two ways: first, by quantifying the unique variance explained by each representation through variance partitioning; and second, by testing whether combining their feature spaces improves held-out brain prediction.

\begin{figure}[h]
    \centering

    \begin{subfigure}{\linewidth}
        \centering
        \includegraphics[width=\linewidth]{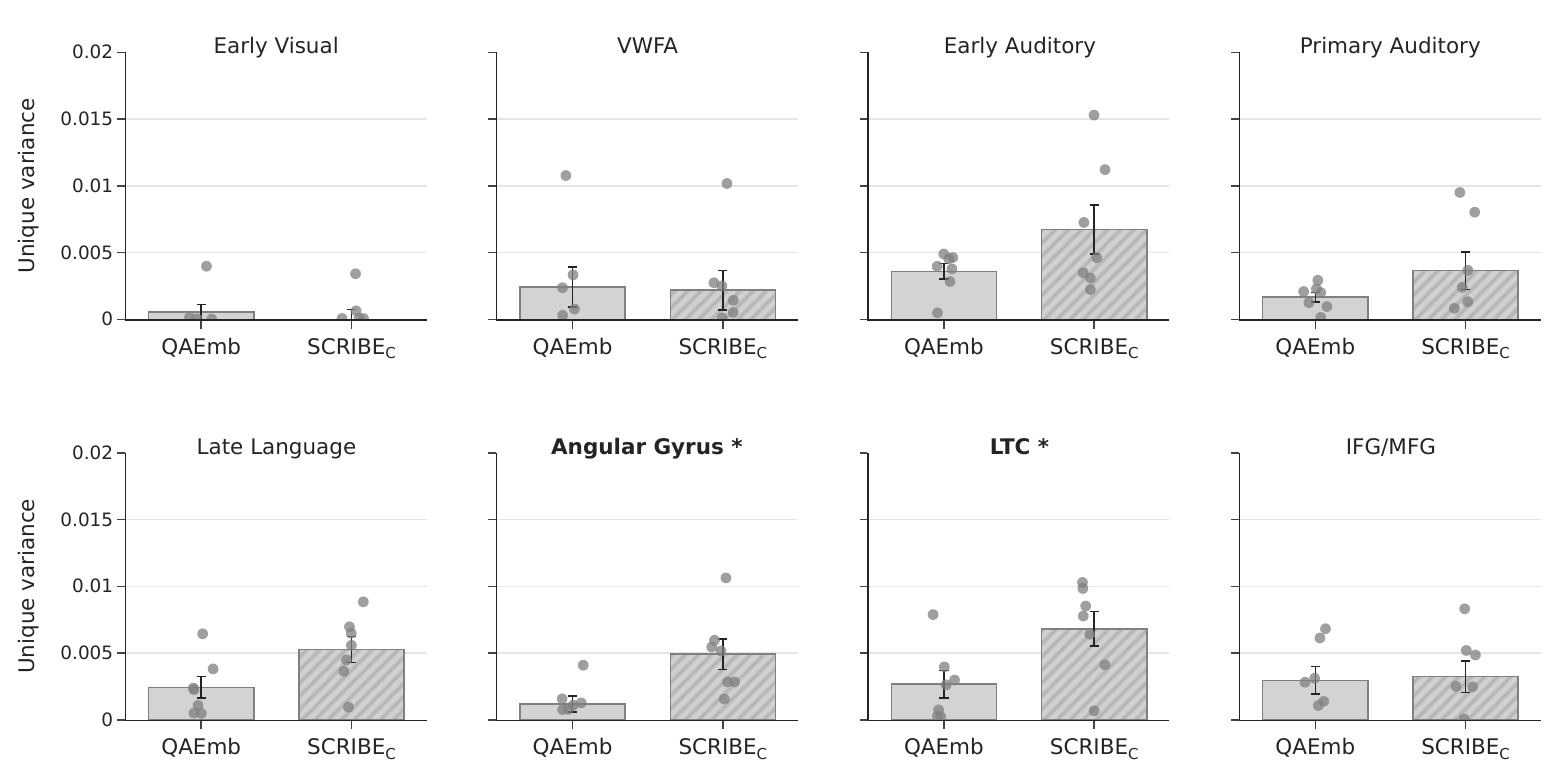}
    \end{subfigure}

    \caption{
    Unique variance explained by $\mathrm{SCRIBE}_{C}$ and QAEmb across cortical regions.
    Bars show the mean unique explained variance across subjects, with error bars indicating the standard error of the mean; individual dots represent subjects.
    $p$-values correspond to one-sided paired Wilcoxon signed-rank tests testing whether the unique variance explained by $\mathrm{SCRIBE}_{C}$ is greater than that explained by QAEmb.
    Asterisks indicate significant differences ($p<0.05$).
    }
    \label{fig:appendix_scribe_qaemb_uniquevar_all_rois}
\end{figure}

\begin{figure}[h]
    \centering
    \begin{subfigure}{0.45\linewidth}
        \centering
        \includegraphics[width=\linewidth]{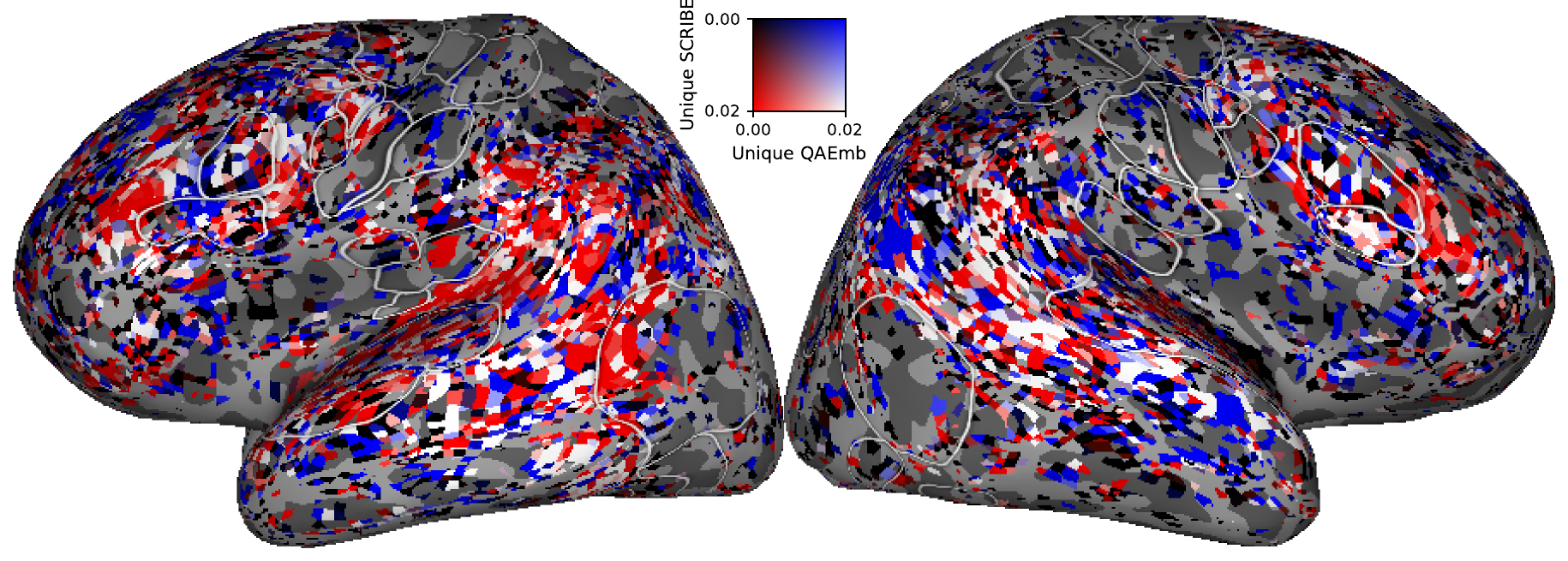}
        \caption{Subject 1.}
    \end{subfigure}
    \hfill
    \begin{subfigure}{0.45\linewidth}
        \centering
        \includegraphics[width=\linewidth]{data/images/appendix/B_results/3_brain/SCRIBE_vs_QA/contrast_inflated_QAvsSCRIBE_subject_2_VP.pdf}
        \caption{Subject 2.}
    \end{subfigure}

    \begin{subfigure}{0.45\linewidth}
        \centering
        \includegraphics[width=\linewidth]{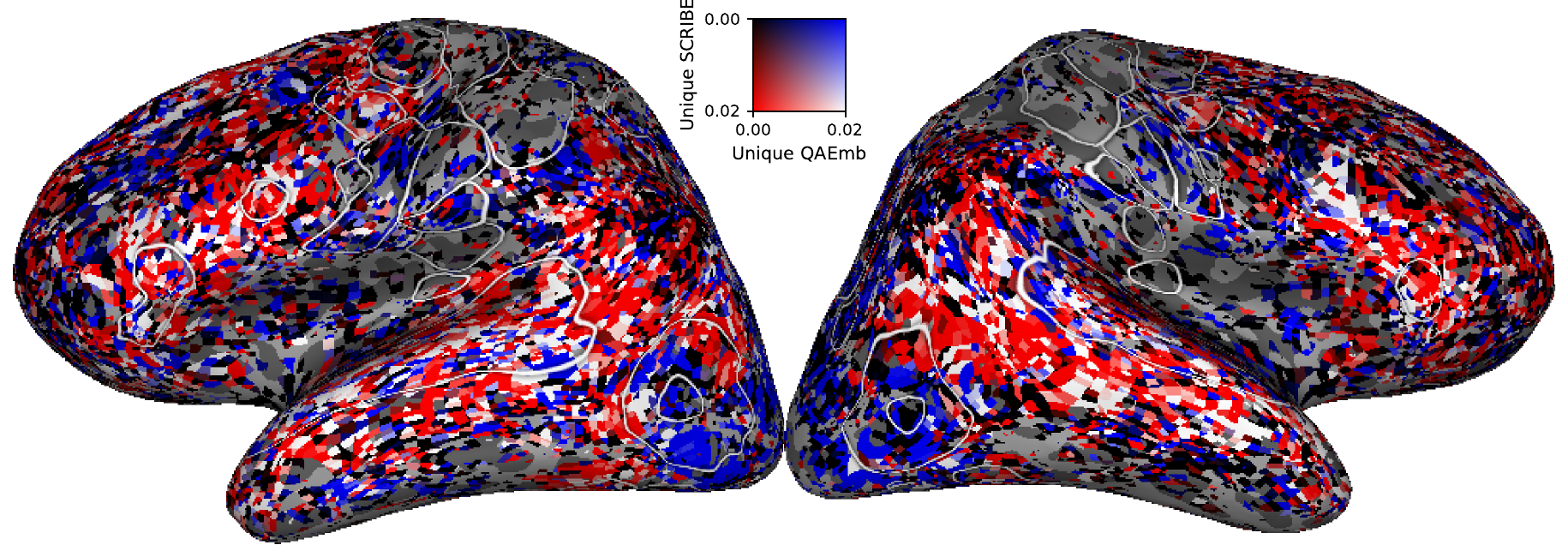}
        \caption{Subject 3.}
    \end{subfigure}
    \hfill
    \begin{subfigure}{0.45\linewidth}
        \centering
        \includegraphics[width=\linewidth]{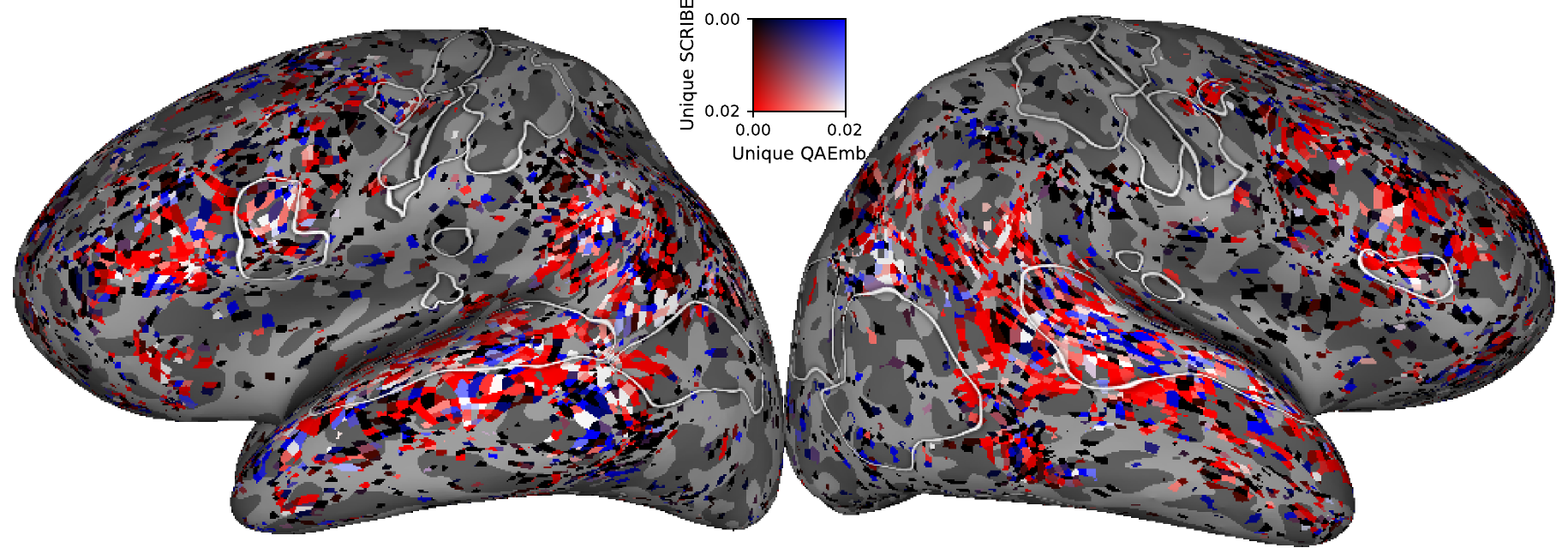}
        \caption{Subject 5.}
    \end{subfigure}
    
    \begin{subfigure}{0.45\linewidth}
        \centering
        \includegraphics[width=\linewidth]{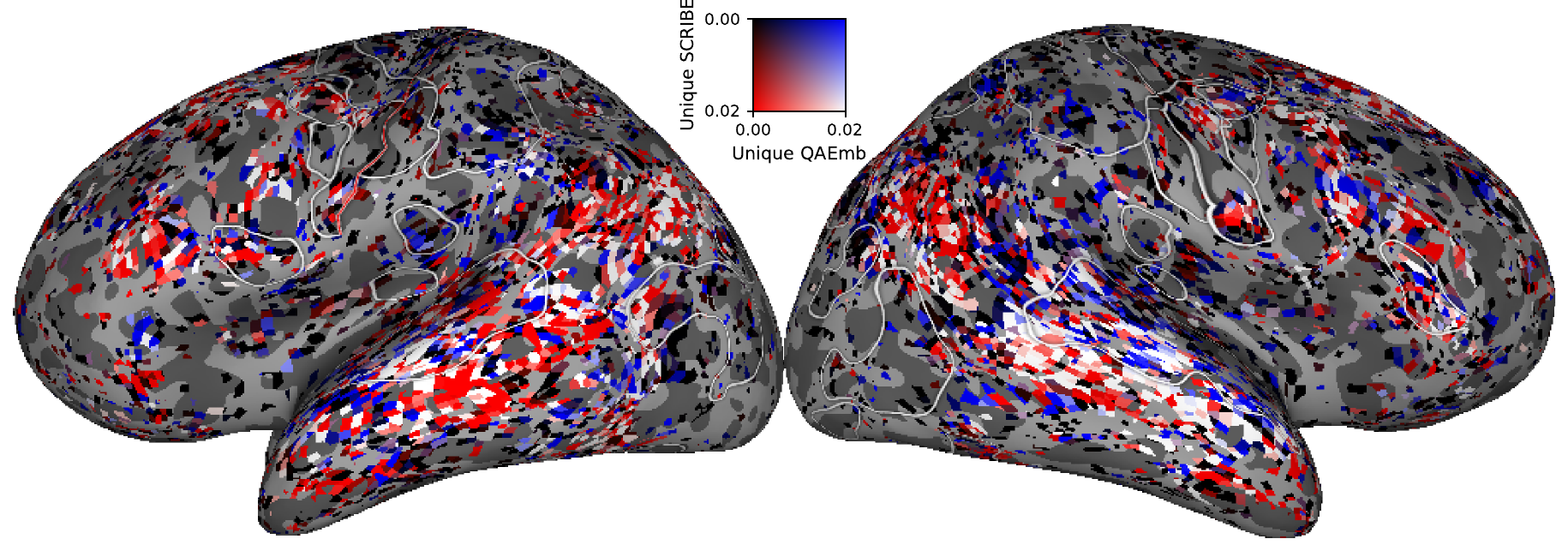}
        \caption{Subject 6.}
    \end{subfigure}
    \hfill
    \begin{subfigure}{0.45\linewidth}
        \centering
        \includegraphics[width=\linewidth]{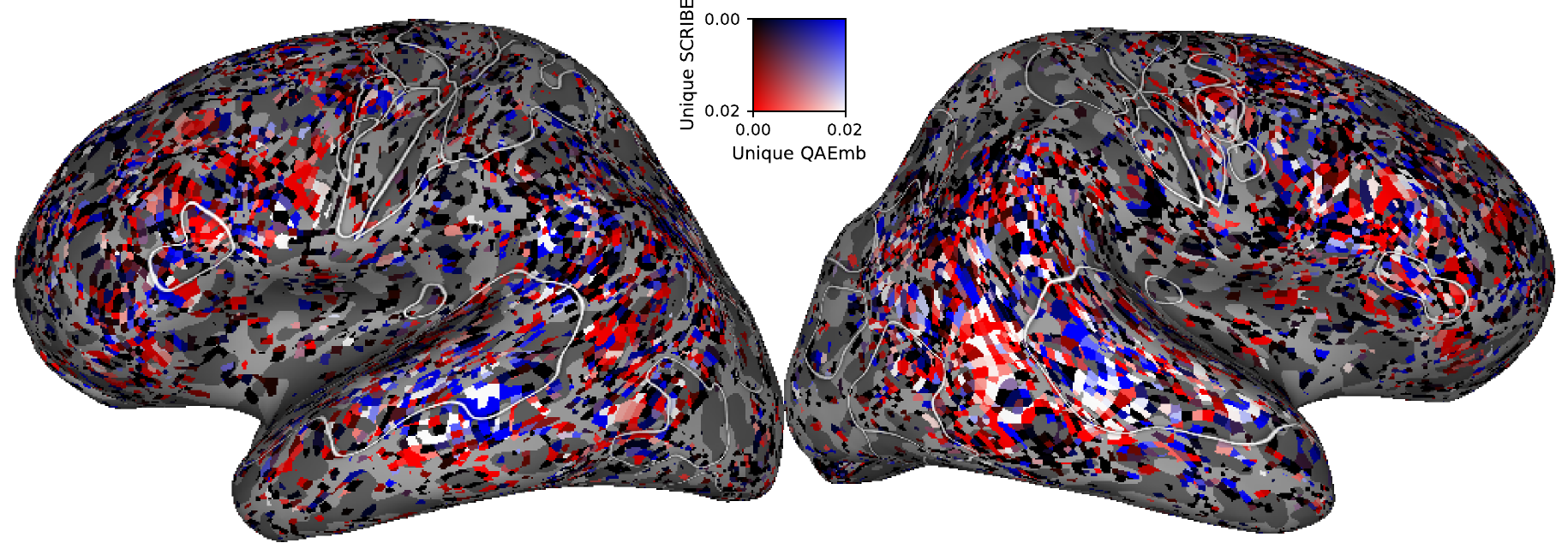}
        \caption{Subject 7.}
    \end{subfigure}

    \begin{subfigure}{0.45\linewidth}
        \centering
        \includegraphics[width=\linewidth]{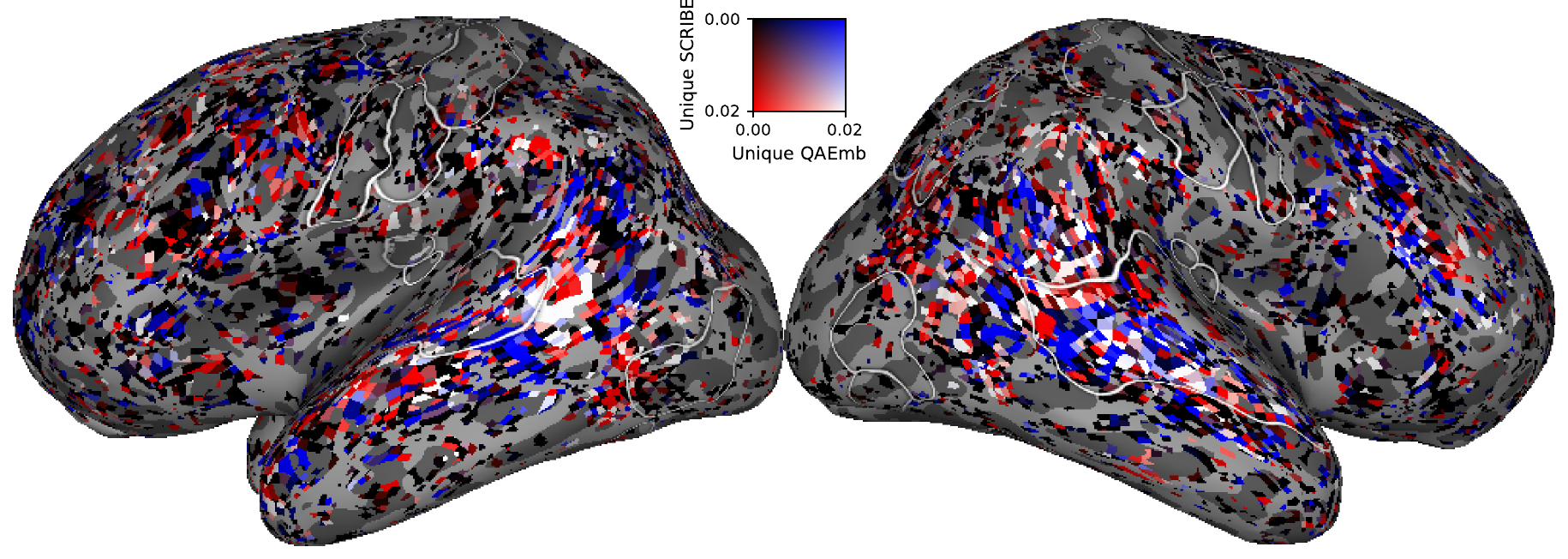}
        \caption{Subject 8.}
    \end{subfigure}
    \caption{
    Voxel-wise unique variance explained by \method{} and QAEmb across subjects.
    Variance partitioning was performed directly between the two representations, without including the syntactic control features.
    Red regions indicate greater unique variance explained by \method{}, whereas blue regions indicate greater unique variance explained by QAEmb.
    Only voxels with a noise ceiling above $0.4$ are shown.
    Across subjects, the maps predominantly show greater unique variance for \method{}, particularly in language-related cortical areas, although this pattern is less pronounced in Subject 8.
    }
    \label{fig:contrastive_brain_plots_qa_vs_scribe}
\end{figure}

Variance partitioning showed that both $\mathrm{SCRIBE}_\mathrm{C}$ and QAEmb explain unique variance in neural responses beyond their shared contribution (Figure~\ref{fig:appendix_scribe_qaemb_uniquevar_all_rois}).
Despite their similar overall encoding performance, $\mathrm{SCRIBE}_\mathrm{C}$ explained more unique variance than QAEmb across several higher-level language and semantic ROIs, with significant differences in the lateral temporal cortex (LTC; $p=0.039$) and angular gyrus (AG; $p=0.023$), and a numerical difference in the late-language network.
These results indicate that the two representations are not fully redundant and that $\mathrm{SCRIBE}_\mathrm{C}$ captures brain-predictive information not accounted for by QAEmb.
To examine the spatial distribution of these unique contributions, Figure~\ref{fig:contrastive_brain_plots_qa_vs_scribe} shows the voxel-wise unique-variance contrast between \method{} and QAEmb for each subject.

To further test whether this complementary information translates into improved brain prediction, we trained an additional encoding model using the concatenation of the $\mathrm{SCRIBE}_\mathrm{C}$ and QAEmb feature spaces and compared its prediction performance with models using either representation alone (Table~\ref{tab:joint_scribe_qaemb_results}).
At the whole-brain level, combining QAEmb with $\mathrm{SCRIBE}_\mathrm{C}$ significantly improved prediction performance over both $\mathrm{SCRIBE}_\mathrm{C}$ and QAEmb individually.
The combined representation achieved a mean noise-ceiling-normalized correlation of $0.2022 \pm 0.0146$, compared with $0.1837 \pm 0.0144$ for $\mathrm{SCRIBE}_\mathrm{C}$ and $0.1778 \pm 0.0125$ for QAEmb (one-sided Wilcoxon signed-rank tests, $p=0.0078$ for both comparisons).
This corresponds to a $13.7\%$ improvement in mean correlation over QAEmb alone.

The same pattern was observed within cortical regions associated with higher-level language and semantic processing.
The combined representation yielded higher mean prediction performance than either individual representation in the late-language network ($0.1577 \pm 0.0213$), LTC ($0.1644 \pm 0.0236$), IFG/MFG ($0.1458 \pm 0.0196$), and AG ($0.1497 \pm 0.0290$).
The combined representation significantly outperformed both individual representations in all four ROIs (one-sided Wilcoxon signed-rank tests, $p<0.05$; Table~\ref{tab:joint_scribe_qaemb_results}).

Together, the variance-partitioning and joint-model analyses provide converging evidence that $\mathrm{SCRIBE}_\mathrm{C}$ and QAEmb capture complementary brain-relevant information.
Both representations account for unique variance in neural responses, while $\mathrm{SCRIBE}_\mathrm{C}$ shows greater unique contributions in several higher-level language and semantic regions.
Moreover, combining the two representations consistently improves held-out prediction performance over either representation alone, indicating that the information captured by \method{} is not subsumed by QAEmb.

\begin{table}[h]
    \centering
    \caption{
    Noise-ceiling-normalized brain encoding correlations for \method{} ($\mathrm{SCRIBE}_{C}$), QAEmb,
    and their combined representation, evaluated across the whole brain and
    semantic language ROIs.
    Values are reported for individual subjects and as mean $\pm$ standard error across subjects.
    $\ast$ denotes that QAEmb + $\mathrm{SCRIBE}_\mathrm{C}$ significantly outperforms both individual representations
    (one-sided Wilcoxon signed-rank tests, $p<0.05$).
    }
    \scriptsize{
    \begin{tabular}{llcccccccc}
    \textbf{Region} & \textbf{Method}
    & \textbf{S1} & \textbf{S2} & \textbf{S3}
    & \textbf{S5} & \textbf{S6} & \textbf{S7} & \textbf{S8}
    & \textbf{Avg $\pm$ STE} \\
    \midrule

    \multirow{3}{*}{Whole Brain}
    & $\mathrm{SCRIBE}_\mathrm{C}$
    & 0.2284 & 0.2263 & 0.1714 & 0.2023 & 0.1800 & 0.1519 & 0.1258
    & 0.1837 $\pm$ 0.0144 \\
    & QAEmb
    & 0.2343 & 0.2115 & 0.1626 & 0.1788 & 0.1554 & 0.1544 & 0.1477
    & 0.1778 $\pm$ 0.0125 \\
    & \textbf{QAEmb + $\mathrm{SCRIBE}_\mathrm{C}$}
    & 0.2579 & 0.2426 & 0.1895 & 0.2140 & 0.1926 & 0.1684 & 0.1504
    & \textbf{0.2022 $\pm$ 0.0146}$^\ast$ \\

    \midrule

    \multirow{3}{*}{Late Language}
    & $\mathrm{SCRIBE}_\mathrm{C}$
    & 0.1673 & 0.1990 & 0.2112 & 0.1348 & 0.1371 & 0.1191 & 0.0536
    & 0.1460 $\pm$ 0.0201 \\
    & QAEmb
    & 0.1715 & 0.1943 & 0.1932 & 0.1136 & 0.1214 & 0.1153 & 0.0615
    & 0.1387 $\pm$ 0.0186 \\
    & \textbf{QAEmb + $\mathrm{SCRIBE}_\mathrm{C}$}
    & 0.1880 & 0.2145 & 0.2253 & 0.1395 & 0.1450 & 0.1285 & 0.0632
    & \textbf{0.1577 $\pm$ 0.0213}$^\ast$ \\

    \midrule

    \multirow{3}{*}{LTC}
    & $\mathrm{SCRIBE}_\mathrm{C}$
    & 0.1991 & 0.2046 & 0.1770 & 0.1609 & 0.1819 & 0.1155 & 0.0377
    & 0.1538 $\pm$ 0.0223 \\
    & QAEmb
    & 0.2026 & 0.1963 & 0.1569 & 0.1405 & 0.1571 & 0.1147 & 0.0415
    & 0.1442 $\pm$ 0.0206 \\
    & \textbf{QAEmb + $\mathrm{SCRIBE}_\mathrm{C}$}
    & 0.2218 & 0.2174 & 0.1870 & 0.1665 & 0.1884 & 0.1258 & 0.0435
    & \textbf{0.1644 $\pm$ 0.0236}$^\ast$ \\

    \midrule

    \multirow{3}{*}{IFG/MFG}
    & $\mathrm{SCRIBE}_\mathrm{C}$
    & 0.1819 & 0.1424 & 0.1948 & 0.1200 & 0.0769 & 0.1350 & 0.0673
    & 0.1312 $\pm$ 0.0182 \\
    & QAEmb
    & 0.1901 & 0.1651 & 0.1853 & 0.0989 & 0.0726 & 0.1251 & 0.0948
    & 0.1331 $\pm$ 0.0178 \\
    & \textbf{QAEmb + $\mathrm{SCRIBE}_\mathrm{C}$}
    & 0.2060 & 0.1690 & 0.2104 & 0.1227 & 0.0837 & 0.1422 & 0.0870
    & \textbf{0.1458 $\pm$ 0.0196}$^\ast$ \\

    \midrule

    \multirow{3}{*}{Angular Gyrus}
    & $\mathrm{SCRIBE}_\mathrm{C}$
    & 0.0926 & 0.2294 & 0.2583 & 0.1236 & 0.1099 & 0.1011 & 0.0648
    & 0.1399 $\pm$ 0.0279 \\
    & QAEmb
    & 0.0922 & 0.2058 & 0.2365 & 0.1006 & 0.0905 & 0.1016 & 0.0602
    & 0.1268 $\pm$ 0.0251 \\
    & \textbf{QAEmb + $\mathrm{SCRIBE}_\mathrm{C}$}
    & 0.1063 & 0.2378 & 0.2771 & 0.1296 & 0.1156 & 0.1118 & 0.0698
    & \textbf{0.1497 $\pm$ 0.0290}$^\ast$ \\

    \bottomrule
    \end{tabular}
    }
    \label{tab:joint_scribe_qaemb_results}
\end{table}

\clearpage
\subsubsection{Controlling for feature-space dimensionality}
\label{appendix:feature_selection}

$\mathrm{SCRIBE}_\mathrm{C}$ and QAEmb differ in the dimensionality of their final feature spaces:
$\mathrm{SCRIBE}_\mathrm{C}$ contains 65 semantic properties, whereas QAEmb uses 35 features selected from an initial set of 606 questions.
We therefore investigated whether differences in brain-encoding performance could be explained simply by the larger dimensionality of $\mathrm{SCRIBE}_\mathrm{C}$.

To control for this possibility, we applied sparse feature selection to our contextual variant of $\mathrm{SCRIBE}_\mathrm{C}$, following the feature-selection approach used by QAEmb.
Feature selection retained 33 of the original 65 semantic dimensions, resulting in a representation with approximately half the dimensionality of the original $\mathrm{SCRIBE}_\mathrm{C}$ space and a dimensionality comparable to QAEmb.
The selected dimensions were: \textit{Bright, Dark, Large, Small, Fast, Slow, Face, Temperature, Pain, Weight, Audition, High, Music, Taste, Smell, Head, UpperLimb, LowerLimb, Landmark, Toward, Number, Time, Long, Self, Cognition, Happy, Sad, Angry, Disgusted, Fearful, Surprised, Needs}, and \textit{Arousal}.

Despite reducing the representation from 65 to 33 dimensions, whole-brain encoding performance was preserved (Table~\ref{tab:scribe_feature_selection}).
The full $\mathrm{SCRIBE}_\mathrm{C}$ representation achieved a mean noise-ceiling-normalized correlation of $0.1837 \pm 0.0144$, compared with $0.1859 \pm 0.0126$ after feature selection.
The difference was not statistically significant (two-sided Wilcoxon signed-rank test, $p=0.5781$).

This preservation of performance was also observed across the examined language and semantic ROIs.
The full and feature-selected representations achieved highly similar correlations in the late-language network ($0.1460 \pm 0.0201$ vs.\ $0.1452 \pm 0.0197$), lateral temporal cortex (LTC; $0.1538 \pm 0.0223$ vs.\ $0.1522 \pm 0.0216$), inferior/middle frontal gyrus (IFG/MFG; $0.1312 \pm 0.0182$ vs.\ $0.1344 \pm 0.0181$), and angular gyrus (AG; $0.1399 \pm 0.0279$ vs.\ $0.1380 \pm 0.0268$; Table~\ref{tab:scribe_feature_selection}).

Together, these results indicate that $\mathrm{SCRIBE}_\mathrm{C}$ does not depend on its larger number of semantic dimensions to achieve its brain-encoding performance.
Reducing the representation from 65 to 33 dimensions preserves both whole-brain prediction accuracy and performance across the examined semantic-language ROIs, suggesting that the observed results are not simply attributable to differences in feature-space dimensionality between $\mathrm{SCRIBE}_\mathrm{C}$ and QAEmb.

\begin{table}[h]
    \centering
    \caption{
    Noise-ceiling-normalized brain encoding correlations for $\mathrm{SCRIBE}_\mathrm{C}$ before and after feature selection, evaluated across the whole brain and semantic language ROIs.
    Feature selection reduces the representation from 65 to 33 semantic dimensions.
    Values are reported for individual subjects and as mean $\pm$ standard error across subjects.
    At the whole-brain level, performance does not significantly differ after feature selection
    (two-sided Wilcoxon signed-rank test).
    }
    \scriptsize{
    \begin{tabular}{llcccccccc}
    \textbf{Region} & \textbf{Method}
    & \textbf{S1} & \textbf{S2} & \textbf{S3}
    & \textbf{S5} & \textbf{S6} & \textbf{S7} & \textbf{S8}
    & \textbf{Avg $\pm$ STE} \\
    \midrule

    \multirow{2}{*}{Whole Brain}
    & $\mathrm{SCRIBE}_\mathrm{C}$ (65)
    & 0.2284 & 0.2263 & 0.1714 & 0.2023
    & 0.1800 & 0.1519 & 0.1258
    & 0.1837 $\pm$ 0.0144 \\
    & $\mathrm{SCRIBE}_\mathrm{C}$ (33)
    & 0.2275 & 0.2101 & 0.1848 & 0.2046
    & 0.1897 & 0.1504 & 0.1340
    & 0.1859 $\pm$ 0.0126 \\

    \midrule

    \multirow{2}{*}{Late Language}
    & $\mathrm{SCRIBE}_\mathrm{C}$ (65)
    & 0.1673 & 0.1990 & 0.2112 & 0.1348
    & 0.1371 & 0.1191 & 0.0536
    & 0.1460 $\pm$ 0.0201 \\
    & $\mathrm{SCRIBE}_\mathrm{C}$ (33)
    & 0.1660 & 0.1897 & 0.2124 & 0.1351
    & 0.1396 & 0.1178 & 0.0561
    & 0.1452 $\pm$ 0.0197 \\

    \midrule

    \multirow{2}{*}{LTC}
    & $\mathrm{SCRIBE}_\mathrm{C}$ (65)
    & 0.1991 & 0.2046 & 0.1770 & 0.1609
    & 0.1819 & 0.1155 & 0.0377
    & 0.1538 $\pm$ 0.0223 \\
    & $\mathrm{SCRIBE}_\mathrm{C}$ (33)
    & 0.1938 & 0.1974 & 0.1762 & 0.1615
    & 0.1814 & 0.1146 & 0.0402
    & 0.1522 $\pm$ 0.0216 \\

    \midrule

    \multirow{2}{*}{IFG/MFG}
    & $\mathrm{SCRIBE}_\mathrm{C}$ (65)
    & 0.1819 & 0.1424 & 0.1948 & 0.1200
    & 0.0769 & 0.1350 & 0.0673
    & 0.1312 $\pm$ 0.0182 \\
    & $\mathrm{SCRIBE}_\mathrm{C}$ (33)
    & 0.1835 & 0.1396 & 0.2056 & 0.1215
    & 0.0876 & 0.1356 & 0.0673
    & 0.1344 $\pm$ 0.0181 \\

    \midrule

    \multirow{2}{*}{Angular Gyrus}
    & $\mathrm{SCRIBE}_\mathrm{C}$ (65)
    & 0.0926 & 0.2294 & 0.2583 & 0.1236
    & 0.1099 & 0.1011 & 0.0648
    & 0.1399 $\pm$ 0.0279 \\
    & $\mathrm{SCRIBE}_\mathrm{C}$ (33)
    & 0.0954 & 0.2148 & 0.2567 & 0.1224
    & 0.1095 & 0.0981 & 0.0690
    & 0.1380 $\pm$ 0.0268 \\

    \bottomrule
    \end{tabular}
    }
    \label{tab:scribe_feature_selection}
\end{table}

\clearpage
\subsubsection{Variance partitioning with syntactic features}
\label{appendix:results_variance_partitioning}

To further characterize the information underlying the brain-encoding performance of \method{}, we performed variance partitioning with an explicit syntactic feature space.
Syntactic features comprised one-hot encoded part-of-speech tags and dependency length obtained with spaCy~\citep{honnibal2020spacy}, together with constituency tree depth obtained using the Benepar parser~\citep{kitaev2018constituency}.
We compared $\mathrm{SCRIBE}_\mathrm{C}$ with QAEmb to assess both how much of their brain-predictive variance overlaps with these syntactic features and how much remains uniquely explained after accounting for them.

\begin{figure}[h]
    \centering
    \begin{subfigure}{0.45\linewidth}
        \centering
        \includegraphics[width=\linewidth]{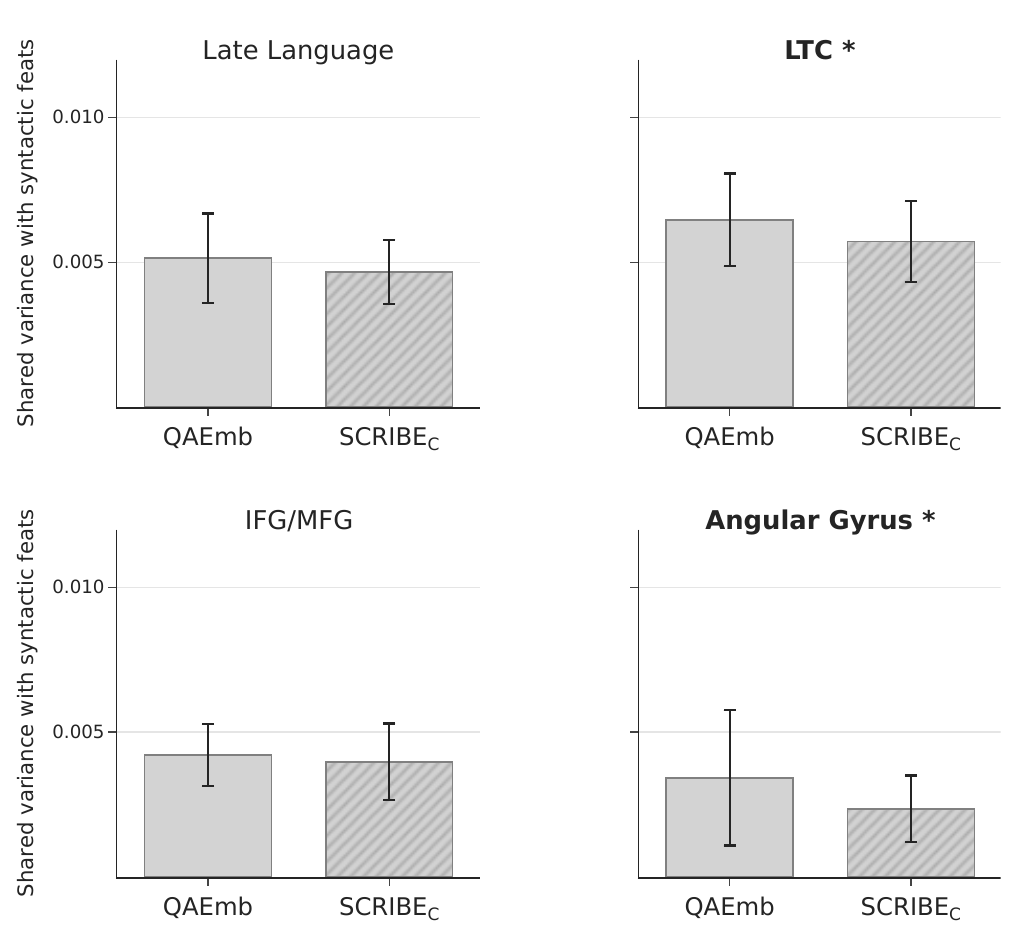}
        \caption{Shared variance with syntactic features.}
        \label{fig:shared_vp_qaemb_vs_scribe_concat-synt}
    \end{subfigure}
    \hfill
    \begin{subfigure}{0.45\linewidth}
        \centering
        \includegraphics[width=\linewidth]{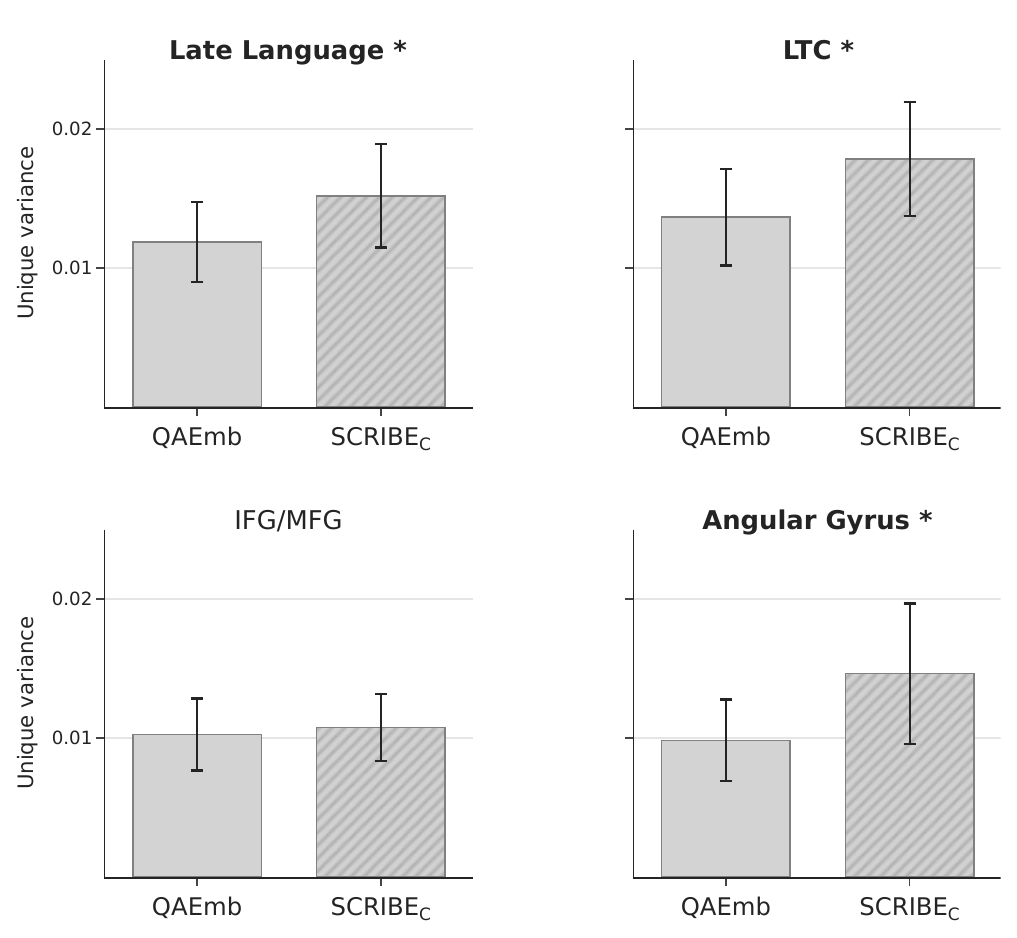}
        \caption{Unique variance beyond syntactic features.}
        \label{fig:unique_vp_qaemb_vs_scribe_concat-synt}
    \end{subfigure}
    \caption{
    Variance partitioning of $\mathrm{SCRIBE}_{C}$ and QAEmb with syntactic features across language-related ROIs.
    (a) Brain-predictive variance shared with the syntactic feature space.
    (b) Unique brain-predictive variance retained by each representation after accounting for the syntactic feature space.
    Bars show means across subjects and error bars indicate the standard error of the mean.
    Bold ROI labels and asterisks indicate significant paired differences between $\mathrm{SCRIBE}_{C}$ and QAEmb (one-sided paired Wilcoxon signed-rank test, $p<0.05$).
    }
    \label{fig:vp_synt_feats_qaemb_vs_scribe}
\end{figure}

Figure~\ref{fig:vp_synt_feats_qaemb_vs_scribe}a shows the variance that $\mathrm{SCRIBE}_\mathrm{C}$ and QAEmb share with the syntactic feature space.
$\mathrm{SCRIBE}_\mathrm{C}$ shares significantly less variance with syntactic features than QAEmb in LTC and AG, with the same numerical direction in Late Language and IFG/MFG.
Thus, the greater brain-prediction accuracy of $\mathrm{SCRIBE}_\mathrm{C}$ relative to QAEmb is not accompanied by greater overlap with the measured syntactic information.

Complementing this result, Figure~\ref{fig:vp_synt_feats_qaemb_vs_scribe}b shows the variance uniquely explained by each representation after accounting for the syntactic feature space.
$\mathrm{SCRIBE}_\mathrm{C}$ retains significantly more unique brain-predictive variance than QAEmb in Late Language, LTC, and AG, with no significant difference in IFG/MFG.

Together, these analyses show that, relative to QAEmb, $\mathrm{SCRIBE}_\mathrm{C}$ retains more unique brain-predictive information while exhibiting less overlap with the measured syntactic features, supporting a more semantically selective interpretation of its neural predictivity.

\begin{figure}[h]
    \centering
    \includegraphics[width=\linewidth]{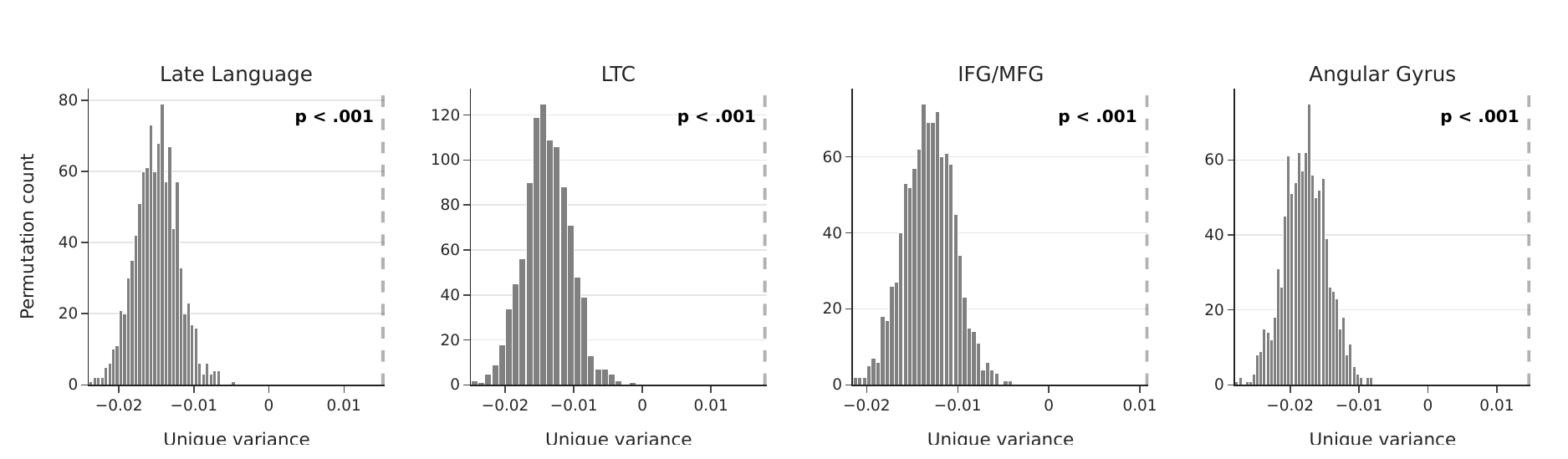}
    \caption{
    Permutation-based significance testing of the unique neural variance explained by $\mathrm{SCRIBE}_\mathrm{C}$ after accounting for syntactic features.
    Histograms show permutation-based null distributions, and dashed lines indicate the observed unique variance.
    }
    \label{fig:permuted_unique_vp_scribe_syntax}
\end{figure}

\clearpage
\subsubsection{Comparison with LLM Embeddings}
\label{appendix:llm_embeddings_vs_SCRIBE}

To compare \method{} with representations extracted directly from an LLM, we evaluated brain-encoding performance using Llama2-7B embeddings. 
We consider both non-contextual embeddings extracted from layer 0 and contextual embeddings extracted from layer 15.

\begin{figure}[h]
    \centering

    \begin{subfigure}[c]{0.42\linewidth}
        \centering
        \includegraphics[width=\linewidth]{
        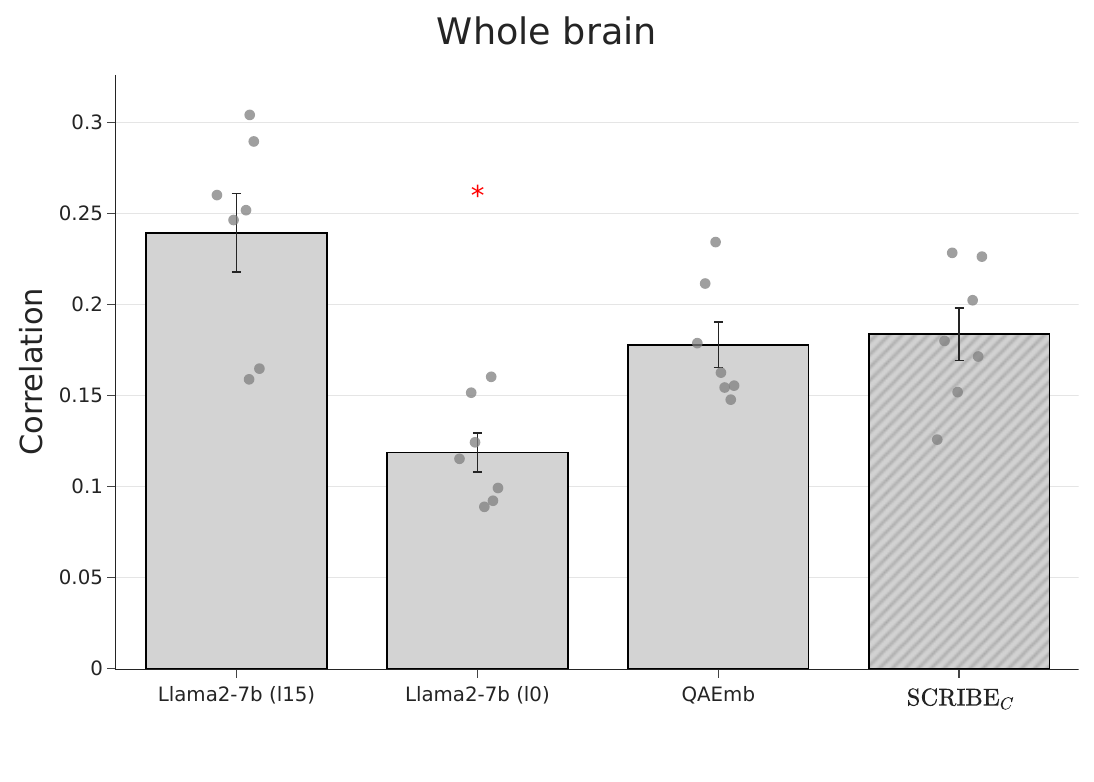}
        \caption{Whole-brain encoding performance.}
        \label{fig:appendix_wholebrain_scribe_llama_qaemb}
    \end{subfigure}
    \hfill
    \begin{subfigure}[c]{0.55\linewidth}
        \centering
        \includegraphics[width=\linewidth]{
        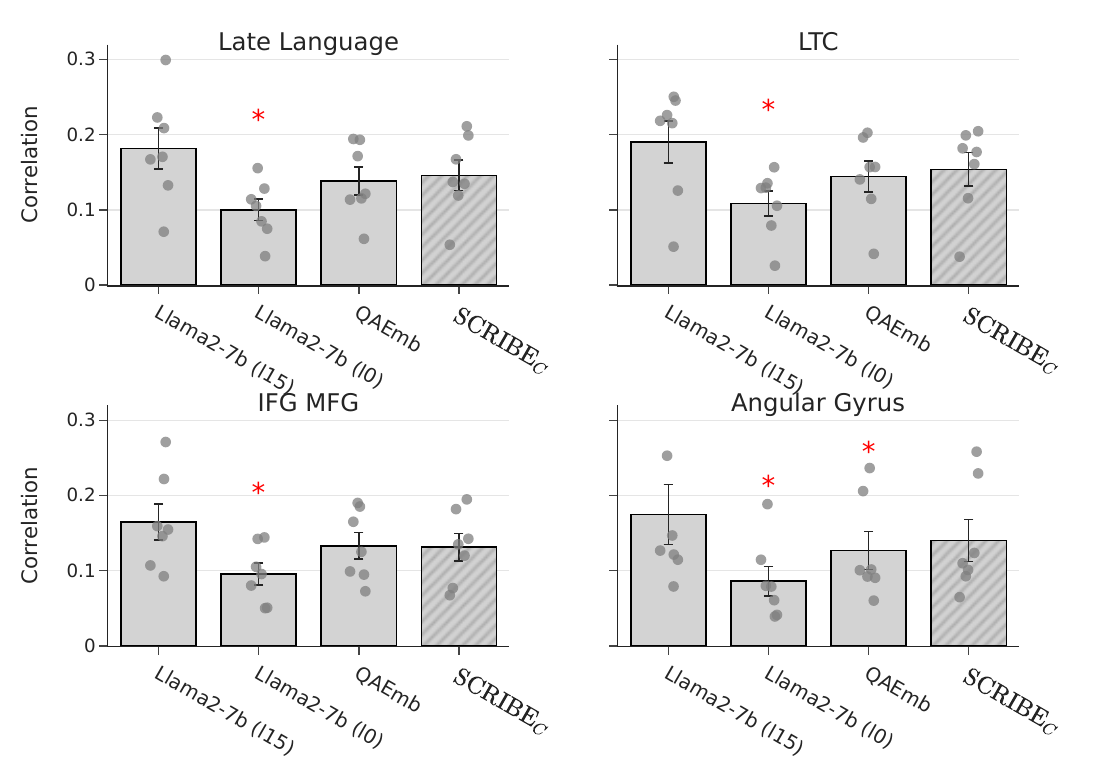}
        \caption{Encoding performance across language-related ROIs.}
        \label{fig:appendix_rois_scribe_llama_qaemb}
    \end{subfigure}

    \caption{
    Brain-encoding performance of Llama2-7b, $\mathrm{QAEmb}$, and 
    $\mathrm{SCRIBE}_{C}$.
    \textbf{(a)} Whole-brain encoding performance.
    \textbf{(b)} Encoding performance across the Late Language, lateral temporal cortex (LTC), IFG/MFG, and Angular Gyrus ROIs.
    Bars show mean noise-ceiling-normalized correlation across subjects, individual points represent subjects, and error bars indicate the standard error of the mean.
    Asterisks indicate significant differences between $\mathrm{SCRIBE}_{C}$ and the corresponding baseline ($p < 0.05$; one-sided Wilcoxon signed-rank test).
    }
    \label{fig:appendix_llama2-b_scribe_qaemb_corrs}
\end{figure}

Figure~\ref{fig:appendix_llama2-b_scribe_qaemb_corrs} shows the whole-brain and region-wise comparison between the non-contextual (layer $0$) and contextual (layer $15$) Llama2-7B embeddings, $\mathrm{QAEmb}$, and $\mathrm{SCRIBE}_{C}$.
Compared with non-contextual Llama2-7B embeddings, 
$\mathrm{SCRIBE}_{C}$ achieved substantially higher whole-brain prediction accuracy ($0.1837 \pm 0.0144$ vs.\ $0.1185 \pm 0.0106$), corresponding to a 55\% relative improvement.
$\mathrm{SCRIBE}_{C}$ also significantly outperformed the non-contextual embeddings in the Late Language ROIs, AG, LTC, and IFG/MFG.
In contrast, contextual Llama2-7B embeddings achieved higher overall brain-encoding performance than $\mathrm{SCRIBE}_{C}$ ($0.2387 \pm 0.0215$ vs.\ $0.1837 \pm 0.0144$).

\begin{figure}[h]
    \centering
    \begin{subfigure}{0.6\linewidth}
        \centering
        \includegraphics[width=\linewidth]{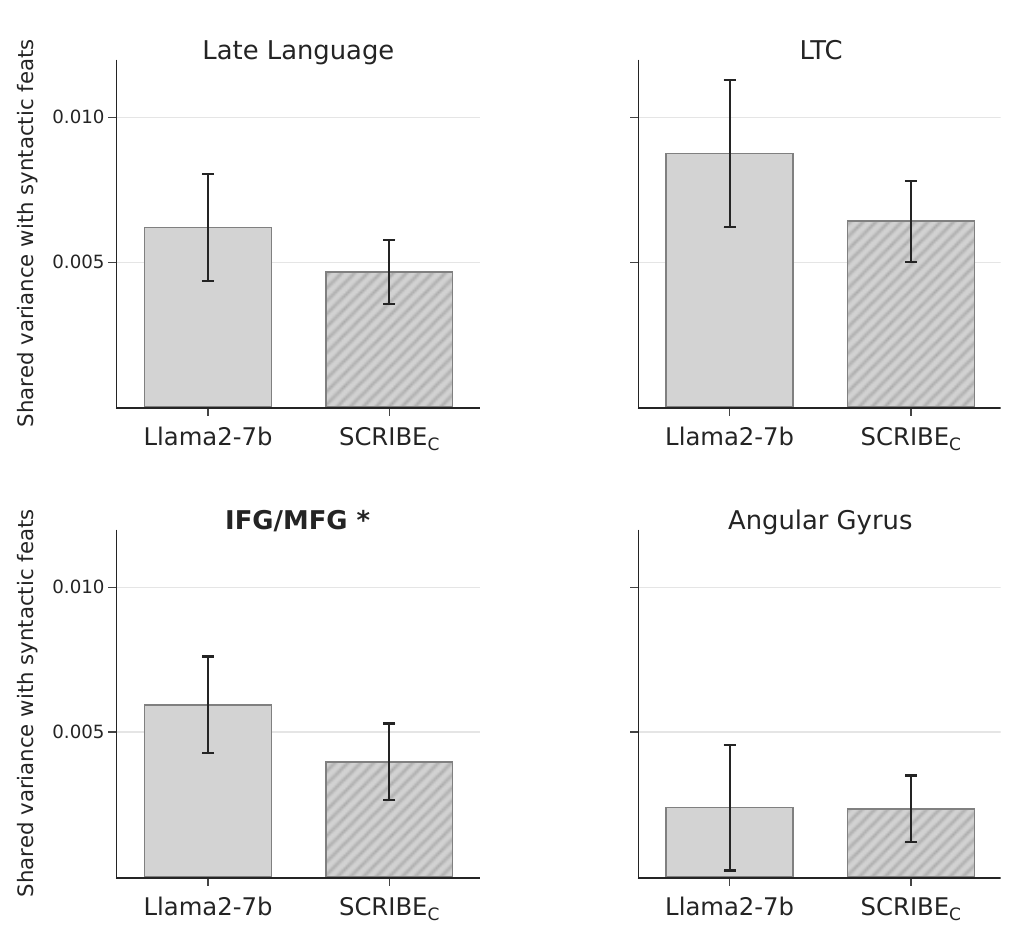}
        \caption{Shared variance with syntax.}
        \label{fig:shared_vp_llama_vs_scribe_concat-synt}
    \end{subfigure}
    \hfill
    \caption{
    Shared brain-predictive variance with syntactic features for contextual Llama2-7B and \method{} ($\mathrm{SCRIBE}_\mathrm{C}$).
    Bars show the mean shared variance across subjects and error bars indicate the standard error of the mean.
    \method{} shares significantly less variance with the syntactic feature space than Llama2-7B in IFG/MFG (one-sided paired Wilcoxon signed-rank test, $p<0.05$), indicated by the bold ROI label and asterisk.
    }
    \label{fig:shared_vp_synt_feats_llama}
\end{figure}

The higher predictive performance of contextual Llama2-7B embeddings does not, however, establish which linguistic information drives this advantage.
Because contextual LLM embeddings may encode multiple sources of linguistic information, we used variance partitioning with explicit syntactic features to further characterize the information underlying their brain predictivity.
Figure~\ref{fig:shared_vp_synt_feats_llama} compares the variance that contextual Llama2-7B and \method{} share with the syntactic feature space.
\method{} exhibits significantly less shared variance than Llama2-7B in IFG/MFG, with the same numerical direction in Late Language and LTC.
This difference is particularly notable given the established involvement of left inferior frontal regions in syntactic processing~\citep{dapretto1999form,heim2003distributed,friederici2011brain}.

\begin{figure}[h]
    \centering
    \includegraphics[width=\linewidth]{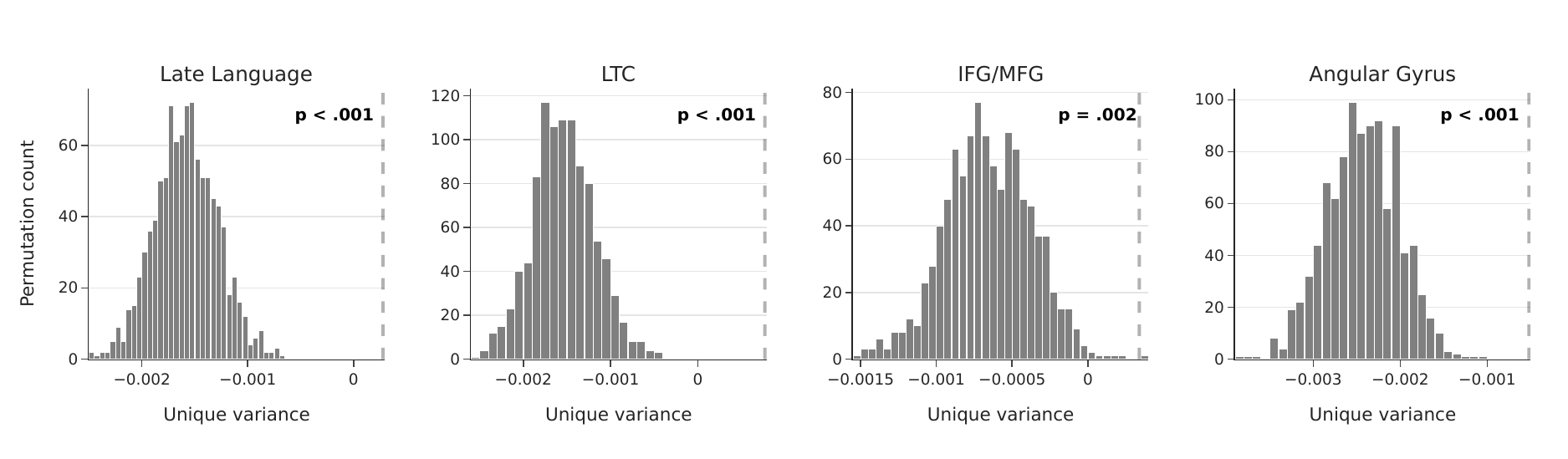}
    \caption{
    Permutation-based significance testing of the unique neural variance explained by $\mathrm{SCRIBE}_\mathrm{C}$ in variance partitioning with Llama2-7b.
    Histograms show permutation-based null distributions, and dashed lines indicate the observed unique variance.
    }
    \label{fig:permuted_unique_vp_scribe_llama}
\end{figure}

Finally, we tested whether the variance uniquely explained by \method{} exceeds chance.
Permutation tests confirmed significant unique variance across all four language-related ROIs (Late Language, LTC, IFG/MFG, and AG; all $p \leq .002$; Figure~\ref{fig:permuted_unique_vp_scribe_llama}).
Thus, despite the higher overall brain-prediction accuracy of contextual Llama2-7B embeddings, \method{} captures significant unique brain-predictive information and shows reduced overlap with syntactic features in IFG/MFG.

\end{document}